%% file: interfpd.tex
\documentclass[runningheads,orivec]{llncs}
\usepackage{interfpd}

\title{\centerline{A second-order theory of texture for depth from focus}}
\titlerunning{A second-order theory of texture for DFF}

\author{Sreekar Sai Ranganathan\orcidlink{0000-0001-7287-9761} \and Ioannis Gkioulekas\orcidlink{0000-0001-6932-4642}}
\authorrunning{S. S. Ranganathan, I. Gkioulekas}
\institute{Carnegie Mellon University, Pittsburgh PA, USA\\
\email{\{ssrangan,igkioule\}@andrew.cmu.edu}
}

\newcommand*{\illusionTeaserFigure}{
	\begin{figure}[t]
		\centering
		\includegraphics[width=\linewidth]{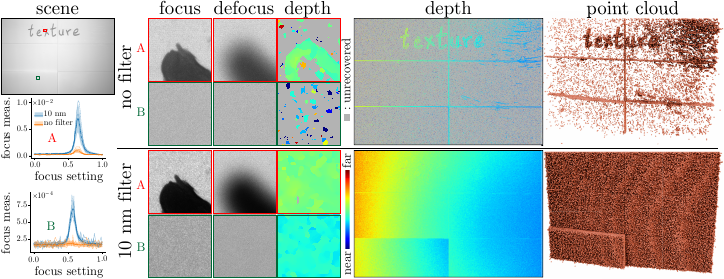}
		\vspace{-2em}
		\caption{Computer vision traditionally considers depth recovery of scenes with sparse texture features impossible with passive methods such as depth from focus (top). We show that \emph{simply adding a narrowband spectral filter to the camera} makes passive depth recovery of such scenes possible (bottom). We explain this surprising finding by developing a theory of \emph{second-order texture}---a form of subjective speckle whose in-focus contrast is enhanced with decreasing spectral bandwidth (crops to the left).}
		\label{fig:illusion-teaser}
	\end{figure}
}

\newcommand*{\typesOfTextureFigure}{
	\setlength{\columnsep}{0.5em}
	\setlength{\intextsep}{-0.15em}
	\begin{wrapfigure}[17]{r}{145pt}
		\centering
		\vspace{-0.5em}
		\includegraphics[width=\linewidth]{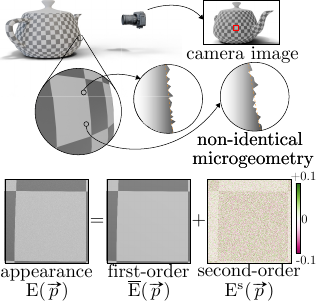}
		\vspace{-2em}
		\caption{First-order texture (checkerboard pattern) is due to visible BRDF variations. Second-order texture (speckle) is due to invisible random microgeometry even at first-order textureless regions.}
		\label{fig:texturetypes}
	\end{wrapfigure}
}

\newcommand*{\setupFigure}{
	\setlength{\columnsep}{0.5em}
	\setlength{\intextsep}{-0.15em}
	\begin{wrapfigure}[10]{r}{80pt}
		\centering
		\includegraphics[width=\linewidth]{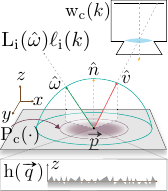}
		\vspace{-2em}
		\caption{Problem setup and notation.}
		\label{fig:setup}
	\end{wrapfigure}
}

\newcommand*{\coherenceAreasFigure}{
	\setlength{\columnsep}{0.5em}
	\setlength{\intextsep}{-0.15em}
	\begin{wrapfigure}[17]{r}{89pt}
		\centering
		\includegraphics[width=\linewidth]{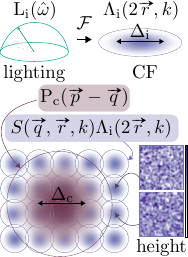}
		\vspace{-2em}
		\caption{Coherent summation within coherence areas results in speckle (random radiance). Incoherent summation of such areas within the PSF reduces speckle contrast.}
		\label{fig:coherence-areas}
	\end{wrapfigure}
}

\newcommand*{\mcsimFigure}{
	\setlength{\columnsep}{0.5em}
	\setlength{\intextsep}{-0.15em}
	\begin{wrapfigure}[16]{r}{170pt}
		\centering
		\vspace{-0.25em}
		\includegraphics[width=\linewidth]{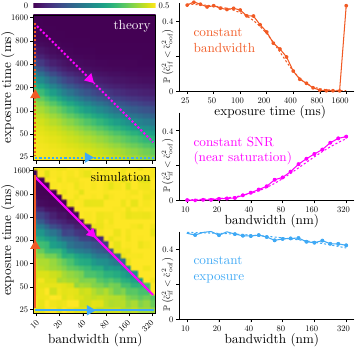}
		\vspace{-2em}
		\caption{Error probability $\Prob\sparen{\csqnoisyif < \csqnoisyoof}$ estimation from theory and simulation.}
		\label{fig:mc-sim}
	\end{wrapfigure}
}

\newcommand*{\underexposedFigure}{
	\setlength{\columnsep}{0.5em}
	\setlength{\intextsep}{-0.15em}
	\begin{wrapfigure}[9]{r}{170pt}
		\centering
		\includegraphics[width=\linewidth]{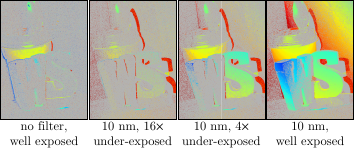}
		\vspace{-2em}
		\caption{DFF improves with a spectral filter even at suboptimal exposures. Example captured outdoors under sunlight.}
		\label{fig:underexposed}
	\end{wrapfigure}
}

\newcommand*{\rmsewrapFig}{
	\setlength{\columnsep}{0.5em}
	\setlength{\intextsep}{-0.15em}
	\begin{wrapfigure}[15]{r}{109.3pt}
		\centering
		\includegraphics[width=\linewidth]{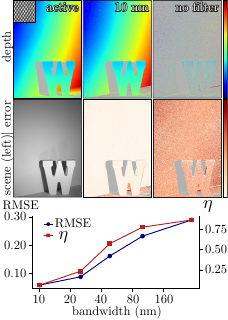}
		\vspace{-2em}
		\caption{Validation of the $\fractionunrecovered$ metric with active DFF.}
		\label{fig:rmsewrapfig}
	\end{wrapfigure}
}

\newcommand*{\figtwoTeaserFigure}{
	\begin{figure*}[t]
		\centering
		\includegraphics[width=\textwidth]{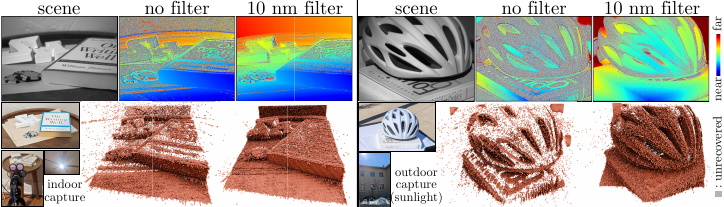}
		\vspace{-2em}
		\caption{We demonstrate fine-scale depth recovery of objects with sparse texture features under ambient lighting indoors (ceiling lights, left) and outdoors (sunlight, right). In both cases, adding a spectral filter dramatically improves DFF performance.
		}
		\label{fig:fig2-teaser}
	\end{figure*}
}

\newcommand*{\contrastevidenceFigure}{
	\begin{figure*}[t]
		\centering
		\includegraphics[width=\textwidth]{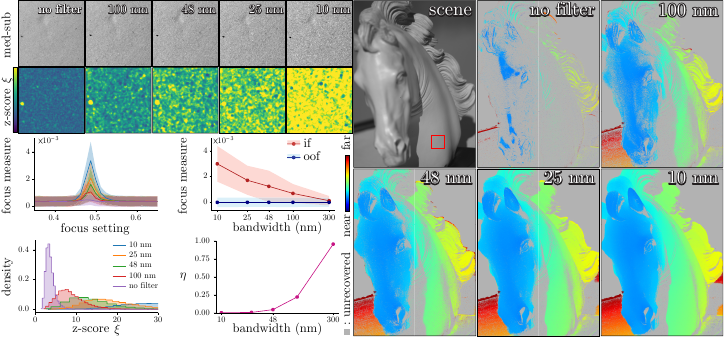}
		\vspace{-2em}
		\caption{
			DFF improves with filters of decreasing bandwidth thanks to enhanced second-order texture, visible in median-subtracted crops (`med-sub'). This enhancement also results in focus measure peaks more distinguishable from noise (top two plots), decreasing the fraction of pixels under a z-score threshold (bottom two plots, gray pixels in depth maps). Example captured outdoors under sunlight.}
		\label{fig:contrast-evidence}
	\end{figure*}
}

\newcommand*{\resolutionbenefitFigure}{
  \begin{figure*}[t]
    \centering
    \includegraphics[width=\textwidth]{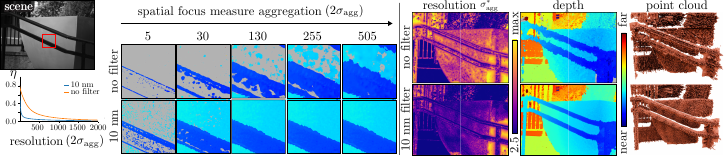}
	\vspace{-2em}
    \caption{Aggregating focus measures over progressively larger patch sizes allows depth recovery at adaptive scales. With the filter, much finer details are recovered at smaller patch sizes (see resolution map). Example captured outdoors under sunlight. }
    \label{fig:resolution-benefit}
	\vspace{1em}
\end{figure*}
}

\newcommand*{\exposureTradeoffFigure}{
	\begin{figure*}[t]
		\centering
		\includegraphics[width=\textwidth]{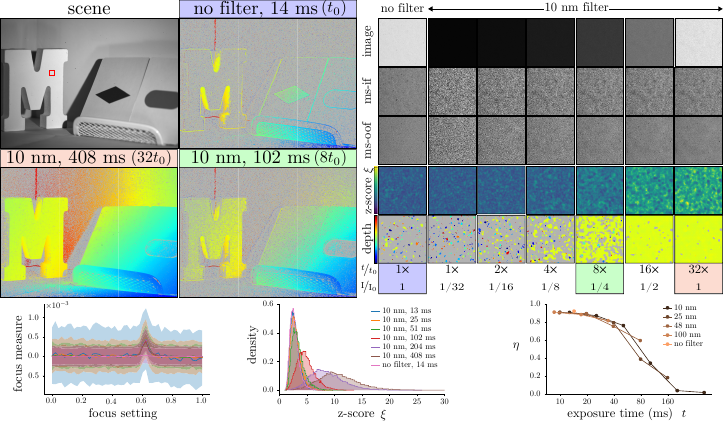}
		\vspace{-2em}
		\caption{Using a spectral filter improves DFF even at suboptimal exposures. At $\nicefrac{1}{4}$ the optimal exposure time, second-order texture is already discernible from noise at median-subtracted crops (`ms-if' vs `ms-oof'), improving focus measures and z-score statistics (bottom plots). Example captured indoors under ceiling lights.}
		\vspace{1em}
		\label{fig:exposure-tradeoff}
	\end{figure*}
}

\newcommand*{\spectralSpeckleChangeFigure}{
	\setlength{\columnsep}{0.5em}
	\setlength{\intextsep}{-0.15em}
	\begin{wrapfigure}[16]{r}{120pt}
		\centering
		\includegraphics[width=\linewidth]{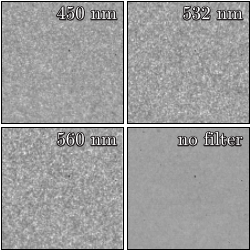}
		\vspace{-2em}
		\caption{In-focus images of a white target using different filters (center $\wavelength$) show uncorrelated speckle patterns. Their incoherent summation without a filter eliminates speckle.}
		\label{fig:speckleChange}
	\end{wrapfigure}
}

\newcommand*{\galleryResultsFigure}{
  \begin{figure*}[t]
    \centering
    \includegraphics[width=\textwidth]{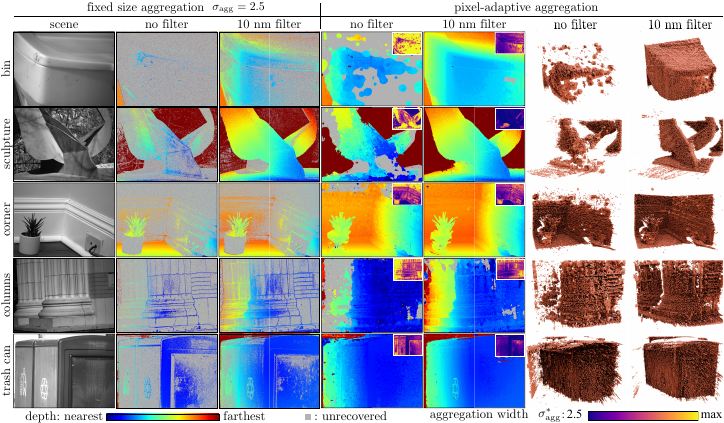}
    \caption{Examples of DFF improvement in indoor and outdoor real-world scenes. Using a spectral filter improves both fixed-size and pixel-adaptive reconstructions. The project website provides interactive visualizations of these and other scenes.}
    \label{fig:other-results}
\end{figure*}
}

\begin{document}

\addtocontents{toc}{\protect\setcounter{tocdepth}{-10}}
\maketitle

\begin{abstract}
	We present a theory of textured appearance of optically rough surfaces based on wave optics, emphasizing the role of texture for passive depth from focus. Our theory shows that even surfaces that traditional computer vision would consider textureless can produce textured appearance, due to subjective speckle from surface microgeometry. We analyze the properties of this \emph{second-order texture}, and show that we can enhance its contrast under natural ambient lighting by simply using a narrowband spectral filter. Doing so results in dramatic improvements in passive depth reconstruction of seemingly textureless scenes, as we demonstrate through extensive theory, simulations, and real-world experiments.
\end{abstract}

\vspace{-1.5em}
\section{Introduction}\label{sec:intro}

\illusionTeaserFigure

\emph{Are textureless scenes recoverable?} \Citet{sundaram1997textureless} posed this fundamental question to the computer vision community nearly three decades ago. Their analysis suggested that little geometrical information exists in images of a textureless scene taken by a standard camera, thus greatly hindering \emph{passive depth sensing} of such scenes. In the succeeding years, computer vision has made great strides towards developing 
methods for passive depth sensing, leveraging focus or parallax cues that can be inferred from texture even under uncontrolled ambient illumination. Despite these advances, the assumption has remained that passive depth sensing of textureless scenes is difficult, if not impossible. In this paper, we challenge this assumption: we show that merely adding a narrowband spectral filter in front of an otherwise standard lens-based camera allows using depth from focus (DFF) to recover textureless objects (\cref{fig:illusion-teaser}).

Understanding this surprising finding requires examining another fundamental question in computer vision: \emph{What is texture?} The conventional definition requires spatial variations of surface reflectance or normal at a scale resolvable by the camera at the working imaging magnification, manifesting as spatially varying intensities in focused images of the surface. Thus, a surface with spatially constant bidirectional reflectance distribution function (BRDF)---the macroscopic description of reflectance---would be considered ``textureless''. 

We challenge this definition by revisiting classical surface appearance models based on wave optics. Our analysis predicts that even such ``textureless'' surfaces will produce non-constant in-focus images showing \emph{subjective speckle}---high-frequency intensity variations due to diffraction at surface microgeometry that the camera cannot resolve. Understanding this effect requires analysis of second-order intensity statistics, instead of just the mean (first-order moment) intensity predicted by classical appearance models. We term this effect \emph{second-order texture}, to distinguish it from the classical \emph{first-order texture}.

Second-order texture will be useful for DFF only if it has high-enough contrast to overcome sensor noise. Prior work assumes that significant subjective speckle is possible only under coherent illumination or at microscopic magnifications. We show that using a narrowband spectral filter (\qtyrange{10}{100}{\nano\meter}) results in high-contrast second-order texture under typical passive computer vision conditions---outdoor and indoor ambient lighting, macroscopic magnifications. Using such a filter also reduces incident flux, necessitating increased exposure time to maintain signal-to-noise ratio (SNR). We show that, even under severe underexposure (3--4 stops), second-order texture is discernible from noise. On the whole, using a narrowband spectral filter can dramatically improve DFF performance (\cref{fig:fig2-teaser}).

To present our findings, first, we provide background on DFF, and characterize depth recoverability conditions as a function of texture contrast and SNR (\cref{sec:background}). Next, we develop our second-order theory of texture, and analyze the factors impacting the contrast of second-order texture (\cref{sec:theory}). Then, we use this analysis to assess, through theory and simulations, how DFF performance varies with filter spectral bandwidth and exposure time (\cref{sec:recoverability}). Last, we perform DFF experiments under ambient indoor and outdoor illumination that validate our findings (\cref{sec:experiments}). The supplemental PDF includes all proofs and additional experimental results. The project website%
\footnote{\projectlink{}}
includes interactive visualizations, code, and datasets.

\figtwoTeaserFigure

\section{Related work}

\paragraph{Passive depth sensing in computer vision.} Passive methods use \emph{parallax} or \emph{focus} to infer depth under ambient illumination. Parallax methods such as stereo \citep{Barnard1980disparity,Nalpantidis2008stereo,hartley_zisserman_2004,scharstein2002taxonomy} infer depth by triangulating correspondences across multi-view images. Among focus methods, depth from focus \citep{Supasorn2015,hazirbas18ddff,grossman1987focus,nayar1994shape,nayar1996real,subbarao1995accurate} infers depth by assessing per-pixel focus in a dense focal stack. Depth from defocus \citep{Subbarao1994defocus,favaro2010recovering,Pentland87,Tang2017,watanabe1998rational} uses two images at different focus settings to estimate defocus blur and thus depth. Other methods change both focus and aperture (confocal stereo \citep{hasinoff2009confocal}), vary focus differentially (focal flow \citep{alexander2016focal,guo2017focal}), or engineer high-frequency defocus blur (coded aperture \citep{levin2007image,veeraraghavan2007dappled,zhou2009good,Zhou2009ICCV,chakrabarti2012depth}). Focus methods can be analyzed as stereo with a baseline equal to the lens aperture \citep{schechner2000depth}, a relationship used in lightfield \citep{bolles1987epipolar,kim2013scene} and dual-pixel \citep{punnappurath2020modeling,Punnappurath_2019_CVPR,garg2019learning,xin2021defocus} methods. Unfortunately, these methods fundamentally require \emph{texture}, to either establish correspondence or assess focus, and are unreliable in textureless scenes. 

\paragraph{Passive interferometric imaging.} Several methods leverage the weak coherence of ambient illumination through \emph{interferometry}---correlation measurements of superimposed light waves---for passive 3D imaging tasks. \emph{Incoherent digital holography} methods take interferometric measurements to create holographic scene representations \citep{rosen2019recent,liu2018incoherent,tahara2022roadmap,tahara2017single,tahara2022palm,muroi2023capturing}. Other such methods can perform 3D localization for occluded imaging \citep{Davy2013,Badon2015,Badon2016,BogerLombard2019,Batarseh2018}. Closer to our work, interferometry has been used for passive depth sensing---through focus \citep{cossairt2014digital}, parallax \citep{chen2024coherence}, or time-of-flight \citep{kotwal2023passive}---of even textureless scenes. Unfortunately, these methods require complex and sensitive optical systems, limiting their practical utility. Like these methods, ours achieves textureless passive depth sensing by leveraging weak coherence effects---subjective speckle from ambient illumination---but it does so through just a very simple modification to a standard depth-from-focus system---mounting a narrowband spectral filter in front of the lens.

\paragraph{Speckle imaging.} Speckle has been extensively studied in optics since the invention of the laser. We refer to dedicated textbooks \citep{goodman2020speckle,mertz2019introduction,akkermans2007mesoscopic}, and limit our discussion to literature in vision and graphics. Computer vision methods have used speckle for tamper detection \citep{shih2012laser}, motion tracking \citep{jo2015spedo,smith2017colux,smith2018tracking}, vibrometry \citep{kichler2025learning,sheinin2022dual}, imaging around corners \citep{metzler2020deep}, and through occluders \citep{xie2024wavemo,alterman2021imaging,boniface2019noninvasive}. All these methods are \emph{active}, introducing coherent illumination into the scene to induce speckle. These applications have also motivated work on speckle rendering \citep{kim2025monte,bar2020rendering,bar2019monte,steinberg2022rendering,liu2025fully}. Lastly, speckle modeling is at the foundation of classical appearance models for rough surfaces \citep{beckmann1987scattering,stam1999diffraction,levin2013fabricating,dong2015predicting}, a foundation we revisit to develop our theory.

\section{Depth from focus \& recoverability conditions}\label{sec:background}

We begin with background on \emph{depth from focus} (DFF) and an analysis of recoverability conditions, focusing on the role of texture and sensor noise. 
DFF uses a \emph{focal stack} of images, captured by stepping the focusing depth of the camera lens. We denote this stack as $\scurly{\intensitynoisy_\focusindex\sparen{\pointcam}}_{\focusindex=1}^{\numfocusindex}$, where $\intensitynoisy_\focusindex\sparen{\pointcam}$ is the sensor's intensity measurement at the pixel $\pointcam$ and focus index $\focusindex$
---we use 2D vectors $\pointcam$ to indicate pixel center locations, and tildes to highlight quantities impacted by sensor noise, such as the noisy intensity $\intensitynoisy$ and its functionals. 
Depth recovery requires determining for each pixel $\pointcam$ the (most) in-focus index $\focusindexif\sparen{\pointcam}$.

DFF estimates $\focusindexif\sparen{\pointcam}$ by finding the image that maximizes a measure of spatial intensity variation (or ``sharpness'') for a patch $\neighborhood_\numsamples\sparen{\pointcam}$ of $\numsamples$ pixels (\eg, $\sqrt{\numsamples}\times\sqrt{\numsamples}$-square) around $\pointcam$. Several such \emph{focus measures} are available \citep{subbarao1998selecting}; we consider the \emph{squared sample coefficient of variation} $\csqnoisy_{\focusindex}\sparen{\pointcam} \coloneq \nicefrac{\fmvaluenoisy_{\focusindex}\sparen{\pointcam}}{\meanintnoisy_{\focusindex}^2\sparen{\pointcam}}$ %
computed from the \emph{sample mean} and \emph{sample variance} of gray patch intensities,
\begin{align}\label{eqn:sample_stats}
	\meanintnoisy_{\focusindex}\paren{\pointcam} \coloneq \frac{1}{\numsamples}\!\sum_{\pointcam'\in\neighborhood_\numsamples\paren{\pointcam}} \!\intensitynoisy_{\focusindex}\paren{\pointcam'}, \quad 
	\fmvaluenoisy_{\focusindex}\paren{\pointcam} \coloneq \frac{1}{\numsamples- 1}\!\sum_{\pointcam'\in\neighborhood_\numsamples\paren{\pointcam}} \!\paren{\intensitynoisy_{\focusindex}\paren{\pointcam'}- \meanintnoisy_{\focusindex}\paren{\pointcam}}^2.
\end{align}
We use $\csqnoisy_{\focusindex}\sparen{\pointcam}$ because it simplifies analysis, but our findings extend to other common focus measures (\cref{sec:experiments}). DFF estimates $\focusindexif\sparen{\pointcam}$ as $\focusindexestimate\sparen{\pointcam} \coloneq \argmax_{\focusindex} \csqnoisy_{\focusindex}\sparen{\pointcam}$. We refer to $\csqnoisy_{\focusindex}\sparen{\pointcam}$ as simply \emph{sample contrast}, and omit $\pointcam$ where convenient.
 
The justification for estimating $\focusindexif$ as  $\focusindexestimate$ is that the contrast $\csqnoisy_{\focusindex}$ will be maximal at the in-focus index because defocus acts as a low-pass filter. This justification is valid under two conditions \citep{subbarao1998selecting,nayar1994shape}:
\begin{enumerate}[nosep]
	\item \emph{Texture condition:} The in-focus patch $\neighborhood_\numsamples\sparen{\pointcam}$ has appreciable \emph{texture}, \ie, spatial intensity variations, in the absence of noise.
	\item \emph{Signal-to-noise ratio condition:} The sensor noise is sufficiently low so that intensity variations due to texture are discernible from those due to noise.
\end{enumerate}
These conditions are affected by scene appearance (surface reflectance, lighting), imaging conditions (lens aperture, exposure, sensor noise), and algorithmic choices (patch size $\numsamples$). In the rest of this section, we characterize the impact and interplay of these factors. We first distinguish between the measured noisy intensities $\intensitynoisy_{\focusindex}$, and the corresponding \emph{noise-free intensities} $\intensity_{\focusindex}$, which we define formally below. We also define the \emph{noise-free contrast, sample variance}, and \emph{mean} $\csq_{\focusindex}$, $\fmvalue_{\focusindex}$, and $\meanint_{\focusindex}$ (respectively) analogously, and use them to define a notion of recoverability.
\begin{dfn}[label={def:recoverability}]{$\control$-recoverability}{recoverability}
	For any $\control \in \bracket{0, 1}$, we say that the in-focus index $\focusindexif$ is \emph{$\control$-recoverable} when $\Prob\sparen{\csqnoisyif < \csqnoisyoof} \le \control$, 
	where: 
	\begin{enumerate*}
		\item $\csqnoisyif$ is the focus measure at $\focusindexif$, and 
		\item $\csqnoisyoof$ is the focus measure at an out-of-focus index $\focusindexoof \neq \focusindexif$ with sufficient defocus such that noise-free intensities are nearly identical, thus $\csq_{\mathrm{oof}} \approx 0$.
	\end{enumerate*}
\end{dfn}
\noindent We use $\control$-recoverability as a weaker, but \emph{analytically tractable}, proxy of true recoverability to formalize the interplay of texture and noise (\cref{pro:marginal_recoverability}). We next characterize the statistics of noise-free and noisy intensities at $\focusindexif$ and $\focusindexoof$---denoted as $\intensityif$ and $\intensityoof$ for the noise-free, and analogously for the noisy case. 
 
\paragraph{Texture model.} The in-focus noise-free intensities $\intensityif\paren{\pointcam}$ are themselves random variables described by a statistical texture model. For weakly textured surfaces, as we define them in \cref{sec:analysis}, it suffices to consider the mean $\marginalmeanintensity$ and variance $\marginalvarintensity$ of their stationary pointwise distribution. 
From the definition of $\focusindexoof$ in \cref{def:recoverability}, the noise-free intensities $\intensityoof$ will all equal $\marginalmeanintensity$ and have zero variance. 

\paragraph{Noise model.} We can relate noisy to noise-free intensities at both $\focusindex\in \scurly{\focusindexif,\focusindexoof}$ using the standard Poisson--Gaussian model \citep{hasinoff2010noise}, $\intensitynoisy_{\focusindex} = \min\paren{\intensity_{\focusindex} + \noisevar_{\focusindex}, \adcmax}$, 
where:
\begin{enumerate*}
	\item $\intensity_{\focusindex}$ is proportional to the \emph{incident flux} $\flux_{\focusindex}$ and \emph{exposure time} $\exptime$, $\intensity_\focusindex \coloneq  \flux_\focusindex\exptime\, \nicefrac{\photoncurrentperflux}{ \gainvar}$, implying that $\marginalmeanflux \coloneq \E{\flux} = \nicefrac{\marginalmeanintensity\gainvar}{\exptime\photoncurrentperflux}$.
	\item The noise term $\noisevar_{\focusindex}$ is a random variable combining photon shot (Poisson) noise and read (Gaussian) noise. It has mean $\E{\noisevar_\focusindex \mid \intensity_\focusindex}=0$ and signal-dependent variance $\Var\sbracket{\noisevar_\focusindex \mid \intensity_\focusindex} = \intensity_\focusindex \nicefrac{1}{\gainvar} +  \nicefrac{\sigmaread^2}{\gainvar^2}$, where $\sigmaread^2 \coloneq \sigmapreamp^2 + {\gainvar^2\sigmapostamp^2}$. 
	\item The constant $\photoncurrentperflux$ depends on quantum efficiency, $\gainvar$ on ISO, $\sigmapreamp$ and $\sigmapostamp$ on readout circuitry, and $\adcmax$ on the saturation limit. 
\end{enumerate*}
This model assumes for simplicity no dark current or discretization, assumptions we revisit in \cref{sec:recoverability}. The marginal variance of noise $\noisevar_{\focusindex}$ thus equals $\marginalvarnoise \coloneq \Var\sbracket{\noisevar_\focusindex} =  \nicefrac{\marginalmeanintensity}{\gainvar} + \nicefrac{\sigmaread^2}{\gainvar^2}$, 
using the fact that the noise-free mean intensity equals $\marginalmeanintensity$ at $\focusindexif$ and $\focusindexoof$.

\paragraph{Recoverability condition.} Using these texture and noise models, we can define:
\vspace{-1em}
\begin{align}\label{eqn:cfvnoise}
	\cfvintensity \coloneq \frac{\marginalvarintensity}{\marginalmeanintensity^2}, 
	\quad \cfvnoise \coloneq \frac{\marginalvarnoise}{\marginalmeanintensity^2} = \frac{1}{\exptime\photoncurrentperflux\marginalmeanflux}\paren{1 + \frac{\sigmaread^2}{\exptime\photoncurrentperflux\marginalmeanflux}}.
\end{align}
The \emph{squared texture contrast} $\cfvintensity$ and \emph{reciprocal squared SNR} $\cfvnoise$ measure texture and noise strength. We can relate them to recoverability as follows. 
\begin{prp}[label={pro:marginal_recoverability}]{Necessary and sufficient condition for $\control$-recoverability}{recoverability}
	For any $\control \in \bracket{0, 1}$, the in-focus index $\focusindexif$ is $\control$-recoverable if and only if:
	\vspace{-1em}
	\def\proxyeqn{\cfvintensity \cdot \frac{1}{\cfvnoise} \cdot \sqrt{\frac{\numsamples -1}{2}}}
	\begin{align}\label{eqn:marginal_recoverability}
		\monofuncmarginal\Bigg(
			\mymathbox{intBlueL}{intBlueLL}{\centering texture}{\vphantom{\proxyeqn}\cfvintensity}
			\;\cdot\;
			\mymathbox{intTealL}{intTealLL}{\centering SNR}{\vphantom{\proxyeqn}\frac{1}{\cfvnoise}}	
			\Bigg)
			\cdot
			\mymathbox{intCrimL}{intCrimLL}{{\centering patch size}}{\vphantom{\proxyeqn}
			\sqrt{\frac{\numsamples -1}{2}}
			} 
			> \cdf^{-1}\paren{1 - \control},
			\vspace{-1em}
	\end{align}
	where: 
	\begin{enumerate*}
		\item $\monofuncmarginal\paren{x} \coloneq \nicefrac{x}{\sqrt{1 + \paren{x + 1}^2}}$ is a monotonically increasing function; and
		\item $\cdf^{-1}\paren{\cdot}$ is the inverse CDF of the standard normal distribution.
	\end{enumerate*}
\end{prp}

\Cref{pro:marginal_recoverability} (proved in supplement) provides an interpretable view of the three key factors impacting recoverability: texture contrast, SNR, and patch size. 

The green highlighted factor in \cref{eqn:marginal_recoverability} suggests improving recoverability by increasing SNR. Doing so is possible by increasing exposure time $\exptime$ (\cref{eqn:cfvnoise}), but only up to the saturation limit $\adcmax$. 
Further increasing SNR is possible by averaging captures from multiple sensor readouts. However, doing so increases acquisition time by more than the increase in total exposure time.

The brown highlighted factor in \cref{eqn:marginal_recoverability} suggests improving recoverability by increasing the patch size $\numsamples$---representing the decreasing impact of noise in the sample contrast $\csqnoisy_{\mathrm{if}}$ as samples increase. However, increasing $\numsamples$ comes at the cost of decreased lateral resolution and increased errors close to depth discontinuities (which are not accounted for in standard DFF).

The classical understanding of texture in computer vision suggests that we cannot improve the blue highlighted factor in \cref{eqn:marginal_recoverability} while maintaining passive operation---\ie, without active illumination to artificially texture surfaces. In what follows, we show this understanding to be incorrect: We present a \emph{second-order} theory of texture (\cref{sec:theory}) based on wave-optical scattering of light, which predicts that it is indeed possible to amplify texture \emph{by simply adding a narrowband spectral filter in front of the camera}. We provide extensive experimental verification (\cref{sec:experiments}) of this prediction, showing that the fundamental requirement for texture in passive depth recovery is not as severe as previously thought. Of course, adding a spectral filter also worsens the SNR factor in \cref{eqn:marginal_recoverability}, assuming fixed exposure time. We show (\cref{sec:recoverability}) that the improvement in texture contrast outweighs the worsening in SNR after only a sublinear increase in exposure time.

\section{A second-order theory of texture}\label{sec:theory}

Further analysis of $\cfvintensity$ in \cref{eqn:marginal_recoverability} requires characterizing the pointwise distribution of pixel intensities $\intensity\sparen{\pointcam}$. We assume for simplicity that the incident flux $\flux_\focusindex\sparen{\pointcam}$ at a pixel is proportional to the irradiance $\irradiance\sparen{\pointcam}$ at its center, $\flux_\focusindex\sparen{\pointcam} = \irradiance_\focusindex\sparen{\pointcam} \pixelresponsewidth^2$. 
It follows that $\cfvintensity  = \cfvirradiance\coloneq \nicefrac{\marginalvarirradiance}{\marginalmeanirradiance^2}$, thus it suffices to analyze $\cfvirradiance $. 

Our goal in this section is thus to characterize the irradiance $\irradiance\sparen{\pointcam}$. In \cref{sec:newtexture}, we use wave-optical principles of light scattering to derive an expression of $\irradiance\sparen{\pointcam}$ for a camera imaging \emph{optically rough} surfaces under ambient illumination. We then explain how this expression relates to appearance models in computer vision \citep{beckmann1987scattering,levin2013fabricating,stam1999diffraction,cuypers2012reflectance}, and why the appearance of real-world rough surfaces is \emph{always} textured due to {subjective speckle}, even for surfaces that classical computer vision would consider textureless. In \cref{sec:analysis} we quantify the strength of this textured appearance as a function of imaging and illumination conditions.

\subsection{First and second-order texture}\label{sec:newtexture}

\setupFigure{}To simplify exposition, and without loss of generality, we assume a thin-lens camera that images a planar, frontoparallel, and optically rough surface under far-field illumination (\cref{fig:setup}). Our problem setting and derivation closely adapt \citet[\S\S 5.8 \& 6.3]{goodman2020speckle}. We use a 3D coordinate system where the $z$ axis is parallel to the optical axis and the macroscopic surface normal $\normaldir$. We use 3D unit-norm vectors $\dirvec$ for directions, and $\dirvec_{xy}$ for their projection on the $x$--$y$ plane. We use 2D vectors $\pointcam$ on the $x$--$y$ plane interchangeably for both sensor and surface points, and identify each sensor point with the surface point it maps to through its central ray $\dirvecout$ (\ie, pinhole projection).

We model the camera lens using its focus-dependent \emph{point spread function} (PSF) $\psf\sparen{\pointcam - \pointcamalt}$, 
and use $\ssf\sparen{\wavenumber}$ for the spectral sensitivity function (SSF) of the sensor. We model the far-field illumination as a spectral radiance environment map that we assume separable in direction and wavenumber, $\radiancein\sparen{\dirvec, \wavenumber}\coloneq \radiancein\sparen{\dirvec} \spectruminc\sparen{\wavenumber}$, 
where $\wavenumber = \nicefrac{2\pi}{\wavelength}$ is the wavenumber and $\wavelength$ the wavelength,
assuming mutually incoherent sources at different directions $\dirvec$. 
Our goal is to compute the irradiance reaching the sensor due to the scattering of $\radiancein\sparen{\dirvec, \wavenumber}$ at the surface point $\pointcam$.

If we zoomed in on the surface around $\pointcam$ (\cref{fig:texturetypes}), we would observe microscopic roughness that we model in Monge form using a \emph{random heightfield} $\height$. This roughness has scale comparable to the visible wavelength (hence, \emph{optically rough}) and is not resolvable by the camera, thus the surface appears macroscopically planar. Instead, the roughness is responsible for the surface's wave-optical scattering behavior, determining its \emph{appearance}. We model this appearance using the Fourier transform of a functional of the heightfield, as follows.

\begin{dfn}[label={def:fs_brdf}]{Sample BRDF}{fs_brdf}
Given a local surface heightfield $\height\paren{\cdot}$, we define the \emph{sample bidirectional reflectance distribution function} (sample BRDF) $\fsrf$ as:
	\begin{align}\label{eqn:fs_brdf}
		\fsrf\paren{\pointcamalt, \hdirvec, \wavenumber} 
		\coloneq
		\frac{1}{\brdfconst} \int_{\R^2} \pstf\paren{\pointcamalt, \pointcamdiff, \wavenumber,\hdirz}
		e^{\imi \wavenumber {\hdirvec_{xy}}\cdot 2\pointcamdiff}\ud \pointcamdiff,
	\end{align}
	where $\pstf$ is a functional of the heightfield $\height$ and reflection coefficient $\reflectcoeff$,
	\begin{align}\label{eqn:surface_scattering}
		\begin{split}
		\pstf\paren{\pointcamalt, \pointcamdiff,\wavenumber, \hdirz} &\coloneq \scatter\paren{\pointcamalt\!+\!\pointcamdiff; \wavenumber, \hdirz} \scatter^\ast\paren{\pointcamalt\!-\!\pointcamdiff; \wavenumber, \hdirz},\\
		\;\;
		 \scatter\paren{\pointtwod; \wavenumber, \hdirz} &\coloneq 
		\reflectcoeff\paren{\pointtwod, \wavenumber} e^{\imi\wavenumber \hdirz \height\paren{\pointtwod}}, \quad \hdirvec\in\R^3.
		\end{split}
	\end{align}
	%
\end{dfn}
We elaborate on roughness assumptions and required properties for $\height$ in the supplement. We can use $\fsrf$, which is random through $\height$, to express the irradiance.%
\footnote{$\fsrf$ is also a scaled Wigner transform of $\scatter\paren{\cdot;\wavenumber,\hdirz}$.}

\begin{prp}[label={pro:irradiance}]{Appearance model}{irradiance}
	The irradiance received at sensor point $\pointcam$ equals:
	\begin{equation} \label{eqn:irradiance_exitance}
		\irradiance\paren{\pointcam} \approx \int_{\R^+} \int_{\R^2} \ssf\paren{\wavenumber} \spectruminc\paren{\wavenumber} \psf\paren{\pointcam -\pointcamalt} \radianceout\paren{\pointcamalt, \dirvecout, \wavenumber} \ud\pointcamalt \ud\wavenumber, \;\text{where the}
	\end{equation}
	\emph{outgoing radiance} $\radianceout$ at surface point $\pointcamalt$, direction $\dirvecout$, and wavenumber $\wavenumber$ is:
	\begin{equation} \label{eqn:fs_reflectance}
		\radianceout\paren{\pointcamalt, \dirvecout, \wavenumber} \coloneq \int_{\unithemi\paren{\normaldir}}\fsrf\paren{\pointcamalt, \dirvec+\dirvecout, \wavenumber} \radiancein\paren{\dirvec} \paren{\dirvec\cdot \normaldir} \ud \sigma\paren{\dirvec}.
	\end{equation}
\end{prp}

\paragraph{Relationship to classical appearance model.} \Cref{eqn:irradiance_exitance} is equivalent to the standard PSF-based image formation model for a thin lens. \Cref{eqn:fs_reflectance} is \emph{almost} the standard reflectance equation in computer vision, except it uses the sample BRDF $\fsrf$ instead of the standard BRDF $\brdf$ for rough surfaces as defined in, \eg, \citet{stam1999diffraction}. This difference is the basis for our texture theory: $\fsrf$, and thus also $\irradiance$, are random variables due to their dependence on $\height$. Classical appearance models eliminate this randomness by using \emph{ensemble averages} $\Exp{\height}{\fsrf}$, $\Exp{\height}{\irradiance}$ over all microgeometry realizations. We make the link to classical models explicit.

\begin{prp}[label={pro:ensemble}]{Ensemble averaging}{ensemble}
	Assuming ergodicity for $\height$, $\Exp{\height}{\fsrf\paren{\pointcamalt, \dirvec + \dirvecout, \wavenumber}} = \brdf\paren{\pointcamalt, \dirvec + \dirvecout, \wavenumber}$.
\end{prp}
From linearity of expectation, replacing $\fsrf$ with $\brdf$ in \cref{eqn:irradiance_exitance,eqn:fs_reflectance} produces the classical radiometric equations for the ensemble-averaged irradiance. 

\paragraph{Subjective speckle as second-order texture.} We use the notation:
\begin{equation}\label{eqn:speckle}
	 \foirradiance\paren{\pointcam} \coloneq \Exp{\height}{\irradiance\paren{\pointcam}}, \quad \soirradiance\paren{\pointcam} \coloneq   \irradiance\paren{\pointcam} - \foirradiance\paren{\pointcam}.
\end{equation}

\typesOfTextureFigure{}The difference $\soirradiance$ between irradiance and its ensemble-average is known in optics as \emph{subjective speckle} \citep{steinberg2022rendering,goodman2020speckle}. 
$\soirradiance$ arises because of wavelength-scale surface microgeometry that \emph{cannot be resolved} by the camera at the working magnification. Resolvable changes in surface geometry or material, including visible surface roughness at high magnifications \citep{nayar1994shape}, are instead represented in the BRDF $\brdf$ (which includes foreshortening), and thus $\foirradiance$.

Returning to DFF, classical computer vision theory defines ``texture'' as spatial variations of the ensemble-averaged irradiance $\foirradiance$ (\cref{fig:texturetypes}). 
However, the presence of subjective speckle $\soirradiance$ suggests that even a surface that would classically be considered \emph{textureless}---\ie, one with spatially constant BRDF---can show irradiance variations due to non-resolvable microgeometry. We use the terms \emph{first-order} and \emph{second-order texture} to distinguish the two types of variations, as one considers only the first moment of $\irradiance$ (ensemble average $\foirradiance$), whereas the other considers also second moments\footnote{$\soirradiance$ captures all higher-order moments; we focus on second-order moments in the context of \cref{pro:marginal_recoverability} for DFF.} of $\irradiance$ (variance of subjective speckle $\soirradiance$).

\paragraph{When is ensemble averaging accurate?} The ensemble averaging assumption of classical computer vision is accurate under geometric camera models (\eg, perspective, orthographic) with an infinitesimal aperture and thus infinite PSF in \cref{eqn:irradiance_exitance}. Most previous derivations of appearance models for rough surfaces \citep{beckmann1987scattering,levin2013fabricating,stam1999diffraction} assume an infinite PSF, making them inaccurate for lens-based cameras under large-aperture or close-focus conditions---and thus DFF. Our derivation of \cref{pro:irradiance} assumes finite PSFs to study implications of deviations from ensemble averaging for DFF.

\subsection{Characterizing second-order texture contrast}\label{sec:analysis}

Our next goal is to derive an expression for the coefficient of variation $\cfvirradiance$---and thus $\cfvintensity$---that accounts for both first-order and second-order texture. We can then use this expression to assess the practicality of second-order texture for DFF, and characterize recoverability improvements.%
We define 
\begin{align}\label{eqn:coherence_function_defn}
	\mcfint\paren{2\pointcamdiff, \wavenumber} \coloneq \int_{\unithemi\paren{\normaldir}}\radiancein\paren{\dirvec}\paren{\dirvec\cdot \normaldir} e^{\imi \wavenumber 2\pointcamdiff\cdot\dirvec_{xy}} \ud \sigma\paren{\dirvec}
\end{align}
as the \emph{coherence function} (CF) of the far-field illumination $\radiancein$, \ie its (projected) Fourier transform. The effective support size of $\mcfint$ is the \emph{coherence area} $\scl^2$ \citep{levin2013fabricating,wolf2007introduction}, which increases as the illumination's directional footprint decreases 
Combining \cref{eqn:coherence_function_defn,eqn:fs_brdf,eqn:fs_reflectance} gives $\radianceout\sparen{\pointcamalt, \dirvecout, \wavenumber} = \mathcal{F}_{\pointcamdiff}\scurly{\pstf\cdot\mcfint}\paren{\dirvecout}$, which we use with \cref{pro:irradiance} to rewrite $\irradiance\sparen{\pointcam}$ as (denoting $\ssfspectrumincprod\paren{\wavenumber}= \ssf\paren{\wavenumber} \spectruminc\paren{\wavenumber}$):
\vspace{-0.5em}
\begin{equation}
	\mymathboxnarrow{intGrayL}{intGrayLL}{\vspace{-0.5 em}\centering spectral integration over band $\bwk$}{\int_{\R^+}\!\!\! \ssfspectrumincprod\paren{\wavenumber}
	\mymathboxnarrow{intCrimL}{intCrimLL}{\vspace{-0.5 em}\centering spatial integration over area $\dfl^2$}{\int_{\R^2}\!\!\! \psf\paren{\pointcam -\pointcamalt} 
	\mymathboxnarrow{intBlueL}{intBlueLL}{\vspace{-0.5 em}\centering spatial integration over area $\scl^2$}{\int_{\R^2}\!\!\! \pstf\paren{\pointcamalt, \pointcamdiff, \wavenumber}
	\mcfint\paren{2\pointcamdiff, \wavenumber} e^{\imi \wavenumber 2\pointcamdiff\cdot \dirvecout_{xy}} \ud{\pointcamdiff}} \ud\pointcamalt} \ud\wavenumber}.
	\vspace{-0.5em}
\end{equation}
\coherenceAreasFigure{}
We can interpret this expression as follows (\cref{fig:coherence-areas}):
\begin{enumerate}
	\item The innermost integral equals $\radianceout\sparen{\pointcamalt, \dirvecout, \wavenumber}$, and aggregates complex and random (due to the dependence of $\pstf$ on $\height$) values over points $\pointcamdiff$ in an area of size $\scl^2$ controlled by the CF $\mcfint$. As the integrand is the product of conjugate-symmetric functions, the integral is real-valued\footnote{$\radianceout$ can take negative values, similar to forms of a generalized radiance~\citep{testorf2010phase,steinberg2026wavetracing}.} for all $\pointcamalt$. 
	\item The intermediate integral equals the spectral density of irradiance $\irradiance\sparen{\pointcam,\wavenumber}$, and aggregates real random $\radianceout\sparen{\pointcamalt, \dirvecout, \wavenumber}$ values over points $\pointcamalt$ in an area of focus-dependent size $\dfl^2$ controlled by the PSF $\psf$.
	\item The outermost integral aggregates real random $\irradiance\sparen{\pointcam,\wavenumber}$ values over wavenumbers $\wavenumber$ in a spectral range $\bwk$ controlled by the illuminant $\spectruminc$ and SSF $\ssf$.
\end{enumerate}
Using optics terminology, the innermost integral performs \emph{coherent summation} within each coherence area---resulting in random $\radianceout\sparen{\pointcamalt, \dirvecout, \wavenumber} $ values manifesting as speckle---whereas the other two perform \emph{incoherent summation} within the PSF area and spectral bandwidth. Assuming $\height$ has sufficiently fast variation within each coherence area, the random $\radianceout\sparen{\pointcamalt, \dirvecout, \wavenumber}$ values are approximately independent for points $\pointcamalt$ separated by more than $\scl$ (\ie, non-overlapping $\scl^2$-sized integration areas), or for sufficiently separated wavenumbers $\wavenumber$---manifesting as uncorrelated speckle. Then, we can understand $\irradiance\sparen{\pointcam}$ as averaging independent real-valued random radiances whose number scales proportionally to $\bwk \nicefrac{\dfl^2}{\scl^2}$. Consequently, the variance of $\irradiance\sparen{\pointcam}$ is inversely proportional to this number, tending to 0 as defocus (and thus $\dfl^2$) or bandwidth (and thus $\bwk$) increase---reducing speckle contrast and converging to ensemble averaging (\cref{fig:speckleChange})---while attaining its maximal value when in focus. We formalize this discussion next.

\begin{prp}[label={pro:cfvirradiance}]{Pointwise coefficient of variation of irradiance}{cfv}
	In focus, the coefficient of variation of intensity and irradiance equal
	\vspace{-1em}
	\begin{align}\label{eqn:total_texture}
		\cfvintensity = \cfvirradiance \approx \mymathbox{intBlueL}{intBlueLL}{\centering second-order}{\frac{1}{\numspectralbuckets\numwindows}\paren{1 + \numwindows\cfvfoirradiance}} + \mymathbox{intCrimL}{intCrimLL}{\centering first-order}{\vphantom{\frac{1}{\numspectralbuckets\numwindows}}\cfvfoirradiance},
		\vspace{-1em}
	\end{align}
	where:
	\begin{enumerate*}
		\item $\cfvfoirradiance$ is the in-focus coefficient of variation for first-order texture $\foirradiance$;
		\item $\numwindows \coloneq \pi \nicefrac{\dflif^2}{\scl^2}$ is the relative size of the in-focus PSF and coherence area; 
		\item $\numspectralbuckets \coloneq G_1 \bwk$ is proportional to the spectral bandwidth, through a constant $G_1$ that depends on statistical properties of $\height$ and the mean wavevector.
	\end{enumerate*}
\end{prp}

\spectralSpeckleChangeFigure{}Notably, we consider a surface without first-order texture, meaning constant ensemble-averaged irradiance $\foirradiance$, and thus $\cfvfoirradiance = 0$. Such a surface would be classically described as ``textureless.'' Yet, \cref{eqn:total_texture} predicts non-zero in-focus texture contrast $\cfvintensity = \nicefrac{1}{\numspectralbuckets\numwindows}$, due to second-order texture. Therefore, from \cref{pro:marginal_recoverability}, even this \emph{first-order textureless} surface is potentially recoverable, so long as the contrast is stronger than noise. \Cref{eqn:total_texture} further predicts that we can increase this contrast by reducing the bandwidth $\bwk$ and thus $\numspectralbuckets$---\eg, by using a spectral filter.%
\footnote{The same principle applies to using a \emph{polarizer} in front of the camera to improve texture contrast, as the two orthogonal (incoherent) polarization states from unpolarized light generally produce uncorrelated speckle~\citep[\S 4.5]{goodman2020speckle}.}
Assessing recoverability requires considering the impact on SNR, as we do in \cref{sec:recoverability}.

\paragraph{Practicality for computer vision.} The factor $\numwindows$ in \cref{eqn:total_texture} helps understand under what conditions second-order texture is practical for DFF. Its in-focus contrast improves with larger coherence areas $\scl^2$ due to illumination dominated by concentrated far-field sources (\eg, outdoor sunlight, indoor small ceiling lights), and worsens with more directionally uniform illumination (\eg, outdoor overcast conditions, indoor large ceiling panels). It also improves with smaller in-focus PSF sizes $\dflif^2$, \ie, smaller f-numbers or reproduction ratios~\citep{kingslake1945effective}. In \cref{sec:experiments} we show that, using bandwidths $\bwl = \qtyrange{10}{100}{\nano\meter}$, 
second-order texture improves DFF in settings typical for passive computer vision: 
\begin{enumerate*}
	\item both outdoor and indoor ambient lighting ($\scl \approx \qtyrange{50}{100}{\micro\meter}$ \citep{kotwal2023passive});
	\item reproduction ratios $\nicefrac{1}{50}$--$\nicefrac{1}{10}$ at f-numbers up to 6 ($\dflif \approx \qtyrange{30}{150}{\micro\meter}$).
\end{enumerate*}
Second-order contrast reduces also with larger pixel sizes \citep{gkioulekas2015transient,chen2024coherence}, favoring smaller pixel pitches ($\qty{3}{\micro\meter}$ in our experiments). We elaborate in the supplement.

\section{Recoverability with second-order texture}
\label{sec:recoverability}

We focus on recoverability for a first-order textureless surface, \ie, $\cfvfoirradiance=0$. From \cref{pro:marginal_recoverability}, it suffices to express the \emph{texture contrast--SNR product} $\contrastsnrprodmarginal^2$ as a function of bandwidth $\bwk$ and exposure time $\exptime$. \Cref{eqn:total_texture} already provides an expression for $\cfvintensity$; whereas from \cref{eqn:cfvnoise}, getting an expression for $\cfvnoise$ requires relating the mean flux $\marginalmeanflux$ to $\bwk$. We prove the following.

\begin{prp}[label={pro:snr}]{Texture contrast--SNR product}{snr}
	In focus, the texture contrast--SNR product $\contrastsnrprodmarginal^2 \coloneq \nicefrac{\cfvintensity}{\cfvnoise}$ equals:
	\vspace{-0.25em}
	\begin{align}\label{eqn:contrastsnrproductsq}
		\contrastsnrprodmarginal^2\paren{\exptime, \bwk} &\approx \frac{1}{\numwindows}
		 \frac{G_2}{G_1} 
		 \frac{\exptime}{\paren{1 + \frac{\sigma^2_{\rm read}}{G_2} \frac{1}{\bwk \exptime}}},
		 \vspace{-1.5em}
	\end{align}
	where the exposure time $\exptime \in \sleftinc{0, \nicefrac{\gainvar \adcmax}{G_2\bwk}}$ has a saturation limit dependent on spectral bandwidth $\bwk$, and the constant $G_2$ depends on spectral reflectance.

\end{prp}

\paragraph{Analysis and simulation.} In \cref{fig:mc-sim}, we visualize the 
probability of error $\Prob\sparen{\csqnoisyif < \csqnoisyoof}$ predicted from \cref{eqn:contrastsnrproductsq,eqn:marginal_recoverability} for different bandwidths $\bwk$\mcsimFigure{}and exposure times $\exptime$. We also use Monte Carlo simulation to numerically estimate $\Prob\sparen{\csqnoisyif < \csqnoisyoof}$ and assess the accuracy of our theory. Assuming a first-order textureless surface, we sampled focal stacks of noisy speckle patterns, then computed sample contrast $\csqnoisy$. We sampled speckle using a statistical model detailed in the supplement, and sensor noise using the model of \cref{sec:background}. The supplement lists all parameters (illumination, surface, sensor, lens) and other details of the visualization and simulation. The results show closely matching theory and simulation predictions for $\Prob\sparen{\csqnoisyif < \csqnoisyoof}$ (lower is better). We also observe:
\begin{enumerate}[nosep,leftmargin=*]
	\item For constant bandwidth (a vertical line), the optimal $\exptime$ is the saturation limit. As texture contrast $\cfvintensity$ is fixed, maximizing $\exptime$ maximizes SNR $\nicefrac{1}{\cfvnoise}$ and $\contrastsnrprodmarginal^2$.
	\item For constant exposure time (a horizontal line), the optimal $\bwk$ is again determined by the saturation limit, \ie, maximizing $\bwk$ till intensities saturate. The decrease in $\cfvintensity$ is outweighed by the increase in $\nicefrac{1}{\cfvnoise}$, improving $\contrastsnrprodmarginal^2$.
	\item For constant bandwidth--exposure time product (a diagonal line, corresponding to constant SNR), optimizing $\contrastsnrprodmarginal^2$ requires minimizing $\bwk$ while proportionally increasing $\exptime$, to stay close to the saturation limit. 
	\item When decreasing $\bwk$, it is also possible to improve $\contrastsnrprodmarginal^2$ by increasing $\exptime$ \emph{sublinearly}, \ie, by less than the amount required to maintain constant SNR.
	\item Due to the saturation limit for a single capture (diagonal discontinuity in simulated results), lowering $\bwk$ with a filter enables much reduced probability of error (top--left corner) than without a filter (bottom--right corner).
\end{enumerate}
In the next section, we validate these observations through extensive experiments.

\contrastevidenceFigure

\section{Experiments \& limitations}\label{sec:experiments}

We perform DFF experiments across multiple scenes using \emph{only pre-existing ambient light}---either sunlight, or standard ceiling lights. 
The supplemental PDF and project website include more experiments, visualizations, code, and data.

\rmsewrapFig{}
\paragraph{Experimental setup.} Our setup uses a machine vision camera (FLIR BFS-U3-122S6M-C), a motorized lens mount for focus control, a photographic lens (Canon EF-S 60mm f/2.8 Macro USM), and narrowband spectral filters (Edmund Optics, different FWHM bandwidths, center wavelength $\approx \qty{530}{\nano\meter}$).

\paragraph{Implementation details.} Given a focal stack $\scurly{\intensity_{\focusindex}\paren{\pointcam}}_{\focusindex=1}^\numfocusindex$, we use a focus measure that aggregates $3\times 3$ Laplacian values over a Gaussian kernel of width $\aggkernelwidth=\nicefrac{5}{2}$ \citep{subbarao1998selecting}. Empirically, this measure performed better than the sample contrast. For simplicity, we continue to use $\csqlap$ for this measure.

\exposureTradeoffFigure
As we do not have access to ground-truth depth in general, we quantify DFF performance using the per-pixel \emph{robust z-score} $\zscore \coloneq \nicefrac{\svbar{\csqlap_{\ast} - \median\sparen{\csqlap_\focusindex}}}{\mad\sparen{\csqlap_\focusindex}}$, where:
\begin{enumerate*}
	\item $\csqlap_\ast$ is the max focus measure at $\focusindexestimate$;
	\item $\median\sparen{\csqlap_\focusindex}$ and $\mad_{\focusindex}\sparen{\csqlap_\focusindex} \coloneq \median_{\focusindex}\svbar{\csqlap_{\focusindex} - \median\sparen{\csqlap_\focusindex}}$ are the median and median absolute deviation (MAD) of the focus measure curve.
\end{enumerate*}
Thus, $\zscore$ quantifies confidence that the focus measure peak corresponds to the true depth.

We use $\zscore$ to filter out noisy depth estimates by thresholding at some level $\zthreshold$ (shown as gray pixels in depth maps), and to compute an error metric $\fractionunrecovered\in [0,1]$ equal to the \emph{fraction of unrecovered pixels} (pixels with $\zscore\paren{\pointcam} < \zthreshold$). We use $\fractionunrecovered$ as a metric to compare DFF performance, as a proxy of $\Prob\sparen{\csqnoisyif < \csqnoisyoof}$. In \cref{fig:rmsewrapfig}, we validate this metric by comparing it to the root-mean-square error (RMSE) with respect to reference depth obtained with \emph{active} DFF (\ie, using a projector to create artificial texture). The figure shows that RMSE and $\fractionunrecovered$ are strongly correlated as expected---wrong depth estimates arise from flat focus curves (low texture or high noise), hence have low $\zscore$ values. We elaborate in the supplement.

\paragraph{Contrast enhancement.} \Cref{fig:contrast-evidence} shows enhanced in-focus texture and drastically improved depth using filters of decreasing bandwidth, while adjusting exposure time to maintain exposure. Median-subtracted images of a weakly\underexposedFigure{}textured patch help visualize the contrast enhancement. The plots show focus curves and z-score statistics. \Cref{fig:illusion-teaser,fig:fig2-teaser,fig:other-results} show more scenes.

\paragraph{Exposure impact.} \Cref{fig:exposure-tradeoff,fig:underexposed} assess the tradeoff between exposure time and spectral bandwidth. In \cref{fig:exposure-tradeoff}, adding a filter of bandwidth $\bwl = \qty{10}{\nano\meter}$ requires increasing exposure time by $32\times$ to maintain proper exposure. Yet DFF performance already improves when we increase exposure time by just $4\times$ (\ie, underexposure by 3 stops), with further improvements closer to proper exposure. The crops and plots show the increase in contrast and z-scores, at a weakly textured patch---where second-order texture gradually overcomes noise---and in aggregate. The same behavior holds for other bandwidths (bottom right) and outdoors (\cref{fig:underexposed}).

\resolutionbenefitFigure

\paragraph{Pixel-adaptive aggregation.} So far we considered DFF with a fixed focus measure aggregation width $\aggkernelwidth$ (or effectively, fixed patch size $\numsamples$) at all pixels. However, the optimal aggregation width can vary per pixel, depending on local texture contrast and noise level. We use a simple adaptive method that, at each pixel, selects the \emph{minimum} aggregation width $\aggkernelwidth^\ast$ that achieves z-score above a threshold $\zthreshold$. We can then use $\aggkernelwidth^\ast$ as a metric to assess improvements in lateral resolution due to second-order texture. \Cref{fig:resolution-benefit} shows that using a spectral filter indeed results in depth maps of significantly higher lateral resolution.

\paragraph{Limitations.} The supplement shows experiments representative of the two main cases where using a spectral filter may not improve DFF performance: The first is translucent surfaces, as subsurface scattering greatly diminishes speckle contrast \citep[\S 6.4.3]{goodman2020speckle}. The second is illumination with large directional footprint (\eg, indoor lights that are very large or very close to the scene, overcast outdoor conditions, or strong indirect light), making $\scl$ much smaller than $\dflif$. We show that the deterioration of contrast, and thus DFF performance, with increasing directional bandwidth is consistent with our theory (\cref{sec:analysis}).

Moreover, improving DFF performance with a spectral filter requires increasing exposure time---even if sublinearly. \Cref{fig:mc-sim,fig:exposure-tradeoff} show that a filter bandwidth $\bwl = \qty{10}{\nano\meter}$ requires a $4\times$ exposure time increase before performance starts to improve. Our results should help assess whether different application settings justify this tradeoff between depth accuracy and acquisition time.

\galleryResultsFigure

\section{Conclusion}

We developed a second-order theory of texture, with significant implications for the definition of texture and its role in passive depth from focus. Our theory shows that:
\begin{enumerate*}
	\item Even surfaces that would traditionally be considered textureless can produce textured appearance, thanks to subjective speckle due to surface microgeometry. 
	\item Using a narrowband spectral filter enhances the in-focus contrast of this texture.
	\item This enhancement can bring about dramatic improvements in depth accuracy under conditions typical for passive computer vision.
\end{enumerate*}
We validated this theory through extensive simulations and experiments.

Our findings invite a broader reexamination of texture and passive depth sensing in computer vision: Though we developed our theory in the context of depth from focus, second-order texture can likely benefit other passive methods using focus or parallax. Further development of algorithms (confocal constancy \citep{hasinoff2009confocal} and focal flow \citep{alexander2016focal} of subjective speckle) and theory (multi-view correlations of subjective speckle \citep{fienup1988imaging} and memory effect \citep{alterman2021imaging}) can realize these benefits. Additionally, the different statistics of speckle and sensor noise hint at the possibility of using learning-based methods to detect speckle even under very low SNR. Future research can adapt related successful approaches in other speckle imaging applications \citep{metzler2020deep,xie2024wavemo}. Lastly, both our work and recent work on passive interferometric depth sensing \citep{cossairt2014digital,kotwal2023passive,chen2024coherence} rely on the weak coherence of ambient light to relax texture requirements. Further research should help understand the relative merits of interferometric and non-interferometric methods.

\section*{Acknowledgements}
We thank Dorian Chan, Aswin Sankaranarayanan, and Matthew O'Toole for helpful discussions about speckle, and Mian Wei for feedback on writing. This work was supported by NSF award 2047341, and a Sloan Research Fellowship. 

\bibliographystyle{splncs04nat_unsrt} 

\bibliography{interfpd}

\input{supplement}

\end{document}

%% file: supplement.tex

\clearpage
\setcounter{page}{1}

\newcommand*{\lightLimitationFigure}{
  \begin{figure*}[t]
    \centering
    \includegraphics[width=\linewidth]{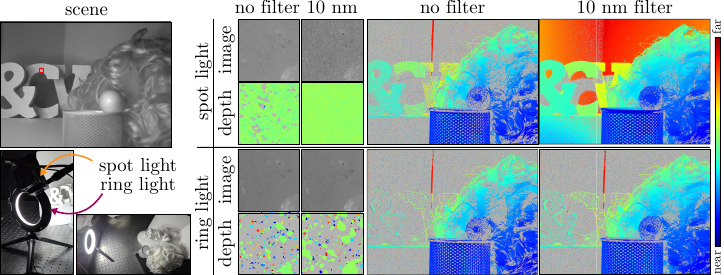}
    \caption{Using a narrowband spectral filter provides negligible improvements on this scene when we illuminate it with a nearby large ring light. The reason is the excessively small coherence area of the illumination, which results in second-order texture of very low contrast---even when spectrally filtered---relative to the noise level. By contrast, using the filter results in drastic improvements under spot-light illumination, which has a much larger coherence area.}
    \label{fig:lightlimitation}
	\vspace{1em}
\end{figure*}
}
\newcommand*{\materialLimitationFigure}{
  \begin{figure*}[t]
    \centering
    \includegraphics[width=\linewidth]{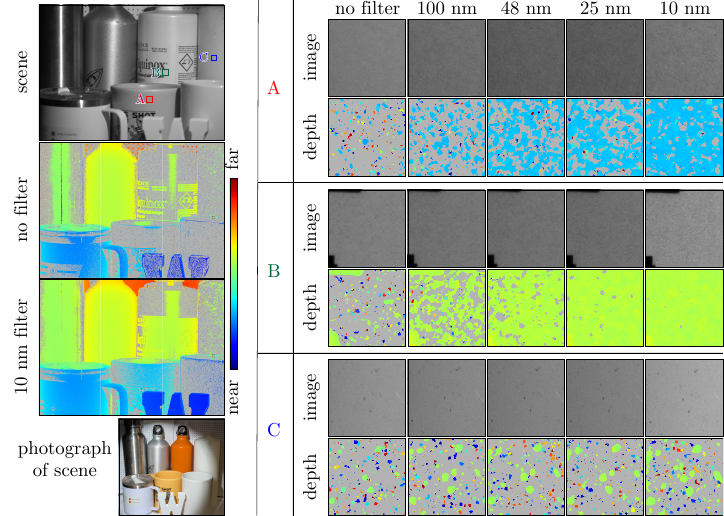}
    \caption{Performance improvements from using a narrowband filter will vary for different optical materials. Crop B, which corresponds to a nearly opaque material (label) shows the strongest improvement. Crop A, which corresponds to a weakly translucent material (ceramic cup) shows reduced but still significant improvement. Crop C, which corresponds to a strongly translucent material (wax) shows negligible improvement. The underlying cause is that strong subsurface scattering in translucent materials (\eg, wax) greatly reduces the contrast of second-order texture (subjective speckle).}
    \label{fig:materiallimitation}
\end{figure*}
}
\newcommand*{\RMSEFigure}{
  \begin{figure*}[t]
    \centering
    \includegraphics[width=\linewidth]{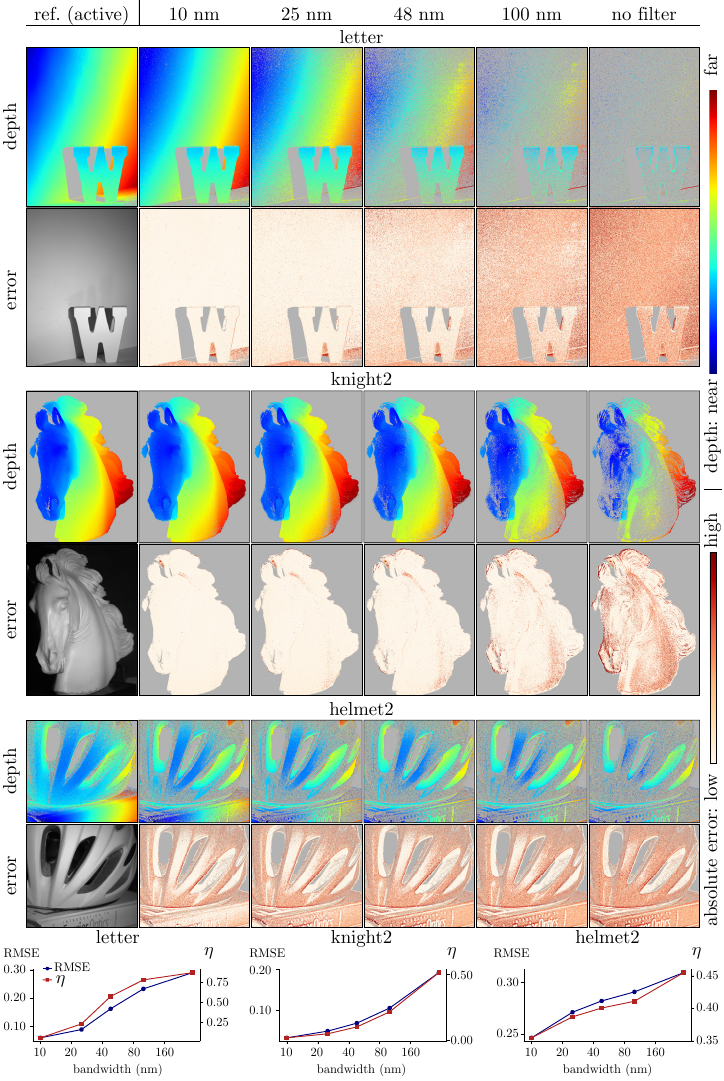}
    \caption{Validation of the $\fractionunrecovered$ metric using RMSE. (scenes captured indoors). For the reference depth map, a projector was used to illuminate the scene with a fine checkerboard pattern to create artificial texture.}
    \label{fig:rmse}
  \end{figure*}
}

\newcommand*{\illumDistanceFigure}{
	\begin{figure*}[t]
		\centering
		\includegraphics[width=\linewidth]{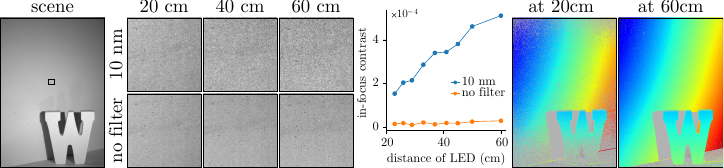}
		\caption{Impact of illumination distance on DFF performance. As the LED moves closer to the scene, the angular bandwidth increases and the coherence area decreases, leading to a reduction in in-focus contrast and DFF performance.}
		\label{fig:illumdistance}
	\end{figure*}
}

\newcommand*{\lightingsetupFigure}{
	\setlength{\columnsep}{0.5em}
	\setlength{\intextsep}{-0.15em}
	\begin{wrapfigure}[14]{r}{56.12pt}
		\centering
		\includegraphics[width=\linewidth]{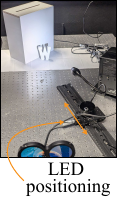}
		\caption{Capture setup with LED positioned on a translation rail.}
		\label{fig:lightingsetup}
	\end{wrapfigure}
}

\newcommand*{\thetaplotFigure}{
	\setlength{\columnsep}{0.5em}
	\setlength{\intextsep}{-0.15em}
	\begin{wrapfigure}[9]{r}{150pt}
		\centering
		\includegraphics[width=\linewidth]{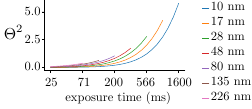}
		\vspace{-2em}
		\caption{Plot of $\contrastsnrprodmarginal^2\paren{\exptime, \bwk}$. Parameters match those in in \cref{fig:mc-sim,fig:mcsimexpanded}.}
		\label{fig:thetaplot}
	\end{wrapfigure}
}

\newcommand*{\mcsimExpandedFigure}{
	\begin{figure*}[t]
		\centering
		\includegraphics[width=\linewidth]{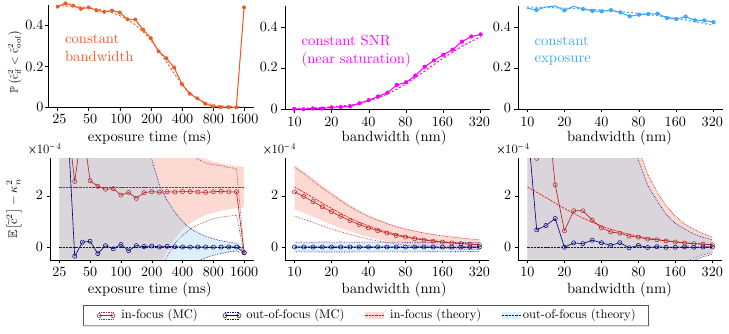}
		\caption{Visualization of the mean and standard deviation of the focus measure (specifically, $\E{\csqnoisy} - \cfvnoise$, and $\sqrt{\Var\bracket{\csqnoisy}}$) along the three paths shown in \cref{fig:mc-sim}. `theory' corresponds to direct analytical evaluation using \crefrange{eqn:fm_mean_if}{eqn:fmstdev}, while `MC' corresponds to Monte-Carlo estimates of the same quantities using samples from simulation. The MC estimates (using $10$k samples per point) have higher variance in regimes of very low exposure.}
		\label{fig:mcsimexpanded}
	\end{figure*}
}


\appendix
\renewcommand{\theHsection}{supp.\thesection}
\renewcommand{\theHsubsection}{supp.\thesubsection}
\renewcommand{\theHsubsubsection}{supp.\thesubsubsection}
\makesupptitle[\textbf{project webpage:} \projectlink]{Supplemental Document}

This document is broadly organized into two parts. \Crefrange{sec:practicalcontrast}{sec:implementation} are focused on additional results, analyses and implementation details, while \crefrange{sec:secondorder_model}{sec:recoverability_spectralfiltered} are dedicated to theoretical derivations including proofs of all of the main results. 

\addtocontents{toc}{\protect\setcounter{tocdepth}{2}}
\section*{Table of Contents}
\supptableofcontents

\clearpage

\section{Practical conditions for second-order texture}
\label{sec:practicalcontrast}
The in-focus contrast of second-order texture, and hence DFF performance, depends on practical factors spanning the imaging system, the illumination, and the surface material. The analysis and experiments presented in this section serve two purposes:
\begin{enumerate*}
	\item they validate our theoretical predictions on the nature of second-order texture in \cref{sec:analysis}; and
	\item they demonstrate limitations of our technique, \ie regimes where using a narrowband filter offers little to no benefit.
\end{enumerate*}

\subsection{Imaging: lens settings and pixel pitch}
\label{sec:lenssettings}
In \cref{sec:analysis}, we showed that the in-focus contrast of second-order texture $\cfvintensity$ is dependent on the in-focus PSF width $\dflif$ via $\numwindows$--the relative size of the in-focus PSF and the coherence area.
We characterize the impact of practical camera parameters---lens $f$-number and reproduction ratio (that together determine PSF width), and sensor pixel pitch---on the in-focus second-order texture contrast and DFF performance, and derive optimal parameter values. We excluded the impact of sensor pixel pitch in our analysis in \cref{sec:theory} for the sake of analytical tractability (by assuming flux is proportional to irradiance, and hence that $\cfvintensity=\cfvirradiance$)--the role of which will become evident below.

\paragraph{Theoretical analysis.}
We can use \cref{eqn:total_texture} to determine how the lens' $f$-number $\fnum$ and reproduction ratio $\repratio$ change second-order texture contrast, through their impact on the lens' in-focus PSF size $\dflif$. At infinity focus, the \emph{image-space} width of the diffraction-limited PSF equals $\wavelength\fnum$ \citep[\S 5.2.2]{mertz2019introduction}. We can account for finite focus with reproduction ratio $\repratio$ by replacing $f$-number $\fnum$ with the \emph{effective $f$-number} $\fnumeff \coloneq \sparen{\repratio + 1} \fnum$.\cite{kingslake1945effective} Lastly, we can convert to object-space widths by dividing by the reproduction ratio $\repratio$, arriving at:
\begin{equation}\label{eqn:if_psf}
	\dflif = \lambda\fnum \frac{\repratio + 1}{\repratio}.
\end{equation}
From \cref{pro:cfvirradiance}, assuming a first-order textureless surface:
\begin{equation}\label{eqn:lens_effects}
	\cfvintensity \propto \frac{1}{\fnum^2}\frac{\repratio^2}{\paren{\repratio + 1}^2},
\end{equation}
where throughout this section we use proportionality to absorb constants that do not depend on the lens or sensor. \Cref{eqn:lens_effects} suggests that the in-focus contrast of second-order texture improves with smaller $f$-number $\fnum$ (larger apertures) and larger reproduction ratio $\repratio$ (larger magnification). However, this analysis does not account for the finite pixel size, which imposes a bound how much we can improve in-focus contrast achieved at an optimal minimum $\fnum$ for any given $\repratio$. We elaborate next.

We denote by $\pixelpitch$ the sensor's pixel pitch (physical width of pixels on the sensor), and by $\magpixelpitch \coloneq \nicefrac{\pixelpitch}{\repratio}$ the \emph{magnified pixel pitch} (width of the surface area imaged by a pixel after accounting for magnification). Pixels optically blur incident irradiance with a $\rect$ kernel of width $\magpixelpitch$. This additional blur is an incoherent summation process that will reduce the contrast of subjective speckle when incident irradiance (in object surface-space) varies considerably within the $\magpixelpitch$-sized spatial extent of a pixel---equivalently when the in-focus PSF width $\dflif$ is significantly smaller than the magnified pixel pitch $\magpixelpitch$. These considerations suggest that, to maximize the in-focus contrast of second-order texture at reproduction ratio $\repratio$, we should set the $f$-number so that $\dflif \approx \magpixelpitch$. From \cref{eqn:if_psf,eqn:lens_effects}, the optimal in-focus texture contrast and corresponding optimal $f$-number become:
\begin{equation}\label{eqn:optimal}
	\cfvintensityopt\paren{\repratio} \propto \frac{\repratio^2}{\pixelpitch^2}\quad \text{achieved at}\quad \fnumopt\paren{\repratio} \coloneq \frac{\pixelpitch}{\lambda\paren{\repratio + 1}}. 
\end{equation}
Further decreasing the $f$-number results in reduced contrast:
\begin{equation}\label{eqn:deterioration}
	\cfvintensity = \cfvintensityopt\sparen{\repratio} \frac{\dflif^2}{\pixelpitch^2} \propto \cfvintensityopt\sparen{\repratio} \frac{\fnum^2\sparen{\repratio + 1}^2}{\pixelpitch^2},\quad \text{ if }\quad \fnum < \fnumopt\paren{\repratio}.
\end{equation} 
Therefore, \cref{eqn:optimal} shows that using sensors with smaller pixel pitch can improve in-focus contrast of second-order texture, and provides guidance on how to optimally set $f$-number $\fnum$ at a given target reproduction ratio $\repratio$.

Lastly, from \cref{pro:marginal_recoverability}, characterizing impact on DFF performance requires accounting for both texture contrast and SNR. We follow \cref{sec:recoverability} and do so by considering how to maximize the texture contrast--SNR product $\contrastsnrprodmarginal^2 = \nicefrac{\cfvintensity}{\cfvnoise}$ for a given target reproduction ratio $\repratio$. We distinguish two scenarios:
\begin{enumerate}[nosep,leftmargin=*]
	\item \emph{Exposure time is not constrained:} We achieve optimal $\contrastsnrprodmarginal^2$ by setting $f$-number $\fnum = \fnumopt\paren{\repratio}$ (\cref{eqn:optimal}) to maximize texture contrast $\cfvintensity$, and exposure time $\exptime$ to saturation (\cref{pro:snr}) to simultaneously maximize SNR $\nicefrac{1}{\cfvnoise}$.
	\item \emph{Exposure time is fixed:} Under Poisson noise-limited conditions, $\nicefrac{1}{\cfvnoise} \propto \nicefrac{1}{\fnum^2}$ (\cref{eqn:cfvnoise}). Thus $\contrastsnrprodmarginal^2 \propto \nicefrac{\repratio^2}{\fnum^4\sparen{\repratio + 1}^2}$ for $\fnum \ge \fnumopt\paren{\repratio}$ (\cref{eqn:lens_effects}), and $\contrastsnrprodmarginal^2 \propto \cfvintensityopt\sparen{\repratio} \nicefrac{\sparen{\repratio + 1}^2}{\pixelpitch^2}$ for $\fnum < \fnumopt\paren{\repratio}$ (\cref{eqn:deterioration}). Therefore, we achieve optimal $\contrastsnrprodmarginal^2$ by setting $f$-number $\fnum$ to the minimum value before saturation, collecting as much light as possible through the lens aperture.
\end{enumerate}

\newcommand*{\apertureFigure}{
	\setlength{\columnsep}{0.5em}
	\setlength{\intextsep}{-0.15em}
	\begin{wrapfigure}[15]{r}{160pt}
		\centering
		\includegraphics[width=\linewidth]{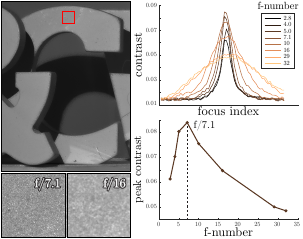}
		\vspace{-2em}
		\caption{Experimental evaluation of second-order texture contrast in a textureless region (marked, top left) as a function of $f$-number. The crops (bottom left) are representative of high and low-contrast cases.}
		\label{fig:aperture}
	\end{wrapfigure}
}
\apertureFigure{} \paragraph{Experimental validation.}
\Cref{fig:aperture} shows experimental results for an indoor scene, using filter bandwidth $\bwl=\qty{10}{\nano\meter}$ and reproduction ratio $\repratio\approx\nicefrac{1}{30}$. We captured focal stacks using different $f$-number settings, adjusting exposure time to achieve proper exposure at each setting, then visualized the focus measure $\csqnoisy_\focusindex$ at a textureless patch as a function of $f$-number. The in-focus contrast of second-order texture, both relative to out-of-focus contrast and in absolute terms, becomes worse as we use more suboptimal settings---$f$ number too large, or too small (pixel pitch limit). The results closely match our theoretical predictions in \cref{eqn:lens_effects,eqn:optimal,eqn:deterioration}.

\subsection{Illumination conditions}
In \cref{sec:analysis}, we pointed out that the in-focus contrast of second-order texture $\cfvintensity$ is dependent on the coherence area $\scl^2$ of the illumination (via $\numwindows$). Our theory therefore predicts that the in-focus contrast of second-order texture, and hence DFF performance, will be worse under \emph{wider} angular footprints of illumination, which result in smaller coherence areas.
This is the only section where we use artificial light sources instead of pre-existing ambient illumination, to facilitate a controlled experiment assessing the impact of coherence area.

\illumDistanceFigure
\lightingsetupFigure{}
\paragraph{Validation of theory.}
\Cref{fig:illumdistance} shows an experiment to validate this prediction. To emulate increasing angular bandwidth while keeping all other conditions constant, we moved an LED light source progressively closer to a scene (\cref{fig:lightingsetup}), and captured a full focal stack at each position. As a translation stage moves an LED closer to the scene, the effective angular footprint of incident illumination increases, and the coherence area decreases.  

The crops show that the in-focus contrast of second-order texture (amplified by introducing a spectral filter) is much weaker when the LED is closer to the scene (\qty{20}{\centi\metre}), compared to when the LED is far (\qty{60}{\centi\metre}). As our theory predicts, in-focus contrast decreases, and DFF performance worsens.

This experiment is complementary to the effect of decreasing \emph{spectral bandwidth} already presented in \cref{fig:contrast-evidence}. These two pieces of evidence together demonstrate that the improvements we see are indeed due to diffractive (wave-optical) effects, and therefore can be attributed to the phenomenon of second-order texture that we describe in our theory.

\paragraph{Limitation.}
This experiment also highlights a limitation of our method in practice: when the angular footprint of illumination is too wide, the texture contrast $\cfvintensity$ is still too small after spectral filtering that it cannot overcome the noise level of the camera. \Cref{fig:lightlimitation} shows experiments on a scene comprising objects of different levels of geometric complexity (in ascending order: walls, letters, meshed pen holder, hair wig). We illuminate this scene using two artificial sources of illumination, an LED spot light and an LED ring light: The spot light has a small directional bandwidth, and thus large coherence area $\scl^2$; the ring light has a much larger directional bandwidth, and thus much smaller coherence area $\scl^2$ (\cref{eqn:coherence_function_defn}). The crops show that the in-focus contrast of second-order texture is much weaker under the ring light than under the spot light. As a result, our method provides only minor performance improvements under the ring light, compared to the much more marked improvements under the spot light.

\lightLimitationFigure
\paragraph{Impact of shadows.} In many of our results, shadowed regions do not show as much improvement as non-shadowed regions (\eg the shadow of the letter in \cref{fig:lightlimitation}). One obvious reason is poor SNR in these regions. Another reason is that shadowed regions are primarily illuminated by indirect light, which typically has a much larger angular footprint than direct light, and thus much smaller coherence area. As a result, second order texture contrast is much weaker in these region, offering little to no improvement in DFF performance. 

\subsection{Material constraints}

While we considered the physics of \emph{surface scattering}, our theory does not account for possible subsurface scattering in the materials of the scene. Subsurface scattering is a phenomenon where light penetrates the surface of a material, scatters internally, and exits at a different location. This can significantly reduce the contrast of second-order texture, as it effectively results in the incoherent superposition of far more speckle patterns\footnote{due to the finite spectral bandwidth, \ie, smaller temporal coherence length than the scattering distance within the object~\citep{goodman2020speckle}} than would be present in a purely surface-scattering material.

\Cref{fig:materiallimitation} shows experiments on a scene comprising objects made from different materials, including materials with varying levels of subsurface scattering (plastic bottle and wax candle on the back right, ceramic cups on the front right). The experiment shows that the performance improvements from our method gradually decrease with increasing amount of scattering (\eg, ceramic cups, crop A), until eventually they become negligible in very translucent materials (\eg, wax candle, crop C). By contrast, our method provides strong performance improvements at parts of the scene that are largely opaque (\eg, the label on the plastic bottle, crop B). This behavior is expected, as subsurface scattering strongly reduces the contrast of subjective speckle---an effect extensively documented and analyzed in optics \citep[\S 6.4.3]{goodman2020speckle}.
\materialLimitationFigure

\section{Interplay between texture contrast and noise}
\label{sec:mcinteraction}
\thetaplotFigure{}
\Cref{sec:recoverability} provided a high-level overview of the tradeoff between second-order texture contrast and SNR for depth recovery performance in terms of the probability of error $\Pr\paren{\csqnoisyif < \csqnoisyoof}$. This section provides an additional layer of detail about what happens under the hood--showing how the texture-contrast-SNR product, and the statistics of the focus measure (mean and standard deviation, derived in \cref{sec:dffanalysis}) are impacted by spectral bandwidth and exposure time, and how that, in turn, results in the behaviour of the probability of error shown in \cref{fig:mc-sim}.

\paragraph{Texture Contrast-SNR product.} 
Just like we visualized the probability of error as a grid as in \cref{fig:mc-sim}, we can also visualize the texture contrast-SNR product $\contrastsnrprodmarginal^2 = \nicefrac{\cfvintensity}{\cfvnoise}$ defined in \cref{sec:recoverability} directly. An increase in $\contrastsnrprodmarginal^2$ directly maps to a decrease in the probability of error through \cref{lem:proberror}. \Cref{fig:thetaplot} shows $\contrastsnrprodmarginal^2$ as a function of exposure time $\exptime$ for different values of spectral bandwidth. Each line stops at the maximum exposure time $\exptime$ that can be used for a given spectral bandwidth, before the camera saturates. 

\paragraph{Focus measure statistics.}
\Cref{lem:marginal_fmstats} (from \cref{sec:dffanalysis}) provides approximate expressions for the mean and variance of the sample contrast focus measure $\csqnoisy$. We have the mean and standard deviation of the in-focus and out-of-focus focus measures:
\begin{align}
	\E{\csqnoisyif} &\approx \cfvintensity + \cfvnoise, \label{eqn:fm_mean_if}\\
	\E{\csqnoisyoof} &\approx \cfvnoise,\label{eqn:fm_mean_oof} \\
	\begin{split}
	\sqrt{\Var\bracket{\csqnoisyif}} &\approx \sqrt{\frac{2}{\numsamples-1}} \paren{\cfvintensity + \cfvnoise}, \\
	\sqrt{\Var\bracket{\csqnoisyoof}} &\approx \sqrt{\frac{2}{\numsamples-1}} \paren{\cfvnoise}.
	\end{split}\label{eqn:fmstdev}
\end{align}
As a consequence, the mean \emph{separation} between the (noisy) in-focus and out-of-focus sample contrast focus measures
\begin{align}
	\E{\csqnoisyif} - \E{\csqnoisyoof} &\approx \cfvintensity\label{eqn:mean_fm_separation}
\end{align}

In addition, we know from \cref{pro:cfvirradiance} that for a first-order textureless surface ($\cfvfoirradiance=0$)
\begin{align}
	\cfvintensity &= \frac{1}{\numspectralbuckets\numwindows}
\end{align}
 and from \cref{eqn:cfvnoise}
 \begin{align}
	\cfvnoise &=  \frac{1}{\exptime\photoncurrentperflux\marginalmeanflux}\paren{1 + \frac{\sigmaread^2}{\exptime\photoncurrentperflux\marginalmeanflux}}
\end{align}

Thus, given the parameters of the capture system, we can plug in these values to compute the mean and standard deviation of the in-focus and out-of-focus focus measures. The bottom row in \Cref{fig:mcsimexpanded} plots these quantities as an extension of \cref{fig:mc-sim}. The `theory' curves show the predicted separation $\E{\csqnoisy_{\focusindex}}-\cfvnoise$ for $\focusindex=\focusindexif$ and $\focusindex=\focusindexoof$ (the latter is uniformly zero, as evident from \cref{eqn:fm_mean_oof}), also labeling a region $\pm 1\times$ standard deviations surrounding it. These regions give us a sense of the width of the distributions of $\csqnoisyif$ and $\csqnoisyoof$.
\mcsimExpandedFigure{}
The \emph{separation} between the in-focus and out-of-focus focus measure distributions provides immediate insight into DFF performance. When the distributions of $\csqnoisyif$ and $\csqnoisyoof$ are well-separated, the chances of a patch being misclassified as in-focus or out-of-focus is low, and hence $\Pr\paren{\csqnoisyif < \csqnoisyoof}$ is low. When they are close together, the opposite is true (worst case: 50-50 chance). 

This plots provide a clearer picture of how capture parameters (exposure time and spectral bandwidth) individually influence the statistics of the focus measures in and out of focus with more granularity.
\begin{enumerate}
	\item 
	If the spectral bandwidth is reduced at a fixed exposure time, the mean separation between the in-focus and out-of-focus focus measures increases in accordance with \cref{eqn:mean_fm_separation} (right sections). 
	\item When the exposure time is increased (left section), the mean separation is unaffected, but the standard deviation of the focus measures decreases, reducing overlap between the distributions and thus reducing the probability of error. 
	\item If the spectral bandwidth is reduced, while also increasing the exposure time to maintain proper exposure (middle section), the mean separation increases, while the in-focus standard deviation increases slightly (in accordance with \cref{eqn:fmstdev}).
\end{enumerate}

\paragraph{Match with Monte-Carlo simulation.} Overlaid on the theory-based curves and shaded regions (from \crefrange{eqn:fmstdev}{eqn:mean_fm_separation}) are the corresponding Monte-Carlo (MC) estimates from simulation. The close match between theory and simulation (for both focus measure statistics, and probability of error) shows that our theoretical analysis is accurate, and that the assumptions made in \cref{sec:dffanalysis} are reasonable. In particular, we assumed that the distributions of $\csqnoisyif$ and $\csqnoisyoof$ are approximately Gaussian, which is validated by the close match between theory and simulation.
{One may notice that the theoretical curves consistently overestimate the mean contrast in the high-contrast regimes (left and middle plots). This can be directly attributed to an assumption we make in the derivation of \cref{lem:marginal_fmstats}, \ie we assume that the intensities of pixels are independent. See \cref{sec:dffanalysis} for details.}

\RMSEFigure
\section{Depth recovery performance evaluation}
\label{sec:evaluation}

We report quantitative metrics across all our scenes, and validate our z-score-based evaluation metric. Visit our webpage (\projectlink) for interactive visualizations of all scenes in the main paper and supplement, including depth maps, surface renders, and comparisons.

\subsection{Quantitative comparison}
\Cref{tab:fracempty_allscenes} aggregates DFF performance metrics across all captured scenes we show in both the main paper and the supplement. As we do not have ground-truth depth, to assess performance, we use the fraction of unrecovered pixels $\fractionunrecovered$ we defined in \cref{sec:experiments}, which serves as a measurable proxy for the probability of error $\Prob\sparen{\csqnoisyif < \csqnoisyoof}$ (thus lower values of $\fractionunrecovered$ represent better DFF performance). The table shows that, consistently across all scenes, performance improves as spectral bandwidth decreases---and while adjusting exposure time to maintain proper exposure---in agreement with our theoretical predictions. The main paper shows quantitative comparisons also for cases of underexposure (\cref{fig:exposure-tradeoff}).

\begin{table}[t]
	\caption{Fraction of unrecovered pixels $\fractionunrecovered$ with $\zscore > \zthreshold \paren{=4.0}$ for all scenes in our results (main paper and supplement). Dashes indicate unavailable (not captured) datasets. Visit \projectlink  for interactive visualizations of all scenes}
	\vspace{-1em}
	\label{tab:fracempty_allscenes}
	\centering
\begin{filecontents*}{data/quant_allscenes.csv}
scene,no filter,100 nm,48 nm,25 nm,10 nm
illusion,0.834,-,-,-,0.004
side table,0.560,0.469,-,-,0.243
helmet,0.742,-,-,0.461,0.313
knight,0.866,0.532,0.427,0.388,0.362
set cards,0.842,0.732,0.621,0.460,0.287
bin,0.853,-,-,0.461,0.583
sculpture,0.552,-,-,-,0.113
corner,0.715,0.645,0.615,0.613,0.644
columns,0.823,0.664,0.668,0.631,0.642
trash can,0.633,0.493,0.501,0.462,0.216
hair \& mesh,0.561,0.423,0.359,0.261,0.141
bottles \& mugs,0.534,0.415,0.371,0.348,0.295
cup,0.893,-,-,-,0.518
\end{filecontents*}
\csvreader[
  tabular=R{1.2in}C{0.5in}C{0.5in}C{0.5in}C{0.5in}C{0.5in},
  table head= \toprule scene & no filter & $\qty{100}{\nano\metre}$  & $\qty{48}{\nano\metre}$ & $\qty{25}{\nano\metre}$  & $\qty{10}{\nano\metre}$ \\\midrule,
  late after line=\\,
  late after last line=\\\bottomrule
]
{data/quant_allscenes.csv}{}
{\textbf{\csvcoli} & \csvcolii & \csvcoliii & \csvcoliv & \csvcolv & \csvcolvi}%
\end{table}
\clearpage
\subsection{Validation of z-score-based metric with active DFF reference}
\Cref{fig:rmse} shows comparisons with a reference from \emph{active} DFF---a projector illuminates the scene with a high-frequency pattern to create artificial texture. We compute RMSE relative to this reference depth, excluding pixels with low-confidence reference. 

We observe: 
\begin{enumerate*}
	\item RMSE decreases as spectral bandwidth narrows, validating our theory.
	\item RMSE is strongly correlated with our $\fractionunrecovered$ metric (fraction of unrecovered pixels).
\end{enumerate*}
 This strong correlation is expected: From DFF theory \citep{nayar1994shape,subbarao1998selecting}, wrong depth estimates with high z-score are \emph{extremely unlikely} (a patch that, when out of focus, serendipitously has far higher noise than at other focus settings). Essentially all wrong depth estimates arise from flat focus-measure curves (low texture or high noise), and hence have low z-score. Therefore, our $\fractionunrecovered$ metric quantifies DFF performance as accurately as RMSE, while being much more practical for outdoor experiments. Capturing active DFF for reference is challenging under sunlight. 

\section{Implementation details}
\label{sec:implementation}

We provide details for simulations and experiments in the main paper and supplement.

\newcommand*{\hardwareFigure}{
	\setlength{\columnsep}{0.5em}
	\setlength{\intextsep}{-0.15em}
	\begin{wrapfigure}[10]{r}{153.14pt}
		\centering
		\includegraphics[width=\linewidth]{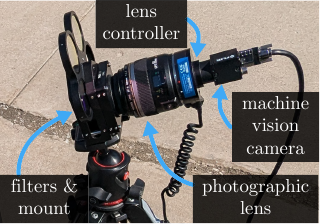}
		\vspace{-2em}
		\caption{Capture setup.}
		\label{fig:hardware}
	\end{wrapfigure}
}

\subsection{Capture setup}

\hardwareFigure{}Our setup comprises only off-the-shelf components for standard depth from focus. We use a machine vision camera (FLIR BlackFly S BFS-U3-122S6M-C) with a monochromatic sensor (Sony IMX304, \qty{1.4}{\centi\meter} diagonal, \qty{3.5}{\micro\meter} pixel pitch, 12 bits). The camera provides access to RAW measurements. We equip the camera with an electronic lens mount (ISSI Canon EF lens controller) for programmatic focus and aperture control, and a photographic lens (Canon EF-S 60mm f/2.8 Macro USM) for imaging. We use standard filter rings and an optical filter wheel to mount and easily rotate filters at the front of the lens. Lastly, we use narrowband spectral filters of \qty{2}{inch} diameter (Edmund Optics, different FWHM bandwidths, center wavelength $\approx \qty{530}{\nano\metre}$). \Cref{fig:hardware} shows a photograph, and \cref{tab:filters} lists all filters we use in our experiments.

\begin{table}[t]
	\centering
	\caption{Spectral filters we use for experiments.}\label{tab:filters}
	\vspace{-1em}
	\setlength{\tabcolsep}{6pt}
	\begin{tabular}{ccc}
		\toprule
		filter model & center wavelength (nm) & full-width half-max (FWHM) (nm) \\
		\midrule
		EO \#65-216 & 532 & 10 \\
		EO \#84-115 & 534 & 25 \\
		EO \#67-045 & 534.5 & 48 \\
		EO \#33-331 & 550 & 100 \\
		\bottomrule
	\end{tabular}
\end{table}

\subsection{Depth recovery algorithms} 

\Cref{alg:singlestepDFF,alg:pixeladaptiveDFF} detail the methods we use in \cref{sec:experiments} for recovering depth from a focal stack. \Cref{alg:singlestepDFF} performs one aggregation over a patch of fixed size to compute a Laplacian-based focus measure. \Cref{alg:pixeladaptiveDFF} is a pixel-adaptive variant: At each pixel, it performs multiple consecutive aggregations---each progressively increasing the effective patch size---until the robust z-score at that pixel reaches a certain threshold. Thus this algorithm varies per-pixel patch size to adapt to areas of different texture contrast. In both algorithms, the routine \textsc{InterpDepth} uses three-point Gaussian interpolation of the focus measure to determine a peak dithered between focus indices \citep{nayar1994shape}.

\begin{algorithm}[t]
	\caption{Depth recovery using single-step aggregation.}\label{alg:singlestepDFF}
	\begin{algorithmic}[1]
		\Procedure{SingleStepDFF}{$\curly{\intensitynoisy_{\focusindex}}_{j=1}^{\numfocusindex}$}
		\For{$\focusindex \in \curly{1,\dots,\numfocusindex}$}
			\State $\csqnoisy_{\focusindex} \gets \Call{LocalFM}{\intensitynoisy_{\focusindex}}$\Comment{focus measure stack}
			\State $\csqnoisy_{\focusindex} \gets {{\csqnoisy_{\focusindex}} \convolve \gaussaggkernel_{\aggkernelwidth}}$ \Comment{Gaussian kernel aggregation}
		\EndFor
		\State $\focusindexestimate\sparen{\pointcam} \gets \argmax_{\focusindex} \csqlap_{\focusindex}\paren{\pointcam}$
		\State $\depth \gets \Call{InterpDepth}{\csqnoisy_{\cdot}, \focusindexestimate}$
		\State $\zscore \gets \Call{Zscore}{\csqnoisy_{\cdot}, \focusindexestimate}$\Comment{pixel-wise z-scores}
		\State \Return $\depth, \zscore$
		\EndProcedure
		\Statex
		\Procedure{LocalFM}{$\intensitynoisy_{\focusindex}$}
		\State \Return $\frac{\bracket{\intensitynoisy_{\focusindex} \convolve \lapkernel}^2}{\bracket{\intensitynoisy_{\focusindex} \convolve \meankernel}^2}$ \Comment{normalized squared Laplacian}
		\EndProcedure
		\Statex
		\Procedure{Zscore}{$\csqnoisy_{\cdot}, \focusindexestimate$}
			\State $\csqnoisy_{\mathrm{med}}\paren{\pointcam} = \median_\focusindex \csqnoisy_{\focusindex}\paren{\pointcam}$
			\State $\mathrm{MAD}\paren{\pointcam} = \median_\focusindex \abs{\csqnoisy_{\focusindex}\paren{\pointcam} - \csqnoisy_{\mathrm{med}}\paren{\pointcam}}$
			\State \Return $\frac{\csqnoisy_{\focusindexestimate} - \csqnoisy_{\mathrm{med}}}{\mathrm{MAD}}$ \Comment{pixel-wise robust z-score}
		\EndProcedure
	\end{algorithmic}
\end{algorithm}

\begin{algorithm}[t]
	\caption{Pixel-adaptive depth recovery using progressive aggregation.}\label{alg:pixeladaptiveDFF}
	\begin{algorithmic}[1]
	\Procedure{PixelAdaptiveDFF}{$\curly{\intensitynoisy_{\focusindex}}_{j=1}^{\numfocusindex}, \zthreshold $}
	\State $\zscore\paren{\pointcam} \gets 1$; $\depth\paren{\pointcam} \gets \mathrm{NaN}$; $\complete\paren{\pointcam} \gets \mathrm{false}$   \Comment{init. z-scores, depth, completion}
	\State $\aggstepnumber \gets 0$
	\For{$\focusindex \in \curly{1,\dots,\numfocusindex}$}
		\State $\csqnoisy_{\focusindex} \gets \Call{LocalFM}{\intensitynoisy_{\focusindex}}$\Comment{init. focus measure stack}
	\EndFor
	\While{$\aggstepnumber < \aggstepmax$ and $\lnot\mathrm{all}\paren{\complete}$}
		\For{$\focusindex\in \curly{1,\dots,\numfocusindex}$}
			\State $\csqnoisy_{\focusindex}\gets \Call{WeightedAggStep}{\csqnoisy_{\focusindex}, \zscore}$ 
		\EndFor
		\State $\focusindexestimate\sparen{\pointcam} \gets \argmax_{\focusindex} \csqlap_{\focusindex}\paren{\pointcam}$
		\State $\zscore \gets \Call{Zscore}{\csqnoisy_{\cdot},\focusindexestimate}$\Comment{update pixel-wise z-scores}
		\State $\completenew \gets \zscore > \zthreshold \land \lnot \complete$
		\State $\depth\bracket{\completenew} \gets \Call{InterpDepth}{\csqnoisy_{\cdot}, \focusindexestimate}\bracket{{\completenew}}$
		\State $\complete  \gets \complete \lor \completenew$
		\State $\aggstepnumber \gets \aggstepnumber + 1$
	\EndWhile
	\State \Return $\depth$
	\EndProcedure
	\Statex
	\Procedure{WeightedAggStep}{$\csqnoisy_{\focusindex}, \zscore$}
	\State \Return $\frac{\bracket{\paren{\csqnoisy_{\focusindex} \cdot \zscore} \convolve \gaussaggkernel_{\aggstepkernelwidth}}}{\bracket{\zscore \convolve \gaussaggkernel_{\aggstepkernelwidth}}}$ \Comment{z-score weighted aggregation}
	\EndProcedure
	\end{algorithmic}
\end{algorithm}

\subsection{Monte Carlo simulation}\label{sec:mcstudy_supp}

Our Monte Carlo simulation is a direct implementation of the description of second-order texture in \cref{sec:cfvirradiance_proof} and the full noise model in \cref{sec:noisemodel}. Below, we explain the sequence of steps alongside implementation details and references to corresponding equations in the supplement and main text. We use a pseudorandom number generator for all sampling steps. \Cref{alg:mcsampler} summarizes the full Monte Carlo pipeline.
\begin{enumerate}[nosep,leftmargin=*]
	\item We simulate a surface that is fronto-parallel to the camera. We model its texture using the blurred outgoing radiance $\radianceoutgblur{\scl}\paren{\gridcenters}$ (\cref{eqn:radianceoutgblur_defn}) as a two-dimensional grid of spacing $\scl$, where $\curly{\gridcenters}$ is the set of grid centers (based on \cref{pro:secondorder_statistical_model} in \cref{sec:cfvirradiance_proof}).
	\item We consider first-order textureless appearance, \ie, $\foradianceoutgblur{\scl}$ is constant across grid locations, or $\cfvforadianceoutgblur{\scl}=\cfvfoirradiance=0$. Additionally, we assume that $\foradianceoutgblur{\scl}$ at each location is proportional to spectral bandwidth $\bwk$ according to \cref{eqn:effectiveatten_approx}, to model brightness attenuation due to narrowing bandwidth; thus: $\foradianceoutgblur{\scl} = \sparen{\foradianceoutgblur{\scl}}_{\mathrm{orig}} \peakfilteratten \bwk$, where $\sparen{\foradianceoutgblur{\scl}}_{\mathrm{orig}}$ is the original value of $\foradianceoutgblur{\scl}$ without spectral filtering.
	\item We sample each cell of the coherence area grid independently from a Gamma distribution according to \cref{pro:secondorder_statistical_model}.
	We report related parameters in \cref{tab:textureparams}.
	\item We compute the irradiance reaching the sensor at the in-focus setting according to the discrete convolution of the PSF $\psf$ with the outgoing radiance grid $\radianceoutgblur{\scl}\paren{\gridcenters}$ in \cref{eqn:irradiance_total_approx}. We model the PSF as a discrete Gaussian kernel with standard deviation $\nicefrac{\dfl}{2}$ as in \cref{eqn:gaussian_psf}. We compute the in-focus PSF width in object space using \cref{eqn:if_psf}. We scale the irradiance at the sensor by a factor $\propto \nicefrac{\repratio^2}{\fnumeff^2}$ to model how the energy reaching the sensor through the lens varies as a function of $\repratio$ and $\fnum$. We report related parameters in \cref{tab:imagingparams}.
	\item We sum the resulting irradiance grid $\irradiance\paren{\pointcam}$---now mapped to image space---over the pixel area $\pixelresponsewidth \times \pixelresponsewidth$ to get the flux $\flux\paren{\pointcam}$ at each pixel at the camera's resolution.
	\item Given the flux at each pixel, we use the full noise model of \cref{eqn:noisemodel_full} to sample the noisy intensity $\intensitynoisy\paren{\pointcam}$ at each pixel. We report related parameters in \cref{tab:sensorparams}.
	\item Given the noisy intensity measurements, we compute the focus measure $\csqnoisy$ (\cref{sec:background}) for a square-shaped patch of size $\sqrt{\numsamples} \times 
	\sqrt{\numsamples}$ ($5\times 5$) centered at the pixel of interest.
	\item We repeat this process multiple times to obtain Monte-Carlo estimates of $\Prob\paren{\csqnoisyif > \csqnoisyoof}$, 
	 for different values of $\bwk$ and $\exptime$. We plot these estimates in \cref{fig:mc-sim}.
\end{enumerate}

\begin{algorithm}[t]
\caption{Monte Carlo focus measure sampler}\label{alg:mcsampler}
\begin{algorithmic}[1]
	\Statex default parameters $\gainvar, \sigmapreamp, \sigmapostamp$ \etc from \crefrange{tab:textureparams}{tab:sensorparams}
	\Statex Pixel resolution $H\times W$.
\Procedure{MCFMSampler}{$\bwk, \exptime$}
	\State $\foradianceoutgblur{\scl} \gets \peakfilteratten \bwk \cdot \foradianceoutgblur{\scl, \mathrm{orig}}$ \Comment{\cref{eqn:effectiveattenfactor_approx}}
	\State $\radianceoutgblur{\scl}\paren{\gridcenters} \gets \Gam\paren{\numspectralbuckets, \frac{\numspectralbuckets}{\foradianceoutgblur{\scl}} }$ \Comment{\cref{eqn:radianceoutcohgrid_conditional_gamma}}
	\State $\irradiance_{\mathrm{if}} \gets \Call{PSFBlur}{\radianceoutgblur{\scl}}$ 
	\State $\intensitynoisy_{\mathrm{if}}\paren{\pointcam} \gets \Call{SensorReadout}{\irradiance_{\mathrm{if}}, \exptime}$ 
	\State $\csqnoisyif\paren{\pointcam} \gets \Call{SampleContrastFM}{\intensitynoisy_{\mathrm{if}}}$ \Comment{in-focus sample}
	\State $\csqnoisyoof\paren{\pointcam} \gets \Call{SampleContrastFM}{\marginalmeanintensity \cdot \text{ones}(H,W)}$ \Comment{out of focus sample}
	\State \Return $\csqnoisyif, \csqnoisyoof$ 
\EndProcedure
\Statex
\Procedure{PSFBlur}{$\radianceoutgblur{\scl}$}
	\State $\irradiance\paren{\pointcamalt} \gets \sum_{\gridindex}\scl^2 \psf\paren{\pointcamalt - \gridcenters} \radianceoutgblur{\scl}\paren{\gridcenters} $ \Comment{\cref{eqn:irradiance_total_approx}}
	\State \Return $\irradiance$
\EndProcedure
\Statex
\Procedure{SensorReadout}{$\irradiance, \exptime$}
\State $\flux\paren{\pointcam} \gets \irradiance\paren{\pointcam} \cdot \pixelresponsewidth^2$ \Comment{sample irradiance coarsely at pixel resolution}
\State $\photoncount \gets \Poisson{\paren{\photoncurrentperflux \flux + \darkcurrent}\exptime}$
\State $\readnoisepreamp \gets \Normal\paren{0, \sigmapreamp^2}$; $\readnoisepostamp \gets \Normal\paren{0, \sigmapostamp^2}$
\State $\intensitynoisy\paren{\pointcam} \gets \min\paren{ \left\lfloor\gain (\min\paren{\photoncount,\fullwellcapacity} + \readnoisepreamp) + \readnoisepostamp \right\rfloor, \adcmax}$\Comment{\cref{eqn:noisemodel_full}}
\State \Return $\intensitynoisy $
\EndProcedure
\Statex
\Procedure{SampleContrastFM}{$\intensitynoisy$}
\State \Return $\frac{\fmvaluenoisy}{\meanintnoisy^2}$\Comment{\cref{eqn:sample_stats}}
\EndProcedure
\end{algorithmic}
\end{algorithm}

\begin{table}[t]
	\centering
	\caption{Parameters for Monte Carlo simulation in \cref{fig:mc-sim}.}
	\vspace{-2em}
	\begin{subtable}{0.6\textwidth}
		\centering
		\caption{Second-order texture parameters.}
		\vspace{-1em}
		\label{tab:textureparams}
		\begin{tabular}{ccc}
			\toprule
			parameter & value & description \\
			\midrule
			$\stdheight$ & $\qty{3}{\micro\metre}$ & RMS height of surface \\
			$\scl$ & $\qty{12}{\micro\metre}$ & width of coherence cell grid \\
			$\bar{\wavelength}$ & \qty{532}{\nano\metre} & central wavelength\\
			\bottomrule
		\end{tabular}
	\end{subtable}\begin{subtable}{0.4\textwidth}
		\centering
		\caption{Camera lens settings.}
		\vspace{-1em}
		\label{tab:imagingparams}
		\begin{tabular}{ccc}
			\toprule
			parameter & value & description \\
			\midrule
			$\repratio$ & 0.01 & reproduction ratio\\
			$ \fnum$ & 7 & $f$-number\\
			\bottomrule
		\end{tabular}
	\end{subtable}\\
	\begin{subtable}{0.48\textwidth}
		\centering
		\caption{Camera sensor parameters.}
		\vspace{-1em}
		\label{tab:sensorparams}
		\begin{tabular}{ccc}
			\toprule
			parameter & value & description \\
			\midrule
			$\gainvar$ &  $\qty{10}{\electron\per\digitalnumber}$ & sensor gain \\
			$\sigmapreamp $ & $\qty{10}{\electron}$& pre-amp noise variance \\
			$\sigmapostamp $ & $\qty{15}{\digitalnumber}$ & post-amp noise variance \\
			$\log_2\paren{\adcmax}$& 12 & ADC bit depth\\
			$\quantumefficiency$ & 0.4 & quantum efficiency\\
			$\darkcurrent$ & $\qty{1.0}{\electron\per\second}$& dark current\\
			$\fullwellcapacity$ & $\qty{35000}{\electron}$ & full-well capacity\\
			$\pixelresponsewidth$ & $\qty{3.45}{\micro\metre}$ & pixel pitch\\
			\bottomrule
		\end{tabular}
	\end{subtable}
\end{table}


\clearpage
\section{Appearance model for second-order texture}
\label{sec:secondorder_model}
We present a derivation of \cref{pro:irradiance} consistent with physical-optics principles for an explicit micro-geometry of the surface, mapping wave quantities to conventional radiometric quantities used in computer vision.
We then explain how traditional BRDF-based image formation models used in computer vision are a first-order abstraction of appearance based on the statistics of the surface heightfield, and then quantify the strength of second-order texture in \cref{sec:cfvirradiance_proof}.
Our formulations largely follow \citet[Sections 5.8 \& 6.3]{goodman2020speckle} and \citet{steinberg2022rendering}.

\subsection{Incident illumination model}
\label{sec:incident_illumination}

Consider the 2D plane macroscopically tangent to the surface of the object at the point of interest with macroscopic outward normal $\normaldir$ (\cref{fig:setup}). At any point $\pointtwod \in \R^2$ in this plane, we model the incoming illumination  as consisting of a superposition of monochromatic plane waves of varying direction and wavelength:
\begin{align}\label{eqn:incident_planewave_sum}
	\incwave\paren{\pointtwod}
	&= \int_{\wavenumber \in \R^+}\!\int_{\unithemi\paren{\normaldir}}\!\!\! \planewaveamp\paren{\dirvec, \wavenumber} e^{\imi \wavenumber \dirvec_{xy} \cdot \pointtwod} \ud\sigma\paren{\dirvec} \ud \wavenumber
\end{align}
Here, $\dirvec$ is the outward unit vector antiparallel to the direction of the incoming plane wave, $\wavenumber = \nicefrac{2\pi}{\lambda}$ represents its (angular) wavenumber, and 
\begin{align}
\unithemi\paren{\normaldir} = \curly{\dirvec\in\R^3, \norm{\dirvec}=1, \dirvec\cdot \normaldir>0}
\end{align}
is the upper-half space of valid outward incident direction vectors.
$\dirvec_{xy}$ is the component of $\dirvec$ parallel to the plane, \ie, $\dirvec = \bracket{\dirvec_{xy},\ \dirz\normaldir}$.

In the context of computer vision with direct and indirect illumination, one may consider these plane waves to have been generated by a variety of light sources in the scene.
The amplitude of these plane waves $\paren{\dirvec, \wavenumber}$ are modeled to be a result of the random process $\{\planewaveamp: \unithemi\paren{\normaldir}\times \R^+ \to \Complex\}$ (Units:$\bracket{\unit{{\watt}^{0.5}\per\steradian}}$). Any observable radiometric quantity such as irradiance or intensity is to be computed by averaging the result over this ensemble. \citep{wolf2007introduction}

\paragraph{Mutual incoherence assumption.}
We assume these plane waves to be \emph{mutually incoherent}, \ie $\planewaveamp\paren{\dirvec, \wavenumber}$ are uncorrelated for any non-identical pair $\paren{\dirvec, \wavenumber}$, that is,
\begin{align}\label{eqn:mutualincoherence_statement}
	\angled{\planewaveamp\paren{\dirvec_1, \wavenumber_1}\planewaveamp^\ast\paren{ \dirvec_2, \wavenumber_2}}
	&= \foreshortenedradianceinc\paren{ \dirvec_1, \wavenumber_1}\delta_{\wavenumber}\paren{\wavenumber_1 - \wavenumber_2}\delta_{\dirvec}\paren{\dirvec_1 - \dirvec_2}
\end{align}
Here, $\angled{\cdot}$ represents an average\footnote{We use the symbol $\angled{\cdot}$ to distinguish it from expectations over texture and heightfields, for which we reserve $\E{\cdot}$ as used in the main text.} over the ensemble $\planewaveamp$, and $\foreshortenedradianceinc\paren{ \dirvec_1, \wavenumber_1}$ equals the \emph{foreshortened radiance}
\begin{align}\label{eqn:foreshortenedradiance}
	\foreshortenedradianceinc\paren{\dirvec, \wavenumber}= \radiancein\paren{\dirvec, \wavenumber} \paren{\dirvec \cdot \normaldir}.
\end{align}
$\radiancein\paren{\dirvec, \wavenumber}$ (Units:$\bracket{\unit{\watt\per\metre\per\steradian}}$) is the incoming spectral radiance environment map introduced in \cref{sec:theory}. The assumption of mutual incoherence allows us make a correspondence between plane wave amplitudes with incoming radiance as defined in radiometry.
In addition, we assume the environment map is separable in $\wavenumber$ and $\dirvec$, \ie $\radiancein\paren{\dirvec, \wavenumber} = \spectruminc\paren{\wavenumber}\cdot\radiancein\paren{\dirvec}$, where the normalized spectral profile is $\spectruminc\paren{\wavenumber}$, which integrates to unity.

The above is a suitable model for real-world ambient illumination, where plane waves from different incident directions due to a variety of far-field light sources (\eg sun, lamps, ceiling lamps, indirect reflections) and wavenumbers are statistically uncorrelated due to randomness of the emission process at each point on the light source(s).
By considering the plane-wave amplitudes $\planewaveamp$ to be position-independent, we have thus described an environment lighting model. This is sufficient for our purposes since we only seek to describe the scattering interaction with the surface in the local neighborhood at the point of interest.

We re-express the complex incident field created by the superposition of these plane waves (\cref{eqn:incident_planewave_sum}) as
\begin{align}
	\incwave\paren{\pointtwod} 
	&= \int_{\wavenumber} \incwave\paren{\pointtwod, \wavenumber}\ud \wavenumber \\
	\incwave\paren{\pointtwod, \wavenumber}
	&= \int_{\unithemi\paren{\normaldir}} \planewaveamp\paren{\dirvec, \wavenumber} e^{\imi \wavenumber \dirvec_{xy} \cdot \pointtwod} \ud\sigma\paren{\dirvec} \label{eqn:incwave_k}
\end{align}

\paragraph{Example: incident irradiance.}
We now provide a simple example of computing an observable radiometric quantity using the above illumination model.
The incident \emph{irradiance} (Units:$\sbracket{\unit{\watt\per\metre\squared}}$) at a point on the local 2D plane is given by the average magnitude-squared of the incident field:\citep{testorf2010phase}
\begin{align}
	\irradianceinc\paren{\pointtwod} &=\angled{\abs{\incwave\paren{\pointtwod}}^2} \\
	&= \int_{\R^+} \angled{\abs{\incwave\paren{\pointtwod, \wavenumber}}^2} \ud \wavenumber\\
	&= \int_{\R^+}
	\irradianceinc\paren{\pointtwod, \wavenumber}  \ud \wavenumber.
\end{align}
The cross-terms vanish since $\planewaveamp\paren{\cdot, \cdot}$ is uncorrelated for non-identical wavenumbers.
We now similarly expand the \emph{spectral density} of incident irradiance $\irradianceinc\paren{\pointtwod, \wavenumber}$ for wavenumber $\wavenumber$, as (Units:$\bracket{\unit{\watt\per\metre}}$)
\begin{align}\label{eqn:irradianceinc}
	\irradianceinc\paren{\pointtwod, \wavenumber}  
	&= \angled{\abs{\incwave\paren{\pointtwod, \wavenumber}}^2}\\ 
	&= \int_{\unithemi\paren{\normaldir}}\!\int_{\unithemi\paren{\normaldir}}\!\!\! 
	\angled{\planewaveamp\paren{\dirvec_1, \wavenumber} \planewaveamp^\ast\paren{\dirvec_2, \wavenumber}} e^{\imi \wavenumber \pointtwod \cdot \paren{\dirvec_{xy,1} - \dirvec_{xy,2}}}\ud\sigma\paren{\dirvec_1} \ud\sigma\paren{\dirvec_2}\\
	&= \spectruminc\paren{\wavenumber}\int_{\unithemi\paren{\normaldir}}\radiancein\paren{\dirvec} \paren{\dirvec \cdot \normaldir}\ud \sigma\paren{\dirvec} 
\end{align}
which is independent of position $\pointtwod$, as expected, since we use an environment map. Hence, we can drop the position dependence and use $\irradianceinc\paren{\wavenumber}$ and $\irradianceinc$ instead.
In summary, thanks to the assumption of mutual incoherence of the incident plane waves, the above example illustrates that for the computation of observable radiometric quantities, we can consider the effect of each incident plane wave $\paren{\dirvec, \wavenumber}$ independently of others (using $\radiancein\paren{\dirvec, \wavenumber}$) and then sum up their contributions linearly.

\subsection{Coherence function of the incident illumination}
\label{sec:coherencefn}
Before proceeding to compute the scattered field and irradiance on the sensor of the camera, we introduce the following definition (\cref{eqn:coherence_function_defn} in main text):

\begin{dfn}[label={dfn:coherencefn}]{Coherence Function}{coherencefunction}
We  define the \emph{coherence function} $\mcfint\paren{\cdot}$ of the illumination environment map  $\radiancein\paren{\cdot}$ as its projected Fourier transform:
	\begin{align}
	\mcfint\paren{\pointtwod; \wavenumber}
&\coloneq \int_{\unithemi\paren{\normaldir}} \radiancein\paren{\dirvec}\paren{\dirvec\cdot \normaldir} 
	e^{\imi \wavenumber \dirvec_{xy} \cdot {\pointtwod}} \ud \sigma\paren{\dirvec}\\
	 \radiancein\paren{\dirvec}
&= \frac{\wavenumber^2}{4\pi^2}\int_{\R^2}\mcfint\paren{\pointtwod; \wavenumber}
	e^{-\imi \wavenumber \dirvec_{xy} \cdot {\pointtwod}} \ud{\pointtwod},\quad \dirvec\in\unithemi\paren{\normaldir} \label{eqn:coherencefn_inverse}
\end{align}
\end{dfn}
It is easily shown using the results of the previous section that the coherence function represents the two-point correlation of the incident field $\incwave\paren{\cdot, \wavenumber}$, where the Fourier variable $\pointtwod$ represents the \emph{separation} between these two points \footnote{Readers with a background in optics may recognize this as a form of the van-Cittert Zernike theorem.}
\begin{align}
	\mcfint\paren{\pointtwod_1 - \pointtwod_2, \wavenumber}&= \angled{\incwave\paren{\pointtwod_1, \wavenumber}\incwave^\ast\paren{\pointtwod_2, \wavenumber}}
\end{align}
Observe that by definition, $\mcfint\paren{\vv{0},\wavenumber} = \nicefrac{\irradianceinc\paren{\wavenumber}}{\spectruminc\paren{\wavenumber}}$.

\paragraph{Example: Gaussian environment lighting.}
As a concrete example, consider the following lighting configuration\footnote{The projected Gaussian here is technically truncated at $\norm{\dirvec_{xy}}<1$. We consider the width $\lightingwidth$ to be sufficiently small such that the truncation effect is negligible.}
\begin{align}\label{eqn:gaussian_lighting}
	\radiancein\paren{\dirvec} &= \radiancepeak \exp\paren{-\frac{\norm{\dirvec_{xy} - \dirvecpeak_{xy}}^2}{2\lightingwidth^2}}
\end{align}
The angular distribution of radiance is a Gaussian lobe (when projected onto the x-y plane) centered at $\dirvecpeak_{xy}$ with (projected) angular spread $2\lightingwidth$ and peak value $\radiancepeak$

The corresponding coherence function (Fourier dual of lighting) according to \cref{dfn:coherencefn} is given by
\begin{align}\label{eqn:gaussian_mcfint}
	\mcfint\paren{\pointtwod; \wavenumber} &= 2\pi{\lightingwidth^2} \radiancepeak\exp\paren{-\frac{\lightingwidth^2 \wavenumber^2}{2}\norm{\pointtwod}^2}\exp\paren{\imi \wavenumber \dirvecpeak_{xy} \cdot \pointtwod}\\
	&= \frac{\irradianceinc\paren{\wavenumber}}{\spectruminc\paren{\wavenumber}} \exp\paren{-\frac{\norm{\pointtwod}^2}{2\scl^2}}\exp\paren{\imi \wavenumber \dirvecpeak_{xy} \cdot \pointtwod}
\end{align}
where $\scl = \nicefrac{1}{\wavenumber\lightingwidth}$ is the \emph{spatial coherence length}, a measure of how quickly the magnitude of field correlation (coherence) decreases with separation, and $\irradianceinc\paren{\wavenumber} = 2\pi{\lightingwidth^2} \radiancepeak \spectruminc\sparen{\wavenumber} $ is the spectral density of incident irradiance at the surface.
We will use the above example in \cref{sec:cfvirradiance_proof} to prove \cref{pro:cfvirradiance} from the main text.

\subsection{Scattering and Measurement}
Our goal now is to express the irradiance measured at a point on the camera sensor given the incident environment map $\radiancein\paren{\dirvec}\spectruminc\paren{\wavenumber}$ and the surface heightfield $\height\paren{\cdot}$, which together (along with the camera lens) determine the scattered field reaching the camera sensor.
For a given point on the camera sensor, consider $\pointcam$ as the point on the tangent plane it is mapped to via pinhole projection (center of the lens aperture). The viewing angle of the camera for point $\pointcam$ is $\dirvecout$ (refer \cref{fig:setup}).

\paragraph{Roughness model.}
We characterize the roughness of the random heightfield $\height$ using the \emph{standard deviation} $\stdheight \coloneqq \sqrt{\var\paren{\height\paren{\vv{x}}}}$ and \emph{correlation length} $\window$ ($\corr\paren{\height\paren{\vv{x}},\height\paren{\vv{y}}} \approx 0$ if $\norm{\vv{x}-\vv{y}} > \window$). We consider
\begin{enumerate*}
	\item  $\stdheight$ comparable to wavelength, 
	\item  $\window$ smaller than the spatial coherence length $\scl$.
\end{enumerate*}
The same conditions underlie standard BRDF models for optically rough surfaces~\citep{beckmann1987scattering,levin2013fabricating,stam1999diffraction}. Thus in practice, they are satisfied by any surface not optically smooth (i.e., near-specular) at visible wavelengths.
The surface roughness conditions are also spelt out in Goodman~\citep[\S 5.10]{goodman2020speckle}

\paragraph{Scattering integral.}
For a single incident plane wave ${\planewaveamp\paren{\dirvec, \wavenumber} e^{\imi \wavenumber \dirvec \cdot \pointthreed}}$, the scattered wave reaching the camera sensor is given by
\begin{align}
	\senswave\paren{\pointcam; \wavenumber, \dirvec} &= \planewaveamp\paren{\dirvec, \wavenumber} \cdot \csfscatfourier\paren{\pointcam; \wavenumber\paren{\dirvecout + \dirvec} }
\end{align}
where we have defined the \emph{scattering integral}~\citep{goodman2020speckle,steinberg2026wavetracing}
\begin{align}\label{eqn:scattering_integral}
	\csfscatfourier\paren{\pointcam; \wavenumber,\hdirvec } &= \int_{\R^2} \csf\paren{\pointcam - \pointtwod}
	\scatter\paren{\pointtwod; \wavenumber, \hdirz}
	e^{\imi \wavenumber \hdirvec_{xy}\cdot {\pointtwod}}  \ud {\pointtwod}
\end{align}
where $\csf\paren{\cdot}$ is the \emph{coherent spread function} (CSF) of the camera describing the complex field impulse response of the camera lens at the depth of the surface (assumed fronto-parallel to the camera), and $\scatter\paren{\pointtwod; \wavenumber, \hdirz} = \reflectcoeff\paren{\pointtwod,\wavenumber}e^{\imi\wavenumber {\hdirz}\height\paren{\pointtwod}} $ as introduced in \cref{dfn:fs_brdf}. The reflection coefficient $\reflectcoeff\paren{\pointtwod,\wavenumber}$ is a slowly varying (compared to $\height$) non-negative factor that models the spatially-varying spectral albedo of the surface.
 Note that $\scatter\paren{\cdot; \wavenumber, \hdirz}$ is a random signal through its dependence on the heightfield $\height\paren{\cdot}$.
This therefore results in the total field
\begin{align}
	\senswave\paren{\pointcam} &= \int_{\R^+}\int_{\unithemi\paren{\normaldir}} \planewaveamp\paren{\dirvec, \wavenumber} \cdot \csfscatfourier\paren{\pointcam; \wavenumber\paren{\dirvecout + \dirvec} } \ud \sigma\paren{\dirvec} \ud \wavenumber
\end{align}
\paragraph{Measured irradiance.}
The irradiance reaching the sensor is the average mean-squared of the field reaching the sensor, \ie
\begin{align}
	\sensirradiance\paren{\pointcam} &= \angled{\abs{\senswave\paren{\pointcam}}^2}\\
	&=  \int_{\wavenumber \in\R^+}\int_{\dirvec\in\unithemi\paren{\normaldir}}	\spectruminc\paren{\wavenumber}\radiancein\paren{\dirvec} \paren{\dirvec \cdot \normaldir} \abs{\csfscatfourier\paren{\pointcam; \wavenumber,\paren{\dirvecout + \dirvec} }}^2 \ud \sigma\paren{\dirvec} \ud\wavenumber
\end{align}
following a simplification similar to the example of incident irradiance in \cref{sec:incident_illumination} based on the mutual incoherence assumption.


To obtain the irradiance \emph{measured} by the camera sensor, we must weight the spectral integration by the sensor spectral sensitivity function $\ssf\paren{\wavenumber}$:
\begin{align}
	\irradiance\paren{\pointcam} &= \int_{\R^+}\ssf\paren{\wavenumber} \spectruminc\paren{\wavenumber}\irradiance\paren{\pointcam,  \wavenumber} \ud \wavenumber
\end{align}
defining
\begin{align}
	\irradiance\paren{\pointcam,  \wavenumber} &= \int_{\unithemi\paren{\normaldir}} 
		\radiancein\paren{\dirvec} \paren{\dirvec \cdot \normaldir} \abs{\csfscatfourier\paren{\pointcam; \wavenumber,{\dirvecout + \dirvec} }}^2 \ud \sigma\paren{\dirvec} 
	\end{align}

We therefore have the following proposition:
\begin{prp}[label={pro:irradiance_exact}]{Measured Sensor Irradiance (Exact)}{irradiance}
	\begin{align}\label{eqn:irradiance_spectrum_integration}
	\irradiance\paren{\pointcam} 
	&= \int_{\R^+}\ssf\paren{\wavenumber} \spectruminc\paren{\wavenumber}\irradiance\paren{\pointcam, \wavenumber}\ud \wavenumber\\
	\irradiance\paren{\pointcam,\wavenumber}
	&= \int_{\unithemi\paren{\normaldir}} 
	\radiancein\paren{\dirvec} \paren{\dirvec \cdot \normaldir} \abs{\csfscatfourier\paren{\pointcam; \wavenumber,{\dirvecout + \dirvec} }}^2 \ud \sigma\paren{\dirvec}  \label{eqn:irradiance_spectrum_exact}
\end{align}
\end{prp}

\subsection{Approximating measured irradiance}
To arrive at \cref{pro:irradiance} in the form of a simple linear system, we now make two simplifying approximations.

\paragraph{Approximation 1: Mean z-component approximation.}
We now make a standard implicit approximation used in similar derivations \citep{stam1999diffraction,levin2013fabricating}.
Consider the scattering integral \cref{eqn:scattering_integral}. The phase term $\scatter\paren{\pointtwod; \wavenumber \hdirz}= e^{\imi\wavenumber {\hdirz}\height\paren{\pointtwod}}$ varies with incident direction $\dirvec$, since $\hdirvec = \dirvec + \dirvecout$ in \cref{eqn:irradiance_spectrum_exact}, preventing an interpretation of $\csfscatfourier$ as a Fourier transform. 

If we consider an illumination environment map similar to \cref{eqn:gaussian_lighting} where the incident illumination is relatively concentrated around a peak direction $\dirvecpeak$ with z-component $\dirpeakz$, then the variation in the z-component of the incident direction $\dirz$ across the environment map is small, \ie $\dirz \approx \dirpeakz$ for all $\dirvec$ with significant contribution.
Formally, the approximation is stated as:
\begin{align}\label{eqn:meanz_approx}
{{ \paren{\height\paren{\pointtwod_1} - \height\paren{\pointtwod_2}}}} 	{\Delta_{\dirz}}\ll{\lambda}\quad \forall \pointtwod_1, \pointtwod_2
\end{align}
This gives us the following approximate form for the scattering integral:
\begin{align}
	\csfscatfourier\paren{\pointcam; \wavenumber\hdirvec } &\approx 
	\int_{\R^2} \csf\paren{\pointcam - \pointtwod}
	\scatter\paren{\pointtwod; \wavenumber, \meanhdirz}
	e^{\imi \wavenumber \hdirvec_{xy}\cdot {\pointtwod}}  \ud {\pointtwod}
\end{align}
using $\scatter\paren{\pointtwod; \wavenumber, \meanhdirz} = e^{\imi\wavenumber \meanhdirz \height\paren{\pointtwod}}$
with the mean direction z-component $\meanhdirz = \dirpeakz + \diroutz$. 

Using this approximate form, we expand the squared magnitude term in \cref{eqn:irradiance_spectrum_exact} as
\begin{align}
	\begin{split}
		\abs{\csfscatfourier\paren{\pointcam; \wavenumber\hdirvec }}^2
		\approx &\int_{\R^2}\int_{\R^2}  \csf\paren{\pointcam - \pointtwod_1}\csf^\ast\paren{\pointcam - \pointtwod_2} 
		\scatter\paren{\pointtwod_1; \wavenumber, \meanhdirz} \scatter^\ast\paren{\pointtwod_2; \wavenumber, \meanhdirz}
		\\
		&e^{\imi \wavenumber \hdirvec_{xy} \cdot \paren{\pointtwod_1 - \pointtwod_2}}
		\ud\paren{\pointtwod_1}
		\ud\paren{\pointtwod_2}
	\end{split} 
\end{align}
Using a change of variables
\begin{align}
	\begin{split}
	\midpt &= \frac{\pointtwod_1  + \pointtwod_2}{2}\\
	\diffpt &= \frac{\pointtwod_1 - \pointtwod_2}{2} \label{eqn:changeofvar}
	\end{split}
\end{align}
we have
\begin{align}
	\begin{split}
		\abs{\csfscatfourier\paren{\pointcam; \wavenumber\hdirvec }}^2
		\approx& 4\int_{\R^2}
		 \int_{\R^2}  \csf\paren{\pointcam - \midpt - \diffpt}\csf^\ast\paren{\pointcam - \midpt + \diffpt} \\
		& \scatter\paren{\midpt + \diffpt; \wavenumber, \meanhdirz} 
		\scatter^\ast\paren{\midpt - \diffpt; \wavenumber, \meanhdirz} 		\\
		&e^{\imi \wavenumber \hdirvec_{xy} \cdot \paren{2\diffpt}}
		\ud{\diffpt} 
		\ud{\midpt}
	\end{split}
\end{align}
Using $\pstf\paren{\pointcamalt, \pointcamdiff; \wavenumber, \meanhdirz}$ from  \cref{dfn:fs_brdf},
\begin{align}
	\pstf\paren{\midpt, \diffpt; \wavenumber, \meanhdirz} \coloneq  \scatter\paren{\midpt + \diffpt; \wavenumber, \meanhdirz} 
	\scatter^\ast\paren{\midpt - \diffpt; \wavenumber, \meanhdirz} 
\end{align}

and plugging the above into \cref{eqn:irradiance_spectrum_exact}, we have an approximate expression
\begin{align}
	\begin{split}
		\irradiance\paren{\pointcam,\wavenumber}
		\approx &4\int_{\R^2}\int_{\R^2}
		\csf\paren{\pointcam - \midpt - \diffpt}\csf^\ast\paren{\pointcam - \midpt + \diffpt} \\
		& \pstf\paren{\midpt, \diffpt; \wavenumber \meanhdirz}\\
		& \Big[ \int_{\unithemi\paren{\normaldir}} 
		\radiancein\paren{\dirvec} \paren{\dirvec \cdot \normaldir} 
		e^{\imi \wavenumber \dirvec_{xy} \cdot \paren{2\diffpt}}
		\ud \sigma\paren{\dirvec} \Big]\\
		&e^{\imi \wavenumber \dirvecout_{xy} \cdot \paren{2\diffpt}}
		\ud{\diffpt} 
		\ud{\midpt}
	\end{split}
\end{align}
Identifying the term in square brackets as the coherence function $\mcfint\paren{2\diffpt; \wavenumber}$ from \cref{dfn:coherencefunction}, we have
\begin{align}\label{eqn:irradiance_approx1}
	\begin{split}
		\irradiance\paren{\pointcam,\wavenumber}
		\approx &4\int_{\R^2}\int_{\R^2}
		\csf\paren{\pointcam - \midpt - \diffpt}\csf^\ast\paren{\pointcam - \midpt + \diffpt} \\
		& \pstf\paren{\midpt, \diffpt; \wavenumber \meanhdirz}	 
		\mcfint\paren{2\diffpt; \wavenumber}
		e^{\imi \wavenumber \dirvecout_{xy} \cdot \paren{2\diffpt}}
		\ud{\diffpt} 
		\ud{\midpt}
	\end{split}
\end{align}

At this point, we bring to attention the definitions of $\radianceout$ from \cref{dfn:fs_brdf} 
\begin{equation}
	\radianceout\paren{\pointcamalt, \dirvecout, \wavenumber} \coloneq \int_{\unithemi\paren{\normaldir}}\fsrf\paren{\pointcamalt, \dirvec+\dirvecout, \wavenumber} \radiancein\paren{\dirvec} \paren{\dirvec\cdot \normaldir} \ud \sigma\paren{\dirvec}.\tag{rep. \ref{eqn:fs_reflectance}}
\end{equation}
and \cref{dfn:fs_brdf}
\begin{align}
	\fsrf\paren{\pointcamalt, \hdirvec, \wavenumber} 
		\coloneq
		\frac{1}{\brdfconst} \int_{\R^2} \pstf\paren{\pointcamalt, \pointcamdiff, \wavenumber, \hdirz}
		e^{\imi \wavenumber {\hdirvec_{xy}}\cdot 2\pointcamdiff}\ud \pointcamdiff,\tag{rep. \ref{eqn:fs_brdf}}
\end{align}
where $\brdfconst$, with dimensions of area, is a constant inversely proportional to the amount of light collection by the lens aperture, scaling as $\brdfconst \propto \nicefrac{\fnumeff^2}{4\repratio^2}$ (the proportionality absorbs the area scale set by the CSF, which cancels in all downstream results). The effective (sensor-side) f-number $\fnumeff = \fnum\paren{1 +\repratio}$, $\fnum$ being the f-number of the camera (refer \cref{sec:practicalcontrast}) 

With the definition of the coherence function in \cref{dfn:coherencefn}, and using the same mean-z-component approximation already used for the scattering integral $\csfscatfourier$, we can write $\radianceout\paren{\midpt, \dirvecout, \wavenumber}$ approximately as
\begin{align}
	\radianceout\paren{\pointcamalt, \dirvecout, \wavenumber}
		&\approx\frac{1}{\brdfconst} \int_{\R^2} 
		\pstf\paren{\midpt, \diffpt; \wavenumber, \meanhdirz}	 
		\mcfint\paren{2\diffpt; \wavenumber}
		e^{\imi \wavenumber \dirvecout_{xy} \cdot \paren{2\pointcamdiff}}
		\ud{\pointcamdiff}
		\label{eqn:radianceout_ft}
\end{align}

\paragraph{Approximation 2: Weak-coherence approximation.}
For typical ambient-lighting conditions, the coherence function (\eg \cref{eqn:gaussian_mcfint}) will have small width $\scl = \nicefrac{1}{\wavenumber\lightingwidth}$.
As long as $\scl$ is small relative to the width of the CSF $\csf\paren{\cdot}$, we can approximate the CSF to be constant within the area of the support of $\mcfint$, resulting in the expression (from \cref{eqn:irradiance_approx1} and \cref{eqn:radianceout_ft}):

\begin{align}\label{eqn:irradiance_spectral_approx}
	\irradiance\paren{\pointcam, \wavenumber} &\approx
		\int_{\R^2}
		\psf\paren{\pointcam - \midpt}
		\radianceout\paren{\midpt, \dirvecout, \wavenumber}
		\ud{\midpt}
\end{align}
where the normalized PSF (integrates to unity) of the camera
$\psf\paren{\pointtwod} \coloneq \nicefrac{\fnumeff^2}{\repratio^2} \abs{\csf\paren{\pointtwod}}^2$.

Combining the above with \cref{eqn:irradiance_spectrum_integration}, we have

\begin{align}
	\irradiance\paren{\pointcam} &\approx 
		\int_{\R^+}\ssf\paren{\wavenumber} \spectruminc\paren{\wavenumber}
		\int_{\R^2}
		\psf\paren{\pointcam - \midpt}
		\radianceout\paren{\midpt, \dirvecout, \wavenumber}
		\ud \midpt
		\ud \wavenumber
\end{align}
thus proving \cref{pro:irradiance}.


\subsection{Ensemble averaging}
\label{sec:ensembleavgproof}

The BRDF defined in \citet{stam1999diffraction} can be equivalently stated using our notation as the (normalized) power spectral density of the scattering function $\scatter$:
\begin{align}
	\brdf\paren{\pointcamalt, \hdirvec, \wavenumber} &= \frac{1}{4\brdfconst}\, \psdscatter\paren{\wavenumber\hdirvec_{xy}}, \\
	\psdscatter\paren{\vv{\kappa}} &\coloneq \lim_{\surfarea\to\infty} \frac{1}{\surfarea}\Exp{\height}{\abs{\int_{\surfarea} \scatter\paren{\pointtwod; \wavenumber, \hdirz} e^{\imi \vv{\kappa} \cdot \pointtwod} \ud\pointtwod}^2},
\end{align}
where $\surfarea$ is the illuminated surface area and $\psdscatter$ is the power spectral density of $\scatter$ (the Fourier transform of its spatial autocorrelation, by the Wiener--Khinchin theorem). The $\nicefrac{1}{\surfarea}$ normalization makes the BRDF independent of the illuminated area (a per-unit-area surface property).
For this section, we are considering a surface where the BRDF is constant (as indicated by the expression).
The BRDF is proportional to the expected value of the squared magnitude of the absolute-squared fourier transform of $\scatter\paren{\pointtwod; \wavenumber, \hdirz}$ over an ensemble of surface heightfield realizations. It is \emph{almost} the mean of the absolute-square of the scattering integral $\csfscatfourier$ in \cref{eqn:scattering_integral}, except for the camera lens CSF term $\csf\paren{\cdot}$. 

The expression for irradiance in our model depends on the particular realization of the surface heightfield $\height\paren{\cdot}$ through the \emph{sample BRDF} $\fsrf$. We first prove \cref{pro:ensemble}---that the ensemble mean of the sample BRDF is the traditional BRDF---and then use the exact model to show how the finite-aperture CSF relates the two.

\paragraph{Proof of \cref{pro:ensemble}.}
 If we compute the mean of our sample BRDF $\fsrf\paren{\pointcamalt, \hdirvec, \wavenumber}$ from \cref{dfn:fs_brdf} over an ensemble of surface heightfield realizations, we have
\begin{align}
	\Exp{\height}{\fsrf\paren{\pointcamalt, \hdirvec, \wavenumber}}
	&= \frac{1}{\brdfconst} \int_{\R^2} \Exp{\height}{\pstf\paren{\pointcamalt, \pointcamdiff, \wavenumber, \hdirz}} e^{\imi \wavenumber \hdirvec_{xy}\cdot 2\pointcamdiff}\ud \pointcamdiff \\
	&= \frac{1}{\brdfconst} \int_{\R^2} \Exp{\height}{\scatter\paren{\pointcamalt + \pointcamdiff; \wavenumber, \hdirz}\scatter^\ast\paren{\pointcamalt - \pointcamdiff; \wavenumber, \hdirz}} e^{\imi \wavenumber \hdirvec_{xy}\cdot 2\pointcamdiff}\ud \pointcamdiff
\end{align}
This step requires only \emph{wide-sense stationarity} of $\scatter$ (equivalently, of the heightfield $\height$): the correlation $\Exp{\height}{\scatter\paren{\pointcamalt + \pointcamdiff}\scatter^\ast\paren{\pointcamalt - \pointcamdiff}}$ is then the autocorrelation of $\scatter$ at separation $2\pointcamdiff$, independent of the center $\pointcamalt$.\footnote{We adopt the stronger \emph{ergodicity} assumption in \cref{pro:ensemble}, as it is the conventional condition invoked for optically rough surfaces~\citep{goodman2020speckle}; it implies wide-sense stationarity, which is all the argument here uses.} Substituting $\pointtwod = 2\pointcamdiff$ (Jacobian $\ud \pointcamdiff = \nicefrac{1}{4}\,\ud \pointtwod$),
\begin{align}
	\Exp{\height}{\fsrf\paren{\pointcamalt, \hdirvec, \wavenumber}}
	&= \frac{1}{4\brdfconst} \int_{\R^2} \Exp{\height}{\scatter\paren{\pointtwod' + \pointtwod; \wavenumber, \hdirz}\scatter^\ast\paren{\pointtwod'; \wavenumber, \hdirz}} e^{\imi \wavenumber \hdirvec_{xy}\cdot \pointtwod}\ud \pointtwod \\
	&= \frac{1}{4\brdfconst}\,\psdscatter\paren{\wavenumber\hdirvec_{xy}},
\end{align}
identifying the Fourier transform of the autocorrelation with the power spectral density $\psdscatter$ (Wiener--Khinchin). Comparing with the definition of $\brdf$, we obtain (\cref{pro:ensemble})
\begin{align}
	\brdf\paren{\pointcamalt, \dirvec + \dirvecout, \wavenumber} = \Exp{\height}{\fsrf\paren{\pointcamalt, \dirvec + \dirvecout, \wavenumber}}.
\end{align}

\paragraph{The traditional BRDF as the pinhole limit.}
If we consider the expected value of the irradiance in \cref{eqn:irradiance_spectrum_exact} from the exact model (\cref{pro:irradiance_exact}), we obtain
\begin{align}
	\Exp{\height}{\irradiance\paren{\pointcam,\wavenumber}}
	&= \int_{\unithemi\paren{\normaldir}} 
	\radiancein\paren{\dirvec} \paren{\dirvec \cdot \normaldir} \Exp{\height}{\abs{\csfscatfourier\paren{\pointcam; \wavenumber,{\dirvecout + \dirvec} }}^2 }\ud \sigma\paren{\dirvec}
\end{align}
\begin{align}
	\Exp{\height}{\abs{\csfscatfourier\paren{\pointcam; \wavenumber,{\dirvecout + \dirvec} }}^2 } &= \Exp{\height}{\abs{\int_{\R^2} \csf\paren{\pointcam - \pointtwod}
	\scatter\paren{\pointtwod; \wavenumber, \hdirz}
	e^{\imi \wavenumber \hdirvec_{xy}\cdot {\pointtwod}}  \ud {\pointtwod}}^2}
\end{align}
If we consider the case of a pinhole camera, then $\csf\paren{\cdot}$ has infinite extent and is constant across the surface, so it factors out of $\csfscatfourier$, leaving the plain Fourier transform of $\scatter$ that defines the traditional BRDF:
\begin{align}
	\Exp{\height}{\abs{\csfscatfourier\paren{\pointcam; \wavenumber,{\dirvecout + \dirvec} }}^2 }
	\propto \Exp{\height}{\abs{\int_{\surfarea} \scatter\paren{\pointtwod; \wavenumber, \hdirz}e^{\imi \wavenumber \hdirvec_{xy} \cdot \pointtwod} \ud\pointtwod}^2}
	\propto \psdscatter\paren{\wavenumber\hdirvec_{xy}} \propto \brdf\paren{\pointcamalt, \hdirvec, \wavenumber},
\end{align}
recovering the traditional BRDF-based rendering equation. The two agree \emph{only} in this pinhole limit: the traditional BRDF omits the camera's coherent spread function $\csf$, whereas our exact model retains it. For any finite aperture the CSF remains inside $\csfscatfourier$ and our model departs from the traditional BRDF---precisely the regime in which defocus and second-order texture (subjective speckle) arise. Our model thus \emph{generalizes} the traditional BRDF to finite-aperture imaging of unresolved microgeometry.

\section{Characterization of second-order texture}
\label{sec:cfvirradiance_proof}
In this section, we use the appearance model derived in the previous section to prove \cref{pro:cfvirradiance} in the main text, characterizing the distribution of irradiance through $\cfvirradiance$.


\cref{eqn:irradiance_spectral_approx}
can be approximated as:
\begin{align}\label{eqn:irradiance_gridsum}
	\irradiance\paren{\pointcam, \wavenumber} 
	&\approx 
	\sum_{\gridindex} \irradiancegrid,\quad \text{where} \\
	\label{eqn:irradiancegrid_defn}
	\irradiancegrid\paren{\wavenumber} 
	&\coloneq \gridspacing^2 \psf\paren{\pointcam - \gridcenters} \radianceoutgblur{\gridspacing}\paren{\gridcenters,\wavenumber}, \\
	\label{eqn:radianceoutgblur_defn}
	\radianceoutgblur{\gridspacing}\paren{\pointcamalt,\wavenumber}
	&\coloneq \int_{\R^2} \normgausskernelgrid\paren{\pointcamaltalt - \pointcamalt} \radianceout\paren{\pointcamaltalt, \dirvecout, \wavenumber}  \ud \pointcamaltalt
\end{align}
defining $\radianceoutgblur{\gridspacing}\paren{\cdot,\wavenumber}$ as the (blurred) local average of outgoing radiance $\radianceout\paren{\cdot, \dirvecout, \wavenumber}$ using a normalized Gaussian kernel in $\R^2$ of width $\gridspacing$ centered at $\gridcenters$
\begin{align}
	\normgausskernelgrid\paren{\pointcamaltalt-\gridcenters} 
	&\coloneq \frac{2}{\pi \gridspacing^2} \exp\paren{-2\frac{\norm{\pointcamaltalt-\gridcenters}^2}{\gridspacing^2}},
\end{align}
with $\curly{\gridcenters}$ forming a rectangular grid of spacing $\gridspacing$ in $\R^2$. 
In the limit of $\gridspacing \to 0$, \cref{eqn:irradiance_gridsum} becomes an exact equality with \cref{eqn:irradiance_spectral_approx}. This approximation is suitable when $\gridspacing$ is smaller than the width of the PSF $\dfl$.

Our objective now is to model the distribution of $\irradiance\paren{\pointcam, \wavenumber}$.
To do this, we consider the approximation above using the specific grid spacing of $\gridspacing = \scl$.
In \cref{sec:expdistproof}, we show that considering the Gaussian environment lighting model in \crefrange{eqn:gaussian_lighting}{eqn:gaussian_mcfint} with spatial coherence length $\scl$, $\radianceoutgblur{\scl}\paren{\pointcamalt,\wavenumber}$ can be modeled as an exponential random variable:
\begin{align}
	\radianceoutgblur{\scl}\paren{\pointcamalt,\wavenumber} \mid \foradianceoutgblur{\scl}\paren{\pointcamalt, \wavenumber} &\sim \Expo\paren{\nicefrac{1}{\foradianceoutgblur{\scl}\paren{\pointcamalt, \wavenumber}}}, \label{eqn:exponential_irradiancegrid}
\end{align}
with mean equal to its first-order (ensemble-averaged) counterpart, $\foradianceoutgblur{\scl}\paren{\cdot,\wavenumber} = \Exp{\height}{\radianceoutgblur{\scl}\paren{\cdot,\wavenumber}}$.
More explicitly in terms of the (first-order) BRDF:
\begin{align}
\foradianceoutgblur{\scl}\paren{\pointcamalt,\wavenumber}&= 
\int_{\R^2} \normgausskernelgrid\paren{\pointtwod - \pointcamalt}
\int_{\unithemi\paren{\normaldir}}\!\!\! 
\Exp{\height}{\fsrf\paren{\pointtwod, \dirvec, \dirvecout, \wavenumber}}\radiancein\paren{\dirvec} \paren{\dirvec\cdot \normaldir} 
\ud\sigma\paren{\dirvec}
\ud \pointtwod
\end{align}
We can also consider $\radianceoutgblur{\scl}\paren{\pointcamalt_{i},\wavenumber}$ and  $\radianceoutgblur{\scl}\paren{\pointcamalt_{j},\wavenumber}$ are independent for $i\ne j$ since they correspond to (mostly) non-overlapping microstructure regions on the surface (and the correlation length of the heightfield is smaller than $\scl$).
We can now approximate \cref{eqn:irradiance_exitance} as:
\begin{align}
	\label{eqn:irradiance_total_approx}
	\irradiance\paren{\pointcam}
	&\approx
	\sum_{\gridindex}\scl^2 \psf\paren{\pointcam - \gridcenters} \radianceoutgblur{\scl}\paren{\gridcenters},\quad \text{where} \\
	\label{eqn:radianceout_wnintegral}
	\radianceoutgblur{\scl}\paren{\pointcamalt}
	&\coloneq \int_{\R^+}\!\! \radianceoutgblur{\scl}\paren{\pointcamalt,\wavenumber} \ssf\paren{\wavenumber} \spectruminc\paren{\wavenumber} \ud \wavenumber
\end{align}
While $\radianceoutgblur{\scl}\paren{\gridcenters,\wavenumber}$ are independent\footnote{conditioned on their first order means, see \cref{sec:numareas_proof}} across the grid locations $\gridcenters$, they are \emph{not} independent across wavenumbers $\wavenumber$ for a fixed $\gridindex$, \ie $\radianceoutgblur{\scl}\paren{\gridcenters,\wavenumber_1}$ and $\radianceoutgblur{\scl}\paren{\gridcenters,\wavenumber_2}$ are not independent as they are both directly dependent on the local surface structure in the neighborhood of $\gridcenters$.
Considering two wavenumbers $\wavenumber_1$, $\wavenumber_2$ they may be correlated or uncorrelated depending on the distribution from which the surface heightfield $\height$ is sampled.

Consider that the normalized spectral envelope of the light reaching the camera 
\begin{align}\label{eqn:spectruminitial}
	\spectruminitial\paren{\wavenumber} \coloneq \frac{\ssf\paren{\wavenumber} \spectruminc\paren{ \wavenumber}}{\int_{\R^+} \ssf\paren{\wavenumber^\prime} \spectruminc\paren{\wavenumber^\prime} \ud\wavenumber^\prime} 
\end{align}
has a gaussian form
\begin{align}\label{eqn:spectruminitial_gaussian}
	\spectruminitial\paren{\wavenumber} &= \frac{1}{\sqrt{2\pi}\bwk} \exp\paren{-\frac{\paren{\wavenumber - \mwk}^2}{2\bwk^2}}
\end{align}
with bandwidth $\bwk$ and mean wavenumber $\mwk$.
\citet{goodman2020speckle} show that for a Gaussian-distributed heightfield, Gaussian correlation function, and a Gaussian spectral profile, the number of uncorrelated spectral buckets for a given wavenumber bandwidth $\bwk$ is given by 
\begin{align}\label{eqn:numspectralbuckets}
 \numspectralbuckets &= \sqrt{1 + 8\pi^2 \paren{\frac{\bwk}{\mwk}}^2\paren{\frac{\stdheight}{\mwl}}^2} \approx G_1{\bwk}
\end{align}
where $G_1 = 2\sqrt{2}\pi \sparen{\nicefrac{\stdheight}{\mwl\mwk}}$,
thereby allowing us to approximately model the integral over wavenumber in \cref{eqn:radianceout_wnintegral} as a sum of $\numspectralbuckets$ uncorrelated exponential random variables.

The conditional distribution of $\radianceoutgblur{\scl}\paren{\pointcamalt}$ can therefore be approximated as a Gamma distribution with shape parameter $\numspectralbuckets$:
\begin{align}\label{eqn:radianceoutcohgrid_conditional_gamma}
	\radianceoutgblur{\scl}\paren{\gridcenters} \mid \foradianceoutgblur{\scl}\paren{\gridcenters} &\sim \Gam\paren{\numspectralbuckets, \frac{\numspectralbuckets}{\foradianceoutgblur{\scl}\paren{\gridcenters}}}.
\end{align}
where we similarly define
\begin{align}
	\foradianceoutgblur{\scl}\paren{\pointcamalt}
	&\coloneq \int_{\R^+}\!\! \foradianceoutgblur{\scl}\paren{\pointcamalt,\wavenumber} \ssf\paren{\wavenumber} \spectruminc\paren{\wavenumber} \ud \wavenumber
\end{align}

\begin{prp}[label={pro:secondorder_statistical_model}]{Model for second-order texture}{texturemodel}
	The irradiance at point $\pointcam$ can be approximately modeled as
	\begin{align}
		\irradiance\paren{\pointcam} &\approx
	\sum_{\gridindex}\scl^2 \psf\paren{\pointcam - \gridcenters} \radianceoutgblur{\scl}\paren{\gridcenters}
	\end{align}
	where the (blurred) outgoing radiance at grid locations $\gridcenters$ with spacing $\scl$ can be modeled as Gamma-distributed random variables
	\begin{align}
		\radianceoutgblur{\scl}\paren{\gridcenters} \mid \foradianceoutgblur{\scl}\paren{\gridcenters} &\sim \Gam\paren{\numspectralbuckets, \frac{\numspectralbuckets}{\foradianceoutgblur{\scl}\paren{\gridcenters}}}.
	\end{align}
\end{prp}

We correspondingly define a first-order counterpart of the approximate irradiance
	\begin{align}
		\foirradiance\paren{\pointcam} &\approx
	\sum_{\gridindex}\scl^2 \psf\paren{\pointcam - \gridcenters} \foradianceoutgblur{\scl}\paren{\gridcenters}
	\end{align}

Using the above model, we show in \cref{sec:numareas_proof} that $\cfvirradiance$ can be written as
\begin{align}
	\cfvirradiance &\coloneq \frac{1}{\numspectralbuckets\numwindows}\paren{1 + \cfvforadianceoutgblur{\scl}} + \cfvfoirradiance \label{eqn:cfvirradiance_foexact}.
\end{align}
where the respective coefficients of variation for $\foirradiance$ and $\foradianceoutgblur{\scl}$
\begin{align}
	\cfvfoirradiance &\coloneq \frac{\Var\bracket{\foirradiance\paren{\pointcam}}}{\marginalmeanirradiance^2},\\
	\cfvforadianceoutgblur{\scl} &\coloneq \frac{\Var\bracket{\foradianceoutgblur{\scl}\paren{\gridcenters}}}{\E{\foradianceoutgblur{\scl}\paren{\gridcenters}}^2}.
\end{align}

If we further assume that first-order texture is independent across the grid locations $\gridcenters$ (\ie $\foirradiance\paren{\gridcenters}$ are independent across $\gridcenters$), then we can write $\cfvfoirradiance = \frac{1}{\numwindows}\cfvforadianceoutgblur{\scl}$, resulting in the simplified expression
\begin{align}
	\cfvirradiance &= \frac{1}{\numspectralbuckets\numwindows}\paren{1 + \cfvforadianceoutgblur{\scl}} + \frac{1}{\numwindows}\cfvforadianceoutgblur{\scl}.\label{eqn:cfvirradiance_foindep}\\
	&= \frac{1}{\numspectralbuckets\numwindows}\paren{1 + \numwindows\cfvfoirradiance} + \cfvfoirradiance
\end{align}
thus proving \cref{pro:cfvirradiance}.


\subsection{Proof of exponential distribution}
\label{sec:expdistproof}

Plugging \cref{eqn:radianceout_ft} into \cref{eqn:radianceoutgblur_defn}, we have
\begin{align}
		\radianceoutgblur{\gridspacing}\paren{\pointcamaltalt,\wavenumber}
		&=
		\frac{1}{\brdfconst} 
		\int_{\R^2} \int_{\R^2} 
		\normgausskernelgrid\paren{\pointcamalt - \pointcamaltalt} 
		\pstf\paren{\pointcamalt, \pointcamdiff, \wavenumber}\mcfint\paren{2\pointcamdiff, \wavenumber}
		e^{\imi \wavenumber \dirvecout_{xy} \cdot \paren{2\pointcamdiff}}
		\ud{\pointcamdiff} 
		\ud\pointcamalt,\\
		&= 
		\frac{2}{\pi \brdfconst\gridspacing^2}
		\int_{\R^2} \int_{\R^2} 
		 e^{-2\frac{\norm{\pointcamalt-\pointcamaltalt}^2}{\gridspacing^2}}
		\pstf\paren{\pointcamalt, \pointcamdiff, \wavenumber}\mcfint\paren{2\pointcamdiff, \wavenumber}
		e^{\imi \wavenumber \dirvecout_{xy} \cdot \paren{2\pointcamdiff}}
		\ud{\pointcamdiff} 
		\ud\pointcamalt
\end{align}
To proceed we consider the special case of the coherence function for Gaussian environment lighting (\crefrange{eqn:gaussian_lighting}{eqn:gaussian_mcfint}):
\begin{align}\label{eqn:gaussian_mcf}
	\mcfint\paren{2\pointcamdiff, \wavenumber}&=\irradianceinc\paren{\wavenumber} \exp\paren{-\frac{\norm{2\pointcamdiff}^2}{2\scl^2}}\exp\paren{\imi \wavenumber \dirvecpeak_{xy} \cdot 2\pointcamdiff} 
\end{align}
Using a change of variables $\temppt_1 = \pointcamalt + \pointcamdiff$, $\temppt_2 = \pointcamalt - \pointcamdiff$, we have
\begin{align}
	\radianceoutgblur{\gridspacing}\paren{\pointcamaltalt,\wavenumber}  \!&=\!\irradianceinc\paren{\wavenumber} \frac{2e^{-\frac{2\norm{\pointcamaltalt}^2}{\gridspacing^2}}}{\pi \brdfconst\gridspacing^2 } \cdot \nonumber\\
	&\int_{\R^2}\int_{\R^2}\!\!
		\scatter\paren{\temppt_1, \wavenumber} 
		e^{-\frac{\norm{\temppt_1}^2}{2}\paren{\frac{1}{\gridspacing^2} + \frac{1}{\scl^2}} + \frac{2\pointcamaltalt\cdot\temppt_1}{\gridspacing^2}}
		e^{\imi \wavenumber \paren{\dirvecout_{xy} + \dirvecpeak_{xy}}\cdot {\temppt_1}}
		\nonumber\\
		&\scatter^\ast\paren{\temppt_2, \wavenumber}
		e^{-\frac{\norm{\temppt_2}^2}{2}\paren{\frac{1}{\gridspacing^2} + \frac{1}{\scl^2}} + \frac{2\pointcamaltalt\cdot\temppt_2}{\gridspacing^2}}
		e^{-\imi \wavenumber \paren{\dirvecout_{xy} + \dirvecpeak_{xy}}\cdot {\temppt_2}}
		\nonumber\\
		&\exp\paren{-{\temppt_1 \cdot \temppt_2}\paren{\frac{1}{\gridspacing^2} - \frac{1}{\scl^2}}}
		\ud{\temppt_1}	  \ud\temppt_2
\end{align}
The integrand above is separable in $\temppt_1$ and $\temppt_2$ except for the cross-term in the last exponential above.
Setting $\gridspacing = \scl$, the cross-term in the exponential vanishes, and the above integral then simplifies to 
\begin{align}
	\radianceoutgblur{\scl}\paren{\pointcamaltalt,\wavenumber}  \!&=\!\frac{2\irradianceinc\paren{\wavenumber} }{\pi\brdfconst\scl^2} \abs{ 
		\int_{\R^2}\!\!
		\scatter\paren{\temppt, \wavenumber}
		e^{-\frac{\norm{\temppt - \pointcamaltalt}^2}{\scl^2}}  
		e^{\imi \wavenumber \paren{\dirvecout_{xy} + \dirvecpeak_{xy}}\cdot \temppt}
		\ud{\temppt}
	}^2\\
	&= \irradianceinc\paren{\wavenumber} \frac{2\pi\scl^2}{\brdfconst}
	 \abs{ 
		\int_{\R^2}\!\!
		\normgausskernel{\sqrt{2}\scl}\paren{\temppt - \pointcamaltalt}
		\scatter\paren{\temppt, \wavenumber}
		e^{\imi \wavenumber \paren{\dirvecout_{xy} + \dirvecpeak_{xy}}\cdot \temppt}
		\ud{\temppt}
	}^2
\end{align}
Without ensemble averaging over the distribution of surface heightfields, the above corresponds to \emph{speckle} with a Gaussian window of width ${\sqrt{2}\scl}$ around $\pointcamaltalt$ on the surface. Since it is a magnitude-squared of a sum of a large number of independent complex random variables (considering correlation length of the height field is sufficiently small relative to $\scl$), we can argue (as in \citet{goodman2020speckle}) that the real and imaginary parts of the integral are therefore normally distributed random variables (by central limit theorem), thereby making the blurred spectral density of outgoing radiance $\radianceoutgblur{\scl}\paren{\pointcamaltalt,\wavenumber}$ exponentially distributed.

\subsection{Spatial integration}
\label{sec:numareas_proof}
From \cref{eqn:irradiance_total_approx} we have the conditional mean of irradiance given the first-order blurred outgoing radiance $\foradianceoutgblur{\scl}\paren{\pointcam}$ as:
\begin{align}
	\E{\irradiance\paren{\pointcam} \mid \foradianceoutgblur{\scl}} &=  \sum_{\gridindex} \scl^2 \psf\paren{\pointcam - \gridcenters} \E{\radianceoutgblur{\scl}\paren{\gridcenters} \mid \foradianceoutgblur{\scl}\paren{\gridcenters}}\\
	&=  \sum_{\gridindex} \scl^2 \psf\paren{\pointcam - \gridcenters} \foradianceoutgblur{\scl}\paren{\gridcenters}\\
	&\approx \int_{\R^2} \psf\paren{\pointcam - \pointtwod} \foradianceoutgblur{\scl}\paren{\pointtwod} \ud \pointtwod
\end{align}
And the marginal mean of irradiance as (using ergodicity):
\begin{align}
	\E{\irradiance\paren{\pointcam}}	
	&= \int_{\R^2} \psf\paren{\pointcam - \pointtwod} \Exp{\height}{\foradianceoutgblur{\scl}\paren{\pointtwod}} \ud \pointtwod.\\
	&= \marginalmeanradianceoutgblur{\scl} = \marginalmeanirradiance
\end{align}
since the PSF integrates to unity.
The conditional variance
\begin{align}
	\Var\bracket{\irradiance \mid \foradianceoutgblur{\scl}} &= 
	\Var\bracket{\sum_{\gridindex} \paren{\scl^2 \psf\paren{\pointcam - \gridcenters}} {\radianceoutgblur{\scl}\paren{\gridcenters}}}\\
	&=\sum_{\gridindex} \paren{\scl^2 \psf\paren{\pointcam - \gridcenters}}^2 \Var\bracket{\radianceoutgblur{\scl}\mid \foradianceoutgblur{\scl}}\\
	&= \sum_{\gridindex} \paren{\scl^2 \psf\paren{\pointcam - \gridcenters}}^2\frac{\paren{\foradianceoutgblur{\scl}\paren{\gridcenters}}^2}{\numspectralbuckets}
\end{align}
using conditional independence of $\radianceoutgblur{\scl}\paren{\gridcenters} \mid \foradianceoutgblur{\scl}\paren{\gridcenters}$ across grid locations.
Therefore the marginal variance of irradiance is given by
\begin{align}
	\Var\bracket{\irradiance} &= \Var\bracket{\E{\irradiance \mid \foradianceoutgblur{\scl}}} + \Exp{}{\Var\bracket{\irradiance \mid \foradianceoutgblur{\scl}}}
\end{align}
The first term
\begin{align}
	\Var\bracket{\E{\irradiance \mid \foradianceoutgblur{\scl}}} &= \Var\bracket{\sum_{\gridindex} \scl^2 \psf\paren{\pointcam - \gridcenters} \foradianceoutgblur{\scl}\paren{\gridcenters}}\\
	&= \Var\bracket{\foirradiance\paren{\pointcam}}
\end{align}
Further simplification isn't possible without information about the correlation structure of first-order texture $\foradianceoutgblur{\scl}\paren{\gridcenters}$ across grid locations, which are not necessarily independent. We will therefore leave it as is.
The second term
\begin{align}
	\E{\Var\bracket{\irradiance \mid \foradianceoutgblur{\scl}}} &= \E{\sum_{\gridindex} \paren{\scl^2 \psf\paren{\pointcam - \gridcenters}}^2\frac{\paren{\foradianceoutgblur{\scl}\paren{\gridcenters}}^2}{\numspectralbuckets}}\\
	&= \frac{1}{\numspectralbuckets}\sum_{\gridindex} \paren{\scl^2 \psf\paren{\pointcam - \gridcenters}}^2\E{\paren{\foradianceoutgblur{\scl}\paren{\gridcenters}}^2}\\
	&= \E{\paren{\foradianceoutgblur{\scl}}^2}\frac{\scl^2}{\numspectralbuckets} \sum_{\gridindex} \scl^2 \paren{ \psf\paren{\pointcam - \gridcenters}}^2\\
	&\approx \E{\paren{\foradianceoutgblur{\scl}}^2}\frac{\scl^2}{\numspectralbuckets} \int_{\R^2} \paren{ \psf\paren{\pointcam - \pointtwod}}^2 \ud \pointtwod
\end{align}

Taking the PSF to be a Gaussian of width $\dfl$ (with parameter $\dfl/2$) centered at $\pointcam$,
\begin{align}\label{eqn:gaussian_psf}
	\psf\paren{\pointcam - \pointtwod} &= \frac{2}{\pi \dfl^2} \exp\paren{-2\frac{\norm{\pointcam - \pointtwod}^2}{\dfl^2}}
\end{align}
\begin{align}
	\int_{\R^2} \paren{ \psf\paren{\pointcam - \pointtwod}}^2 \ud \pointtwod &= \frac{1}{\pi \dfl^2}
\end{align}
and thus the pointwise marginal variance of irradiance
\begin{align}
	\Var\bracket{\irradiance\paren{\pointcam}} &= \frac{1}{\numspectralbuckets}\paren{\frac{\scl^2}{\pi\dfl^2}}\E{\paren{\foradianceoutgblur{\scl}}^2} + \Var\bracket{\foirradiance\paren{\pointcam}}\\
	&= \frac{1}{\numspectralbuckets}\paren{\frac{\scl^2}{\pi\dfl^2}}\paren{\marginalmeanradianceoutgblur{\scl}^2 + \Var\bracket{\foradianceoutgblur{\scl}}} + \Var\bracket{\foirradiance\paren{\pointcam}}
\end{align}
Therefore the coefficient of variation of irradiance is given by
\begin{align}
	\cfvirradiance &= \frac{\Var\bracket{\irradiance\paren{\pointcam}}}{\E{\irradiance\paren{\pointcam}}^2}\\
	&= \frac{1}{\numspectralbuckets}\paren{\frac{\scl^2}{\pi\dfl^2}}\paren{1 + \frac{\Var\bracket{\foradianceoutgblur{\scl}}}{\E{\foradianceoutgblur{\scl}}^2}} + \frac{\Var\bracket{\foirradiance}}{\E{\foirradiance}^2}\\
	&= \frac{1}{\numspectralbuckets\numwindows}\paren{1 + \cfvforadianceoutgblur{\scl}} + \cfvfoirradiance
\end{align}
since $\marginalmeanirradiance = \E{\foirradiance\paren{\pointcam}} = \E{\foradianceoutgblur{\scl}} = \marginalmeanradianceoutgblur{\scl}$, and defining $\numwindows = \frac{\pi \dfl^2}{\scl^2}$ as the effective number of coherence areas contributing to the irradiance at a sensor point through the PSF.







\section{Noise model}
\label{sec:noisemodel}
The (noisy) intensity $\intensitynoisy$ (in DN, Digital Numbers) recorded by the sensor given an incident flux $\flux$ $\bracket{\unit{\watt}}$ is modeled as follows
\begin{align}\label{eqn:noisemodel_full}
	\begin{split}
	\intensitynoisy\paren{\pointcam} &= \min\paren{ \left\lfloor\gain (\min\paren{\photoncount,\fullwellcapacity} + \readnoisepreamp) + \readnoisepostamp \right\rfloor, \adcmax}\\
	\photoncount \mid \flux &\sim \Poisson{\paren{\photoncurrentperflux \flux + \darkcurrent}\exptime}\\
	\readnoisepreamp &\sim \Normal\paren{0, \sigmapreamp^2}, \quad \bracket{\unit{\electron}}\\
	\readnoisepostamp\paren{\pointcam} &\sim \Normal\paren{0, \sigmapostamp^2}, \quad \bracket{\unit{\digitalnumber}}\\
	\photoncurrentperflux &= \frac{ \quantumefficiency}{\energyperphoton}, \quad \bracket{\unit{\photoelectron  \joule^{-1}}}\\
	\energyperphoton &= \frac{\planck \lightspeed}{\wavelength}, \quad \bracket{\unit{\joule}}
	\end{split}
\end{align}
where, $\fullwellcapacity$ is the full-well capacity of the pixel (in electrons), $\adcmax$ is the maximum ADC value (in DN),
 $\gain$ is the gain factor (in $\bracket{\unit{\digitalnumber \per \electron}}$) determined by the ISO setting,
  $\quantumefficiency$ is the quantum efficiency,
$\darkcurrent$ is the dark current (in $\bracket{\unit{\electron \second^{-1}}}$), 
$\exptime$ is the exposure time (in $\bracket{\unit{\second}}$),
 $\readnoisepreamp$ is the pre-amplifier read noise (in electrons),
  and $\readnoisepostamp$ is the post-amplifier read noise (in DN) separate from noise due to ADC quantization. Parameters of the above noise model used for our numerical studies for \cref{fig:mc-sim} are specified in \cref{tab:sensorparams}.
  
\paragraph{Simplifying assumptions for analysis.}
For analytical tractability of (noisy) focus measure statistics, we make the following simplifying assumptions:
\begin{enumerate}
	\item We operate in a regime where saturation effects (full-well and ADC) can be neglected.
	\item We ignore quantization effects, \ie, we consider $\intensitynoisy$ to be a continuous random variable.
	\item We consider dark current to be negligible, \ie, $\darkcurrent \approx 0$.
\end{enumerate}
Resulting in the simplified model:
\begin{align}\label{eqn:noisemodel_analysis}
	\begin{split}
	\intensitynoisy\paren{\pointcam} &= \gain \paren{\photoncount + \readnoisepreamp} + \readnoisepostamp \\
	\photoncount  \mid \flux &\sim \Poisson{\photoncurrentperflux\flux \exptime}
	\end{split}
\end{align}
Under the above assumptions, we define the noise-free intensity $\intensity$ as the conditional mean of $\intensitynoisy$ given the incident flux $\flux$:
\begin{align}
	\intensity \coloneq \Exp{}{\intensitynoisy \mid \flux} &=   \frac{\photoncurrentperflux  \exptime}{\gainvar} \flux\\
	\Var\bracket{\intensitynoisy \mid \flux} = \Var\bracket{\intensitynoisy \mid \intensity} &=
	\frac{1}{\gainvar^2}
	\bracket{
		 \photoncurrentperflux  \exptime \flux + 
		 {\sigmapreamp^2}}
		 + \sigmapostamp^2
	\\
	&= 		 \frac{\intensity}{\gainvar} + 
		 \frac{{\sigmapreamp^2}}{\gainvar^2}
		 + \sigmapostamp^2
\end{align}
We may rewrite \cref{eqn:noisemodel_analysis} in an additive form, defining $\noisevar \coloneq \intensitynoisy - \intensity$:
\begin{align} \label{eqn:conditionalnoisevariance}
	\begin{split}
	\intensitynoisy &= \intensity + \noisevar\\
	\Var\bracket{\noisevar \mid \intensity} = \Var\bracket{\intensitynoisy \mid \intensity} &=
		 \frac{\intensity}{\gainvar} + 
		 \frac{{\sigmapreamp^2}}{\gainvar^2}
		 + \sigmapostamp^2
	\end{split}
\end{align}
which is the model described in \cref{sec:background} (excluding saturation).

\begin{enumerate}[resume]
	\item As a final simplification, we assume the  incident irradiance $\irradiance\paren{\pointcam}$ $[\unit{\watt\per\metre\squared}]$ is uniform over the pixel area 
 	 \begin{align}
		\flux &= \irradiance\paren{\pointcam} \pixelarea, \quad \bracket{\unit{\watt}}
	\end{align}
	where $\pixelarea$ is the pixel area (in $\bracket{\unit{\metre^2}}$).
	This therefore implies the noise-free intensity is proportional to the incident irradiance as follows: 
	\begin{align} \label{eqn:intensity_irradiance_prop_assumption}
		\intensity &= \frac{\photoncurrentperirradiance \exptime}{\gainvar}\irradiance\paren{\pointcam} \\
		\photoncurrentperirradiance &\coloneq \frac{ \quantumefficiency \pixelarea}{\energyperphoton}, \quad \bracket{\unit{\photoelectron  \joule^{-1} \metre^2}}
	\end{align}
We may also write the conditional mean and variance of $\intensitynoisy$ given $\irradiance$ as:
\begin{align}
	\begin{split}
	\Exp{}{\intensitynoisy \mid \irradiance} &=   \frac{\photoncurrentperirradiance  \exptime}{\gainvar} \irradiance\\
	\Var\bracket{\intensitynoisy \mid \irradiance} &=
	\frac{1}{\gainvar^2}
	\bracket{
		 \photoncurrentperirradiance  \exptime \irradiance + 
		 {\sigmapreamp^2}}
		 + \sigmapostamp^2
	\end{split}
\end{align}
\label{item:fluxproportionality}

\end{enumerate}

We do not make the above assumptions (other than proportionality of flux to irradiance, \cref{item:fluxproportionality}) in our numerical implementation \cref{fig:mc-sim}, which is implemented according to \cref{eqn:noisemodel_full} with practical values chosen for sensor parameters including full-well capacity, dark current \etc, that can be found in \cref{tab:sensorparams}. Further details of the numerical implementation can be found in \cref{sec:mcstudy_supp}.

\section{Depth from focus analysis}
\label{sec:dffanalysis}
As described in \cref{sec:background}, given pixel intensity measurements in a patch $\neighborhood_\numsamples$ of $\numsamples$ pixels, the squared sample coefficient of variation focus measure
\begin{equation}
	\csqnoisy_{\focusindex}\paren{\pointcam} \coloneq \frac{\fmvaluenoisy_{\focusindex}\paren{\pointcam}}{\meanintnoisy_{\focusindex}\paren{\pointcam}},
\end{equation}
is computed from the \emph{sample variance} and \emph{sample mean} of patch intensities,
\begin{align}
	\meanintnoisy_{\focusindex}\paren{\pointcam} &\coloneq \frac{1}{\numsamples}\!\!\sum_{\pointcam'\in\neighborhood_\numsamples\paren{\pointcam}} \!\!\intensitynoisy_{\focusindex}\paren{\pointcam'}, \\
	\fmvaluenoisy_{\focusindex}\paren{\pointcam} &\coloneq \frac{1}{\numsamples- 1}\!\!\sum_{\pointcam'\in\neighborhood_\numsamples\paren{\pointcam}} \!\!\paren{\intensitynoisy_{\focusindex}\paren{\pointcam'}- \meanintnoisy_{\focusindex}\paren{\pointcam}}^2
\end{align}

The noise-free focus measure $\csq$ is defined similarly :
\begin{align}
	\csq_{\focusindex}\paren{\pointcam} &\coloneq \frac{\fmvalue_{\focusindex}\paren{\pointcam}}{\meanint_{\focusindex}\paren{\pointcam}},\\
	\fmvalue_{\focusindex}\paren{\pointcam} &\coloneq \frac{1}{\numsamples- 1}\!\!\sum_{\pointcam'\in\neighborhood_\numsamples\paren{\pointcam}} \!\!\paren{\intensity_{\focusindex}\paren{\pointcam'}- \meanint_{\focusindex}\paren{\pointcam}}^2\\
	\meanint_{\focusindex}\paren{\pointcam} &\coloneq \frac{1}{\numsamples}\!\!\sum_{\pointcam'\in\neighborhood_\numsamples\paren{\pointcam}} \!\!\intensity_{\focusindex}\paren{\pointcam'}, 
\end{align}

For the purposes of analysis, we also define a normalized version of the sample variance
\begin{align}\label{eqn:normsv_defn}
	\normsvnoisy\paren{\pointcam} &\coloneq \frac{\fmvaluenoisy\paren{\pointcam}}{\marginalmeanintensity^2}\\
	\normsv\paren{\pointcam}	&\coloneq \frac{\fmvalue\paren{\pointcam}}{\marginalmeanintensity^2}
\end{align}
where $\marginalmeanintensity = \E{\intensity} = \E{\intensitynoisy}$ is the marginal mean of intensities over the patch. This normalized sample variance is distinct from the actual focus measure defined above (and not directly measureable exactly). For analytical tractability, we study the statistics of $\normsvnoisy$ instead of $\csqnoisy$, to approximately characterize $\csqnoisy$.

\subsection{Statistics of the focus measure}
\label{sec:fmstats}

Given the simplified sensor noise model in \cref{eqn:conditionalnoisevariance}, we can denote the pointwise marginal mean and variance of measured (noisy) pixel intensity $\intensitynoisy$ as
\begin{align}
	\marginalmeanintensitynoisy \coloneq \Exp{}{\readout}  &=   \marginalmeanintensity\\
	\marginalvarintensitynoisy \coloneq \Var\bracket{\readout} &=
	\marginalvarintensity +  \frac{\marginalmeanintensity}{\gainvar} + 
		 \frac{{\sigmapreamp^2}}{\gainvar^2}
		 + \sigmapostamp^2		
\end{align}
where the marginal mean and variance of true intensities (as a consequence of assuming uniform irradiance over the pixel area, \cref{eqn:intensity_irradiance_prop_assumption}) are given by
\begin{align}
	\marginalmeanintensity \coloneq \Exp{}{\intensity} &= \frac{\photoncurrentperirradiance \exptime}{\gainvar} \marginalmeanirradiance\\
	\marginalvarintensity \coloneq \Var\bracket{\intensity} &= \frac{\photoncurrentperirradiance^2 \exptime^2}{\gainvar^2} \marginalvarirradiance
\end{align}
and as a consequence of this assumption,
\begin{align}
	\cfvintensity &\coloneq \frac{\marginalvarintensity}{\marginalmeanintensity^2} = \frac{\marginalvarirradiance}{\marginalmeanirradiance^2} = \cfvirradiance
\end{align}
We can now write the coefficient of variation of noisy intensity as
\begin{align}\label{eqn:cfvintensitynoisy}
	\cfvintensitynoisy \coloneq \frac{\marginalvarintensitynoisy}{\marginalmeanintensitynoisy^2} &= \cfvintensity+\cfvnoise\\
	&=\cfvirradiance + \cfvnoise
\end{align}

where we defined in \cref{eqn:cfvnoise} the squared reciprocal of pixel intensity SNR as the coefficient of variation of noise $\cfvnoise$:
\begin{align}\tag{rep. \ref{eqn:cfvnoise}}
	\cfvnoise &\coloneq \frac{\marginalvarnoise}{\marginalmeanintensity^2} = \frac{1}{\exptime\photoncurrentperflux\marginalmeanflux}\paren{1 + \frac{\sigmaread^2}{\exptime\photoncurrentperflux\marginalmeanflux}}
\end{align}
with the marginal variance of noise $\noisevar$,
\begin{align}\label{eqn:marginalvarnoise}
	\marginalvarnoise \coloneq \Var\bracket{\noisevar} = \E{\Var\bracket{\noisevar \mid \intensity}} + \Var\bracket{0} &=
		 \frac{\marginalmeanintensity}{\gainvar} + 
		 \frac{{\sigmapreamp^2}}{\gainvar^2}
		 + \sigmapostamp^2
\end{align}
using the conditional variance from \cref{eqn:conditionalnoisevariance}.
The pointwise marginal distribution of pixel intensity $\intensitynoisy$ at each pixel is identical across pixels. We make the additional assumption that these pixel intensities are \emph{independent across pixels}.\footnote{This assumption is technically strictly valid only when the $f$-number of the camera is sufficiently small, such that the width of the PSF is smaller than the pixel size (see \cref{sec:lenssettings}). If the PSF width is too large, then adjacent pixels will have correlated intensities since they see overlapping regions of the surface. We use this assumption to simplify the analysis of focus measure statistics.}
Therefore the sample variance of noisy pixel intensities has marginal statistics:
\begin{align*}
	\Exp{}{\fmvaluenoisy} &= \Var\bracket{\readout} = \marginalvarintensitynoisy\\
	\Var\bracket{\fmvaluenoisy} &= \frac{\paren{\marginalvarintensitynoisy}^2}{\numsamples}\left(\kurtosis - \frac{\numsamples-3}{\numsamples-1}\right)
\end{align*}
where
$\kurtosis$ is the kurtosis (ratio of fourth moment to squared-variance) of the marginal distribution of pixel intensities. If the distribution of pixel intensities is approximated to be to Gaussian, then $\kurtosis \approx 3$.

Normalizing by $\marginalmeanintensitynoisy^2$ we then have the marginal statistics of $\normsvnoisy$:
\begin{lem}[label={lem:marginal_fmstats}]{Marginal statistics of $\normsvnoisy$}{fmstats}
\begin{align} \label{eqn:meancsq_general}
	\E{\normsvnoisy} &= \frac{\marginalvarintensitynoisy}{\marginalmeanintensitynoisy^2} = \cfvintensitynoisy = \cfvintensity +\cfvnoise\\
	\Var\bracket{\normsvnoisy} &= \frac{\paren{\cfvintensitynoisy}^2}{\numsamples}\paren{\kurtosis - \frac{\numsamples-3}{\numsamples-1}}\\
	&= \frac{2\paren{\cfvintensitynoisy}^2}{\numsamples-1} \quad \text{(if $\kurtosis \approx 3$, effectively $\intensitynoisy\sim \Normal\paren{\marginalmeanintensity, \marginalvarnoise}$)}
\end{align}
\end{lem}

\emph{Statistics of focus measure for first-order textureless surface.}
For a first-order textureless surface, we have $\cfvintensity = \cfvirradiance = \nicefrac{1}{\numwindows\numspectralbuckets}$, and so we have
 \begin{align*}
	\E{\normsvnoisy} &= 	\frac{1}{\numwindows\numspectralbuckets} 
  + \cfvnoise\\
\Var\bracket{\normsvnoisy} &= \frac{2}{\numsamples - 1}\curly{
	\frac{1}{\numwindows\numspectralbuckets} 
  + \cfvnoise
	}^2
\end{align*}


\subsection{Proof of \cref{pro:marginal_recoverability}}
\label{sec:recoverabilityproof}
\cref{sec:fmstats} establishes the mean and variance of $\normsvnoisy$. We now present a proof for \cref{pro:marginal_recoverability}, characterizing the $\control$-recoverability of a surface from depth from focus, as defined in \cref{def:recoverability}.

We indicate the in-focus and out-of-focus normalized sample variances as $\normsvnoisyif$ and $\normsvnoisyoof$ respectively, and the corresponding noise-free focus measures as $\normsvif$ and $\normsvoof$.

\begin{enumerate}
	\item 
	$\normsvnoisyif$ is the value of $\normsvnoisy$ evaluated at the sharpest focus setting of the camera.
	\item $\normsvnoisyoof$ is the value of $\normsvnoisy$ evaluated at a focus setting of the camera that is sufficiently far from the in-focus setting, such that the true noise-free intensities in the patch are effectively constant (\ie, the texture is completely blurred out), or $\intensity\paren{\pointcam} \approx \marginalmeanintensity$. This therefore implies $\normsvoof = 0$.
\end{enumerate}

To proceed, we model $\normsvnoisyif$ and $\normsvnoisyoof$ in and out of focus to be normally-distributed random variables, with the mean and variance expressions derived in \cref{sec:fmstats}. The conclusions of this analysis are supported by our Monte-Carlo simulations in \cref{fig:mc-sim}, thereby justifying the use of this assumption for the purpose of analytical tractability.


Modeling $\normsvnoisyif$ and $\normsvnoisyoof$ to be normally-distributed, with the mean and variance derived in \cref{lem:marginal_fmstats}.
\begin{align}
	\normsvnoisyif &\sim \Normal\paren{\cfvintensity + \cfvnoise,\quad \frac{2}{\numsamples-1}\paren{\cfvintensity + \cfvnoise}^2}\\
	\normsvnoisyoof &\sim \Normal\paren{\cfvnoise,\quad \frac{2}{\numsamples-1}\paren{\cfvnoise}^2}
\end{align}

In addition, since the out-of-focus intensity variations in $\normsvnoisyoof$
are purely due to sensor noise, we can say that $\normsvnoisyif$ and $\normsvnoisyoof$ are uncorrelated random variables. Therefore,
\begin{align}
\Prob\paren{{\normsvnoisy_{\rm if}} > {\normsvnoisy}_{\rm oof}}
 &= 	\cdf\paren{\frac{\Exp{}{\normsvnoisy_{\rm if}} - \Exp{}{\normsvnoisy_{\rm oof}}
		}{\sqrt{
		\Var\bracket{\normsvnoisy_{\rm if}} + \Var\bracket{\normsvnoisy_{\rm oof}}
		}}
		}\\
	&= 
	\cdf\paren{\frac{ \cfvintensity
	}{\sqrt{
		\frac{2}{\numsamples - 1}\curly{
			\paren{
	\cfvintensity
  + \cfvnoise
				}^2 
				+
				\paren{\cfvnoise}^2
				}
		}}
		}\\
		&= 
	\cdf\paren{\frac{\contrastsnrprodmarginal^2}{\sqrt{
		\frac{2}{\numsamples - 1}\curly{
			\paren{	\contrastsnrprodmarginal^2
  + 1
				}^2
				+
				1
				}		}}
		}
\end{align}
where $\cdf$ is the CDF of the standard normal distribution, and we've defined $\contrastsnrprodmarginal^2$ in \cref{sec:recoverability}.
\begin{align}
	\contrastsnrprodmarginal^2 &\coloneq \frac{\cfvintensity}{\cfvnoise}
\end{align}
\begin{lem}[label={lem:proberror}]{Probability of Error}{proberror}
	Modeling $\normsvnoisyif$ and $\normsvnoisyoof$ to be normally-distributed, with the mean and variance derived in \cref{lem:marginal_fmstats}, the \emph{probability of error}
	\begin{align}
		\Prob\paren{{\normsvnoisy_{\rm if}} < {\normsvnoisy}_{\rm oof}} = 1 - \cdf\paren{\frac{\contrastsnrprodmarginal^2}{\sqrt{
			\frac{2}{\numsamples - 1}\curly{
				\paren{	\contrastsnrprodmarginal^2
				+ 1
				}^2
				+
				1
				}		}}
				}
	\end{align}
\end{lem}

Therefore we have a necessary and sufficient condition for $\control$-recoverability (\cref{def:recoverability}) of a surface using depth from focus:
\begin{align}
	\frac{\contrastsnrprodmarginal^2 }{\sqrt{1 + \paren{\contrastsnrprodmarginal^2 + 1}^2}}\cdot \sqrt{\frac{\numsamples -1}{2}} > \cdf^{-1}\paren{1 - \control}
\end{align}
thus proving \cref{pro:marginal_recoverability}.

Notice that as $\contrastsnrprodmarginal$ is increased, the LHS saturates to $\sqrt{\frac{\numsamples -1}{2}}$, which is the maximum possible value of the argument of the CDF. Therefore, for a fixed value of $\numsamples$, there is a maximum achievable probability of correct recovery, which is given by $\cdf\paren{\sqrt{\frac{\numsamples -1}{2}}}$. 



\section{Recoverability with second-order texture}
\label{sec:recoverability_spectralfiltered}

In this section, we derive \cref{pro:snr}, which characterizes the contrast-SNR product $\contrastsnrprodmarginal$ as a function of exposure time $\exptime$ and bandwidth $\bwk$, for a first-order textureless surface. For a first-order textureless surface, we have $\cfvfoirradiance = 0$, and so from \cref{pro:cfvirradiance}
\begin{align}
	\cfvintensity &= \frac{1}{\numwindows\numspectralbuckets}
\end{align}
with $\numspectralbuckets \approx G_1 \bwk$ (\cref{eqn:numspectralbuckets}).
We can similarly expand $\cfvnoise$ using the noise model in \cref{eqn:cfvnoise}.
\begin{align}
	\cfvnoise &= 
	 \frac{1}{\exptime\photoncurrentperflux\marginalmeanflux}\paren{1 + \frac{\sigmaread^2}{\exptime\photoncurrentperflux\marginalmeanflux}}
\end{align}

Spectral filtering reduces the mean flux $\marginalmeanflux$ by a factor of $\effectiveattenfactor$ (described in \cref{sec:spectral_filter_attenuation}), which is a function of the filter bandwidth $\bwk$.
\begin{align}\label{eqn:effectiveatten_approx}
	\effectiveattenfactor\coloneq \frac{\marginalmeanfluxpost}{\marginalmeanfluxpre} \approx \peakfilteratten \frac{\bwk}{\bwksrc}
\end{align} 
where $\marginalmeanfluxpre$ and $\marginalmeanfluxpost$ are the mean flux before and after spectral filtering respectively, and $\peakfilteratten$ is the peak attenuation of the spectral filter. For sufficiently narrow spectral filter bandwidth, the effective bandwidth after filtering is approximately the same as that of the filter, and so we abuse notation and use $\bwk$ for both the spectral filter bandwidth as well as the effective bandwidth after filtering. We denote the fixed bandwidth of the original (unfiltered) source spectrum by $\bwksrc$ (written $\bwk$ in \cref{eqn:spectruminitial_gaussian}). These are clarified in \cref{sec:spectral_filter_attenuation}.

We define
\begin{align}	
	G_2 &\coloneq \frac{\photoncurrentperflux \marginalmeanfluxpre \cdot \peakfilteratten}{\bwksrc}
\end{align}
and plugging in $\photoncurrentperflux\marginalmeanfluxpost = G_2 \bwk$ into the expression for $\cfvnoise$, we have the contrast-SNR product for a first-order textureless surface as a function of exposure time and (filter) bandwidth
	\begin{align}
		\contrastsnrprodmarginal^2\paren{\exptime, \Delta\wavenumber} &\approx \frac{1}{\numwindows}
		 \cdot \frac{G_2}{G_1} 
		 \cdot\frac{\exptime}{\paren{1 + \frac{\sigma^2_{\rm read}}{G_2} \frac{1}{\bwk \exptime}}}
	\end{align}
Since bandwidth affects $\marginalmeanfluxpost$, the upper limit of exposure time until saturation is bandwidth-dependent. A simple notion to model this is to limit the exposure time until $\marginalmeanintensity = \photoncurrentperflux\exptime\frac{\marginalmeanfluxpost}{\gainvar} = \frac{G_2 \bwk}{\gainvar}\exptime$ equals $\adcmax$.
	\begin{align}
		\exptime \in \left[0, \frac{1}{\Delta\wavenumber}\cdot\frac{\gainvar \adcmax}{G_2}\right)
	\end{align}
thus resulting in \cref{pro:snr}.

\subsection{Attenuation due to spectral filtering}
\label{sec:spectral_filter_attenuation}
Consider the original normalized spectral profile (before spectral filtering) is denoted as $\spectruminitial\paren{\wavenumber}$, as in \cref{eqn:spectruminitial}.
Let the attenuation profile of the spectral filter introduced be $\filteratten\paren{\wavenumber}$. 
Thus the effective attenuation in irradiance due to the spectral filter is given by the factor
\begin{align}
	\effectiveattenfactor\coloneq \frac{\marginalmeanfluxpost}{\marginalmeanfluxpre} &= \int_{\R^+} \filteratten\paren{\wavenumber}\spectruminitial\paren{\wavenumber} \ud\wavenumber
\end{align}

\subsection{Gaussian spectral profile}
The spectral profile of various light sources and object reflectance are wide and varied.
However, to build intuition for the effect of spectral mismatch between the spectral filter and the illumination and reflectance spectrum, we consider a Gaussian profile for both $\spectruminitial\paren{\wavenumber}$ and $\filteratten\paren{\wavenumber}$.

We have previously introduced a Gaussian form for $\spectruminitial$ in \cref{eqn:spectruminitial_gaussian}
\begin{align}
	\spectruminitial\paren{\wavenumber} &= \frac{1}{\sqrt{2\pi}\bwksrc} \exp\paren{-\frac{\paren{\wavenumber - \mwkoriginal}^2}{2\bwksrc^2}}
\end{align}
The spectral filter has a peak attenuation $\peakfilteratten$,
\begin{align}
		\filteratten\paren{\wavenumber} &= \peakfilteratten\exp\paren{-\frac{\paren{\wavenumber - \mwkfilter}^2}{2\bwfilter^2}}
\end{align}
The effective spectral profile after introducing the spectral filter will also maintain a Gaussian form, 
\begin{align}
	\spectruminitial\paren{\wavenumber} \filteratten\paren{\wavenumber} &= \effectiveattenfactor \cdot \bracket{\frac{1}{\sqrt{2\pi}\bweff} \exp\paren{-\frac{\paren{\wavenumber - \mwkeffective}^2}{2\bweff^2}}}
\end{align}
with effective mean wavenumber and bandwidth given by
\begin{align}
	\mwkeffective &= \mwkoriginal \frac{\bwfilter^2}{\bwfilter^2 + \bwksrc^2} + \mwkfilter \frac{\bwksrc^2}{\bwfilter^2 + \bwksrc^2}\\
	\frac{1}{\bweff^2} &= \frac{1}{\bwksrc^2} + \frac{1}{\bwfilter^2}  
\end{align}
and effective attenuation factor
\begin{align}
	\effectiveattenfactor &= \peakfilteratten \frac{\bwfilter}{\sqrt{\bwksrc^2 + \bwfilter^2}} \exp\paren{- \frac{\paren{\mwkoriginal - \mwkfilter}^2}{2\paren{\bwksrc^2 + \bwfilter^2}}}.
\end{align}
implying a quadratic-exponential attenuation as the filter center frequency $\mwkfilter$ moves away from the original center frequency $\mwkoriginal$.

\paragraph{Narrowband limit.} If we consider a spectral filter with a bandwidth $\bwfilter\ll \bwksrc$, then we can approximate the effective attenuation factor as
\begin{align}
	\effectiveattenfactor &\approx \peakfilteratten \frac{\bwfilter}{\bwksrc} \exp\paren{- \frac{\paren{\mwkoriginal - \mwkfilter}^2}{2\bwksrc^2}}
\end{align}
which is linear in the filter bandwidth $\bwfilter$. Further, the effective bandwidth after filtering $\bweff$ is approximately the same as the filter bandwidth $\bwfilter$ in this case.
If we additionally consider that the width of the original spectrum $\bwksrc$ is sufficiently wide compared to the spectral mismatch $\paren{\mwkoriginal - \mwkfilter}$, then we can further approximate the attenuation factor as
\begin{align}\label{eqn:effectiveattenfactor_approx}
	\effectiveattenfactor &\approx \peakfilteratten \frac{\bwfilter}{\bwksrc} \approx \peakfilteratten \frac{\bweff}{\bwksrc}
\end{align}

\subsection{Wavenumber and wavelength bandwidths}
In our results, we specify bandwidths in terms of \emph{wavelength} bandwidths as $\bwl$ (specified in $\unit{\nano\metre}$) because this is the common convention for specification of spectral filters as is more familiar to readers. However, we need to convert this to wavenumber bandwidth $\bwk$ for our analysis and monte-carlo numerical study (\cref{eqn:numspectralbuckets}). Considering a central wavelength $\mwl$, we consider the mean (angular) wavenumber $\mwk = \nicefrac{2\pi}{\mwl}$.
\begin{align}
\wavenumber_\mathrm{low} &= \frac{2\pi}{\mwl + \frac{\bwl}{2}}\\
\wavenumber_\mathrm{high} &= \frac{2\pi}{\mwl - \frac{\bwl}{2}}\\
\bwk &= \wavenumber_\mathrm{high} - \wavenumber_\mathrm{low}\\
&= 2\pi \cdot \frac{\bwl}{\mwl^2 - \frac{\bwl^2}{4}} 
\end{align}
Under the condition that $\bwl \ll \mwl$,
\begin{align}
	\bwk &\approx \frac{2\pi}{\mwl} \cdot \frac{\bwl}{\mwl} = \mwk \cdot \frac{\bwl}{\mwl}
\end{align}

%% file: interfpd.bib
@String(PAMI  = {IEEE Trans. Pattern Anal. Mach. Intell.})

@String(IJCV  = {Int. J. Comput. Vis.})

@String(CVPR  = {IEEE Conf. Comput. Vis. Pattern Recog.})

@String(ICCV  = {Int. Conf. Comput. Vis.})

@String(ECCV  = {Eur. Conf. Comput. Vis.})

@String(ACCV  = {Asian Conf. Comput. Vis.})

@String(TOG   = {ACM Trans. Graph.})

@String(CGF   = {Comput. Graph. Forum})

@String(ICCP     = {IEEE Int. Conf. Comput. Photography})

@String(SIGGRAPH = {ACM SIGGRAPH})

@String(OPTICA   = {Optica})

@String(PRL      = {Phys. Rev. Lett.})

@String(NATCOMM  = {Nat. Commun.})

@String(JOSA     = {J. Opt. Soc. Am.})

@String(JOPT     = {J. Opt.})

@String(OPTCON   = {Opt. Contin.})

@String(OPTENG   = {Opt. Eng.})

@String(AOP      = {Adv. Opt. Photonics})

@String(APB      = {Appl. Phys. B})

@String(APPLSCI  = {Appl. Sci.})

@String(PRLETT   = {Pattern Recognit. Lett.})

@String(IJOM     = {Int. J. Optomechatron.})

@String(LASERSCI = {Laser Science})

@book{mertz2019introduction,
  title={Introduction to optical microscopy},
  author={Mertz, Jerome},
  year={2019},
  publisher={Cambridge University Press}
}

@article{levin2013fabricating,
  title={Fabricating {BRDF}s at high spatial resolution using wave optics},
  author={Levin, Anat and Glasner, Daniel and Xiong, Ying and Durand, Fr{\'e}do and Freeman, William and Matusik, Wojciech and Zickler, Todd},
  journal = TOG,
  volume={32},
  number={4},
  pages={1--14},
  year={2013},
  publisher={ACM New York, NY, USA}
}

@article{muroi2023capturing,
  title={Capturing videos at 60 frames per second using incoherent digital holography},
  author={Muroi, Tetsuhiko and Nobukawa, Teruyoshi and Katano, Yutaro and Hagiwara, Kei},
  journal = OPTCON,
  volume={2},
  number={11},
  pages={2409--2420},
  year={2023},
  publisher={Optica Publishing Group}
}

@article{cuypers2012reflectance,
  title={Reflectance model for diffraction},
  author={Cuypers, Tom and Haber, Tom and Bekaert, Philippe and Oh, Se Baek and Raskar, Ramesh},
  journal = TOG,
  volume={31},
  number={5},
  pages={1--11},
  year={2012},
  publisher={ACM New York, NY, USA}
}

@Article{	  Davy2013,
  title		= {Green's Function Retrieval and Passive Imaging from
		  Correlations of Wideband Thermal Radiations},
  author	= {Davy, Matthieu and Fink, Mathias and de Rosny, Julien},
  journal = PRL,
  volume	= {110},
  issue		= {20},
  pages		= {203901},
  numpages	= {5},
  year		= {2013},
  month		= {May},
  publisher	= {American Physical Society},
}

@Article{	  Badon2015,
  title		= {Retrieving Time-Dependent {Green's} Functions in Optics with
		  Low-Coherence Interferometry},
  author	= {Badon, Amaury and Lerosey, Geoffroy and Boccara, Albert C.
		  and Fink, Mathias and Aubry, Alexandre},
  journal = PRL,
  volume	= {114},
  issue		= {2},
  pages		= {023901},
  numpages	= {5},
  year		= {2015},
  month		= {Jan},
  publisher	= {American Physical Society},
}

@Article{	  Badon2016,
  author	= {Amaury Badon and Dayan Li and Geoffroy Lerosey and A.
		  Claude Boccara and Mathias Fink and Alexandre Aubry},
  journal = OPTICA,
  number	= {11},
  pages		= {1160--1166},
  publisher	= {OSA},
  title		= {Spatio-temporal imaging of light transport in highly
		  scattering media under white light illumination},
  volume	= {3},
  month		= {Nov},
  year		= {2016},
}

@Article{	  BogerLombard2019,
  author	= {Boger-Lombard, Jeremy and Katz, Ori},
  title		= {Passive optical time-of-flight for non line-of-sight
		  localization},
  journal = NATCOMM,
  year		= {2019},
  month		= {Jul},
  day		= {26},
  volume	= {10},
  number	= {1},
  pages		= {3343},
}

@article{kingslake1945effective,
  title={The effective aperture of a photographic objective},
  author={Kingslake, R},
  journal = JOSA,
  volume={35},
  number={8},
  pages={518--520},
  year={1945},
  publisher={Optical Society of America}
}

@Article{	  Batarseh2018,
  author	= {Batarseh, M. and Sukhov, S. and Shen, Z. and Gemar, H. and
		  Rezvani, R. and Dogariu, A.},
  title		= {Passive sensing around the corner using spatial
		  coherence},
  journal = NATCOMM,
  year		= {2018},
  month		= {Sep},
  day		= {07},
  volume	= {9},
  number	= {1},
  pages		= {3629},
}

@article{tahara2022roadmap,
  title={Roadmap of incoherent digital holography},
  author={Tahara, Tatsuki and Zhang, Yaping and Rosen, Joseph and Anand, Vijayakumar and Cao, Liangcai and Wu, Jiachen and Koujin, Takako and Matsuda, Atsushi and Ishii, Ayumi and Kozawa, Yuichi and Okamoto, Ryo and Oi, Ryutaro and Nobukawa, Teruyoshi and Choi, Kihong and Imbe, Masatoshi and Poon, Ting-Chung},
  journal = APB,
  volume={128},
  number={11},
  pages={193},
  year={2022},
  publisher={Springer}
}

@article{tahara2017single,
  title={Single-shot phase-shifting incoherent digital holography},
  author={Tahara, Tatsuki and Kanno, Takeya and Arai, Yasuhiko and Ozawa, Takeaki},
  journal = JOPT,
  volume={19},
  number={6},
  pages={065705},
  year={2017},
  publisher={IOP Publishing}
}

@inproceedings{tahara2022palm,
  title={Palm-sized single-shot phase-shifting incoherent digital holography system with birefringent materials},
  author={Tahara, Tatsuki and Kozawa, Yuichi and Ishii, Ayumi and Okamoto, Ryo},
  booktitle = LASERSCI,
  pages={JTu5B--52},
  year={2022},
  organization={Optica Publishing Group}
}

@article{liu2018incoherent,
  title={Incoherent digital holography: a review},
  author={Liu, Jung-Ping and Tahara, Tatsuki and Hayasaki, Yoshio and Poon, Ting-Chung},
  journal = APPLSCI,
  volume={8},
  number={1},
  pages={143},
  year={2018},
  publisher={MDPI}
}

@article{rosen2019recent,
  title={Recent advances in self-interference incoherent digital holography},
  author={Rosen, Joseph and Vijayakumar, A and Kumar, Manoj and Rai, Mani Ratnam and Kelner, Roy and Kashter, Yuval and Bulbul, Angika and Mukherjee, Saswata},
  journal = AOP,
  volume={11},
  number={1},
  pages={1--66},
  year={2019},
  publisher={Optica Publishing Group}
}

@InProceedings{	  Punnappurath_2019_CVPR,
  author	= {Punnappurath, Abhijith and Brown, Michael S.},
  title		= {Reflection Removal Using a Dual-Pixel Sensor},
  booktitle = CVPR,
  year		= {2019},
  pages = {1556--1565},
}

@InProceedings{	  xin2021defocus,
  title		= {Defocus map estimation and deblurring from a single
		  dual-pixel image},
  author	= {Xin, Shumian and Wadhwa, Neal and Xue, Tianfan and Barron,
		  Jonathan T and Srinivasan, Pratul P and Chen, Jiawen and
		  Gkioulekas, Ioannis and Garg, Rahul},
  booktitle = ICCV,
  pages		= {2228--2238},
  year		= {2021}
}

@article{gkioulekas2015transient,
	author = {Gkioulekas, Ioannis and Levin, Anat and Durand, Fr\'{e}do and Zickler, Todd},
	title = {Micron-scale light transport decomposition using interferometry},
	year = {2015},
	issue_date = {August 2015},
	publisher = {Association for Computing Machinery},
	address = {New York, NY, USA},
	volume = {34},
	number = {4},
	journal = TOG,
	month = jul,
	articleno = {37},
	numpages = {14},
	pages = {37:1--37:14}
}

@InProceedings{	  garg2019learning,
  title		= {Learning single camera depth estimation using
		  dual-pixels},
  author	= {Garg, Rahul and Wadhwa, Neal and Ansari, Sameer and
		  Barron, Jonathan T.},
  booktitle = ICCV,
  year		= {2019},
  pages = {7628--7637},
}

@InProceedings{	  punnappurath2020modeling,
  author	= {Abhijith Punnappurath and Abdullah Abuolaim and Mahmoud
		  Afifi and Michael S. Brown},
  booktitle = ICCP,
  title		= {Modeling Defocus-Disparity in Dual-Pixel Sensors},
  year		= {2020},
  pages = {1--12},
}

@article{schechner2000depth,
  title={Depth from defocus vs. stereo: How different really are they?},
  author={Schechner, Yoav Y and Kiryati, Nahum},
  journal = IJCV,
  volume={39},
  pages={141--162},
  year={2000},
  publisher={Springer}
}

@article{nayar1994shape,
  title={Shape from focus},
  author={Nayar, Shree K and Nakagawa, Yasuo},
  journal = PAMI,
  volume={16},
  number={8},
  pages={824--831},
  year={1994},
  publisher={IEEE}
}

@article{hasinoff2009confocal,
  title={Confocal stereo},
  author={Hasinoff, Samuel W and Kutulakos, Kiriakos N},
  journal = IJCV,
  volume={81},
  pages={82--104},
  year={2009},
  publisher={Springer}
}

@inproceedings{kotwal2023passive,
  title={Passive micron-scale time-of-flight with sunlight interferometry},
  author={Kotwal, Alankar and Levin, Anat and Gkioulekas, Ioannis},
  booktitle = CVPR,
  pages={4139--4149},
  year={2023}
}

@inproceedings{chen2024coherence,
  title={Coherence As Texture---Passive Textureless {3D} Reconstruction by Self-interference},
  author={Chen, Wei-Yu and Sankaranarayanan, Aswin C and Levin, Anat and O'Toole, Matthew},
  booktitle = CVPR,
  pages={25058--25066},
  year={2024}
}

@article{kim2013scene,
  title={Scene reconstruction from high spatio-angular resolution light fields.},
  author={Kim, Changil and Zimmer, Henning and Pritch, Yael and Sorkine-Hornung, Alexander and Gross, Markus H},
  journal = TOG,
  volume={32},
  number={4},
  pages={73:1--73:12},
  year={2013}
}

@book{akkermans2007mesoscopic,
  title={Mesoscopic physics of electrons and photons},
  author={Akkermans, Eric and Montambaux, Gilles},
  year={2007},
  publisher={Cambridge university press}
}

@book{wolf2007introduction,
  title={Introduction to the Theory of Coherence and Polarization of Light},
  author={Wolf, Emil},
  year={2007},
  publisher={Cambridge university press}
}

@inproceedings{cossairt2014digital,
  title={Digital refocusing with incoherent holography},
  author={Cossairt, Oliver and Matsuda, Nathan and Gupta, Mohit},
  booktitle = ICCP,
  pages={1--9},
  year={2014},
  organization={IEEE}
}

@inproceedings{chakrabarti2012depth,
  title={Depth and deblurring from a spectrally-varying depth-of-field},
  author={Chakrabarti, Ayan and Zickler, Todd},
  booktitle = ECCV,
  pages={648--661},
  year={2012},
  organization={Springer}
}

@article{dong2015predicting,
  title={Predicting appearance from measured microgeometry of metal surfaces},
  author={Dong, Zhao and Walter, Bruce and Marschner, Steve and Greenberg, Donald P},
  journal = TOG,
  volume={35},
  number={1},
  pages={1--13},
  year={2015},
  publisher={ACM New York, NY, USA}
}

@InProceedings{	  zhou2009good,
  title		= {What are good apertures for defocus deblurring?},
  author	= {Zhou, Changyin and Nayar, Shree K.},
  booktitle = ICCP,
  year		= {2009},
  pages = {1--8},
}

@InProceedings{	  Zhou2009ICCV,
  title		= {Coded Aperture Pairs for Depth from Defocus},
  author	= {Changyin Zhou and Stephen Lin and Shree K. Nayar},
  booktitle = ICCV,
  year		= {2009},
  pages = {325--332},
}

@Article{	  veeraraghavan2007dappled,
  title		= {Dappled photography: Mask enhanced cameras for heterodyned
		  light fields and coded aperture refocusing},
  author	= {Veeraraghavan, Ashok and Raskar, Ramesh and Agrawal, Amit
		  and Mohan, Ankit and Tumblin, Jack},
  journal = TOG,
  year		= {2007},
  volume = {26},
  number = {3},
  pages = {69},
}

@Article{	  levin2007image,
  title		= {Image and depth from a conventional camera with a coded
		  aperture},
  author	= {Levin, Anat and Fergus, Rob and Durand, Fr{\'e}do and
		  Freeman, William T.},
  journal = TOG,
  year		= {2007},
  volume = {26},
  number = {3},
  pages = {70},
}

@inproceedings{guo2017focal,
  title={Focal track: Depth and accommodation with oscillating lens deformation},
  author={Guo, Qi and Alexander, Emma and Zickler, Todd},
  booktitle = ICCV,
  pages={966--974},
  year={2017}
}

@InProceedings{	  alexander2016focal,
  title		= {Focal flow: Measuring distance and velocity with defocus
		  and differential motion},
  author	= {Alexander, Emma and Guo, Qi and Koppal, Sanjeev and
		  Gortler, Steven and Zickler, Todd},
  booktitle = ECCV,
  pages		= {667--682},
  year		= {2016},
  organization	= {Springer}
}

@Article{	  watanabe1998rational,
  title		= {Rational filters for passive depth from defocus},
  author	= {Watanabe, Masahiro and Nayar, Shree K.},
  journal = IJCV,
  year		= {1998},
  volume = {27},
  number = {3},
  pages = {203--225},
}

@InProceedings{	  Tang2017,
  author	= {Tang, Huixuan and Cohen, Scott and Price, Brian and
		  Schiller, Stephen and Kutulakos, Kiriakos N.},
  booktitle = CVPR,
  title		= {Depth from Defocus in the Wild},
  year		= {2017},
  pages = {2740--2748},
}

@Article{	  Pentland87,
  author	= {Pentland, Alex Paul},
  title		= {A new sense for depth of field},
  journal = PAMI,
  year		= {1987},
  volume = {9},
  number = {4},
  pages = {523--531},
}

@InProceedings{	  favaro2010recovering,
  title		= {Recovering thin structures via nonlocal-means
		  regularization with application to depth from defocus},
  author	= {Favaro, Paolo},
  booktitle = CVPR,
  year		= {2010},
  pages = {1133--1140},
}

@Article{	  Subbarao1994defocus,
  title		= {Depth from defocus: A spatial domain approach},
  journal = IJCV,
  year		= {1994},
  author	= {Subbarao, Murali and Surya, Gopal},
  volume = {13},
  number = {3},
  pages = {271--294},
}

@article{bolles1987epipolar,
  title={Epipolar-plane image analysis: An approach to determining structure from motion},
  author={Bolles, Robert C and Baker, H Harlyn and Marimont, David H},
  journal = IJCV,
  volume={1},
  number={1},
  pages={7--55},
  year={1987},
  publisher={Springer}
}

@Article{	  grossman1987focus,
  title		= {Depth from focus},
  journal = PRLETT,
  year		= {1987},
  author	= {P. Grossmann},
  volume = {5},
  number = {1},
  pages = {63--69},
}

@InProceedings{	  hazirbas18ddff,
  author	= {Hazirba\c{s}, Caner and Soyer, Sebastian Georg and Staab, Maximilian Christian and Leal-Taix\'e, Laura and Cremers, Daniel},
  title		= {Deep Depth From Focus},
  booktitle = ACCV,
  year		= {2018},
  pages = {525--541},
}

@InProceedings{	  Supasorn2015,
  author	= {Supasorn Suwajanakorn and Carlos Hernandez and Steven M.
		  Seitz},
  booktitle = CVPR,
  title		= {Depth from focus with your mobile phone},
  year		= {2015},
  pages = {3497--3506},
}

@book{goodman2020speckle,
  title={Speckle phenomena in optics: theory and applications},
  author={Goodman, Joseph W},
  year={2020},
  publisher={SPIE}
}

@inproceedings{sundaram1997textureless,
  title={Are textureless scenes recoverable?},
  author={Sundaram, Hari and Nayar, Shree},
  booktitle = CVPR,
  pages={814--820},
  year={1997},
  organization={IEEE}
}

@article{nayar1996real,
  title={Real-time focus range sensor},
  author={Nayar, Shree K and Watanabe, Masahiro and Noguchi, Minori},
  journal = PAMI,
  volume={18},
  number={12},
  pages={1186--1198},
  year={1996},
  publisher={IEEE}
}

@inproceedings{hasinoff2010noise,
  title={Noise-optimal capture for high dynamic range photography},
  author={Hasinoff, Samuel W and Durand, Fr{\'e}do and Freeman, William T},
  booktitle = CVPR,
  pages={553--560},
  year={2010},
  organization={IEEE}
}

@article{subbarao1995accurate,
  title={Accurate recovery of three-dimensional shape from image focus},
  author={Subbarao, Muralidhara and Choi, Tao},
  journal = PAMI,
  volume={17},
  number={3},
  pages={266--274},
  year={1995},
  publisher={IEEE}
}

@article{subbarao1998selecting,
  title={Selecting the optimal focus measure for autofocusing and depth-from-focus},
  author={Subbarao, Muralidhara and Tyan, J-K},
  journal = PAMI,
  volume={20},
  number={8},
  pages={864--870},
  year={1998},
  publisher={IEEE}
}

@inproceedings{jo2015spedo,
  title={Spedo: 6 dof ego-motion sensor using speckle defocus imaging},
  author={Jo, Kensei and Gupta, Mohit and Nayar, Shree K},
  booktitle = ICCV,
  pages={4319--4327},
  year={2015}
}

@inproceedings{stam1999diffraction,
  title={Diffraction shaders},
  author={Stam, Jos},
  booktitle = SIGGRAPH,
  pages={101--110},
  year={1999}
}

@book{beckmann1987scattering,
  title={The scattering of electromagnetic waves from rough surfaces},
  author={Beckmann, Petr and Spizzichino, Andre},
  year={1987},
  publisher = {Artech House},
}

@Book{		  hartley_zisserman_2004,
  title		= {Multiple View Geometry in Computer Vision},
  author	= {Hartley, Richard and Zisserman, Andrew},
  year		= {2004},
  publisher	= {Cambridge University Press}
}

@Article{	  Nalpantidis2008stereo,
  author	= {Nalpantidis, Lazaros and Sirakoulis, Georgios and
		  Gasteratos, Antonios},
  year		= {2008},
  title		= {Rev. of Stereo Vision Algorithms: Software to Hardware},
  journal = IJOM,
  volume = {2},
  number = {4},
  pages = {435--462},
}

@Article{	  scharstein2002taxonomy,
  title		= {A taxonomy and evaluation of dense two-frame stereo
		  correspondence algorithms},
  author	= {Scharstein, Daniel and Szeliski, Richard},
  journal = IJCV,
  year		= {2002},
  volume = {47},
  number = {1},
  pages = {7--42},
}

@Article{	  Barnard1980disparity,
  author	= {Barnard, Stephen T. and Thompson, William B.},
  journal = PAMI,
  title		= {Disparity Analysis of Images},
  year		= {1980},
  volume = {2},
  number = {4},
  pages = {333--340},
}

@article{smith2017colux,
  title={{CoLux}: Multi-object {3D} micro-motion analysis using speckle imaging},
  author={Smith, Brandon M and Desai, Pratham and Agarwal, Vishal and Gupta, Mohit},
  journal = TOG,
  volume={36},
  number={4},
  pages={1--12},
  year={2017},
  publisher={ACM New York, NY, USA}
}

@article{fienup1988imaging,
  title={Imaging correlography with sparse arrays of detectors},
  author={Fienup, James R and Idell, Paul S},
  journal = OPTENG,
  volume={27},
  number={9},
  pages={778--784},
  year={1988},
  publisher={SPIE}
}

@article{boniface2019noninvasive,
  title={Noninvasive light focusing in scattering media using speckle variance optimization},
  author={Boniface, Antoine and Blochet, Baptiste and Dong, Jonathan and Gigan, Sylvain},
  journal = OPTICA,
  volume={6},
  number={11},
  pages={1381--1385},
  year={2019},
  publisher={Optical Society of America}
}

@inproceedings{smith2018tracking,
  title={Tracking multiple objects outside the line of sight using speckle imaging},
  author={Smith, Brandon M and O'Toole, Matthew and Gupta, Mohit},
  booktitle = CVPR,
  pages={6258--6266},
  year={2018}
}

@inproceedings{sheinin2022dual,
  title={Dual-shutter optical vibration sensing},
  author={Sheinin, Mark and Chan, Dorian and O'Toole, Matthew and Narasimhan, Srinivasa G},
  booktitle = CVPR,
  pages={16324--16333},
  year={2022}
}

@inproceedings{kichler2025learning,
  title={Learning to See Inside Opaque Liquid Containers using Speckle Vibrometry},
  author={Kichler, Matan and Bagon, Shai and Sheinin, Mark},
  booktitle = ICCV,
  pages={9466--9476},
  year={2025}
}

@inproceedings{shih2012laser,
  title={Laser speckle photography for surface tampering detection},
  author={Shih, Yi Chang and Davis, Abe and Hasinoff, Samuel W and Durand, Fr{\'e}do and Freeman, William T},
  booktitle = CVPR,
  pages={33--40},
  year={2012},
  organization={IEEE}
}

@inproceedings{xie2024wavemo,
  title={{WaveMo}: learning wavefront modulations to see through scattering},
  author={Xie, Mingyang and Guo, Haiyun and Feng, Brandon Y and Jin, Lingbo and Veeraraghavan, Ashok and Metzler, Christopher A},
  booktitle = CVPR,
  pages={25276--25285},
  year={2024}
}

@article{alterman2021imaging,
  title={Imaging with local speckle intensity correlations: theory and practice},
  author={Alterman, Marina and Bar, Chen and Gkioulekas, Ioannis and Levin, Anat},
  journal = TOG,
  volume={40},
  number={3},
  pages={1--22},
  year={2021},
  publisher={ACM New York, NY}
}

@article{bar2019monte,
  title={A Monte Carlo framework for rendering speckle statistics in scattering media},
  author={Bar, Chen and Alterman, Marina and Gkioulekas, Ioannis and Levin, Anat},
  journal = TOG,
  volume={38},
  number={4},
  pages={1--22},
  year={2019},
  publisher={ACM New York, NY, USA}
}

@article{liu2025fully,
  title={A Fully-statistical Wave Scattering Model for Heterogeneous Surfaces},
  author={Liu, Zhengze and Huo, Yuchi and Peng, Yifan and Wang, Rui},
  journal = TOG,
  volume={44},
  number={4},
  pages={1--17},
  year={2025},
  publisher={ACM New York, NY, USA}
}

@article{bar2020rendering,
  title={Rendering near-field speckle statistics in scattering media},
  author={Bar, Chen and Gkioulekas, Ioannis and Levin, Anat},
  journal = TOG,
  volume={39},
  number={6},
  pages={1--18},
  year={2020},
  publisher={ACM New York, NY, USA}
}

@article{kim2025monte,
  title={A Monte Carlo Rendering Framework for Simulating Optical Heterodyne Detection},
  author={Kim, Juhyeon and Benko, Craig and Wrenninge, Magnus and Villemin, Ryusuke and Barber, Zeb and Jarosz, Wojciech and Pediredla, Adithya},
  journal = TOG,
  volume={44},
  number={4},
  pages={1--19},
  year={2025},
  publisher={ACM New York, NY, USA}
}

@article{metzler2020deep,
  title={Deep-inverse correlography: towards real-time high-resolution non-line-of-sight imaging},
  author={Metzler, Christopher A and Heide, Felix and Rangarajan, Prasana and Balaji, Muralidhar Madabhushi and Viswanath, Aparna and Veeraraghavan, Ashok and Baraniuk, Richard G},
  journal = OPTICA,
  volume={7},
  number={1},
  pages={63--71},
  year={2020},
  publisher={Optical Society of America}
}

@article{steinberg2022rendering,
author = {Steinberg, Shlomi and Yan, Ling-Qi},
title = {Rendering of Subjective Speckle Formed by Rough Statistical Surfaces},
year = {2022},
issue_date = {February 2022},
publisher = {Association for Computing Machinery},
address = {New York, NY, USA},
volume = {41},
number = {1},
journal = TOG,
month = feb,
articleno = {2},
numpages = {23},
pages = {2:1--2:23}
}

@book{testorf2010phase,
  title={Phase-space optics: fundamentals and applications},
  author={Testorf, Markus E and Hennelly, Bryan M and Ojeda-Casta{\~n}eda, Jorge},
  publisher = {The McGraw-Hill Companies, Inc.},
  year={2010}
}

@article{steinberg2026wavetracing,
 	author = {Steinberg, Shlomi and Pharr, Matt},
 	title = {Wave Tracing: Generalizing The Path Integral To Wave Optics},
 	journal = CGF,
  pages = {e70322},
  year = {2026}
 }
